\documentclass{article}

\usepackage{graphicx}
\usepackage{wrapfig}
\usepackage{multirow} 
\usepackage{algorithm}
\usepackage{algpseudocode} 
\usepackage{tikz}
\usetikzlibrary{arrows.meta,positioning,shapes.geometric}
\usepackage{array}
\usepackage{tabularx}
\usepackage{amsmath}
\usepackage{amsthm}
\usepackage{makecell}
\usepackage[eandd,final]{neurips_2026}
\usepackage{amssymb}
\usepackage{marvosym}

\ifdefined\DeclareUnicodeCharacter
\DeclareUnicodeCharacter{202F}{\,}
\fi

\usepackage[utf8]{inputenc} 
\usepackage[T1]{fontenc}    
\usepackage{hyperref}       
\hypersetup{hidelinks}
\usepackage{url}            
\usepackage{booktabs}       
\usepackage{amsfonts}       
\usepackage{nicefrac}       
\usepackage{microtype}      
\usepackage{xcolor}         
\usepackage{placeins}
\usepackage{fvextra}

\usepackage{blindtext}
\usepackage{xcolor}
\usepackage{etoolbox}
\newif\ifcrchanges
\crchangesfalse

\theoremstyle{definition}
\newtheorem{definition}{Definition}
\newtheorem{problem}{Problem}
\newenvironment{prompttemplate}[1]{%
  \par\smallskip\noindent\emph{#1.}\par\vspace{-0.4em}%
  \begin{quote}%
}{%
  \end{quote}\vspace{-0.2em}%
}

\title{MM-OptBench: A Solver-Grounded Benchmark for Multimodal Optimization Modeling}

\author{%
  Zhong Li$^{\clubsuit}$ \quad
  Qi Huang$^{\blacklozenge}$ \quad
  Yuxuan Zhu$^{\spadesuit}$ \quad
  Mohammad Mohammadi Amiri$^{\spadesuit}$ \\
  \textbf{Niki van Stein}$^{\blacklozenge}$ \quad
  \textbf{Thomas B{\"a}ck}$^{\blacklozenge}$ \quad
  \textbf{Matthijs van Leeuwen}$^{\blacklozenge}$ \\
  \textbf{Zaiwen Wen}$^{\bigstar}$ \quad
  \textbf{Lincen Yang}$^{\blacklozenge}$\textsuperscript{(\Letter)} \\
  $^{\clubsuit}$Great Bay University,
  $^{\blacklozenge}$ Leiden University,\\
  $^{\spadesuit}$ Rensselaer Polytechnic Institute,
  $^{\bigstar}$Peking University \\
  \textsuperscript{(\Letter)} Corresponding author:
  \texttt{l.yang@liacs.leidenuniv.nl} (Lincen Yang)
}

\begin{document}

\maketitle

\begin{abstract}
    Optimization modeling translates real decision-making problems into mathematical optimization models and solver-executable implementations. Although language models are increasingly used to generate optimization formulations and solver code, existing benchmarks for this capability are almost entirely text-only. This omits many optimization-modeling tasks that arise in operational practice, where requirements are described in text but essential instance information is often conveyed through visual artifacts such as tables, graphs, maps, schedules, and planning dashboards. We introduce \emph{multimodal optimization modeling}, a benchmark setting in which models must construct both a mathematical formulation and executable solver code from a text-and-visual problem specification. To evaluate this setting, we develop a solver-grounded framework that generates structured optimization instances, verifies each instance with an exact solver, and builds both the model-facing inputs and the hidden reference files from the same verified source. We instantiate the framework as MM-OptBench, a benchmark of 780 solver-verified instances spanning 6 optimization families, 26 subcategories, and 3 structural difficulty levels. We then evaluate 9 multimodal large language models (MLLMs), including 6 frontier general-purpose models and 3 math-specialized models, with aggregate, family-level, difficulty-level, and failure-mode analyses. The results show that the task remains far from solved: the best two models reach 52.1\% and 51.3\% pass@1, while on average across the six general-purpose MLLMs, pass@1 is 43.4\% on easy instances and 15.9\% on hard instances. All three math-specialized MLLMs solve 0/780 instances. Post-hoc extraction diagnostics and oracle-reading evaluation reveal difficulties in both multimodal grounding and downstream solver-artifact generation. MM-OptBench provides a concrete testbed for solver-grounded, decision-oriented multimodal intelligence. The dataset and code are available at \url{https://github.com/ZhongLIFR/MM-OptBench}.
\end{abstract}

\section{Introduction}

Optimization modeling~\citep{hart2017pyomo,xiao2025survey} translates real decision-making tasks into mathematical optimization models and solver-executable implementations. In practice, \textit{domain stakeholders} describe goals, operating rules, resource limits, data sources, and application semantics; \textit{optimization specialists} then formalize these semantics as decision variables, feasibility constraints, and objectives; and finally implementation-oriented \textit{modelers} encode the formulation in solver code. Importantly, optimization modeling supports decision-making in domains such as logistics, manufacturing, energy, and finance~\citep{antoniou2007practical,singh2012overview,anand2017comparative}. Yet reliable modeling remains difficult to scale because it requires domain knowledge, operations-research expertise, careful data--semantics alignment, and repeated validation of both formulation and implementation~\citep{xiao2023chain,huang2025orlm,jiang2025llmopt}. Even small mistakes---for example, a missing coupling constraint or a mismatched parameter binding---can change the feasible region or objective and lead to invalid decisions. These costs and risks make automated optimization modeling an important research direction~\citep{fan2025artificial,xiao2025survey,wang2025large}.

\begin{figure}[h]
    \centering
    \includegraphics[width=0.98\linewidth]{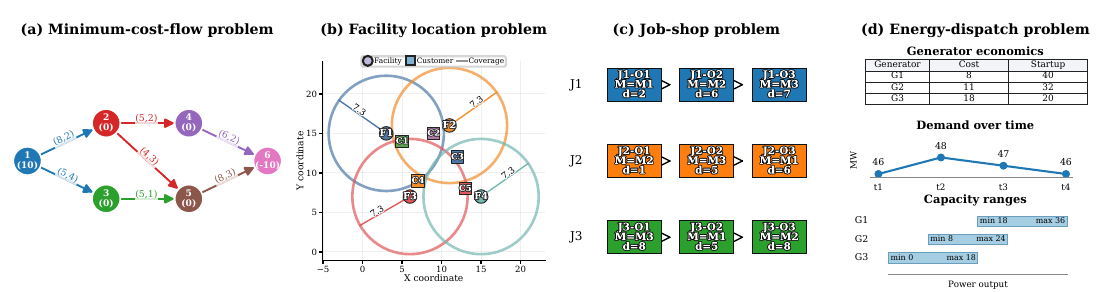}
    \caption{Compact exemplar MM-OptBench instances in which textual rules and structured visual artifacts jointly specify the optimization model. Full-size examples are in Figures~\ref{fig:appendix_flow_graph}, \ref{fig:appendix_facility_diagram}, \ref{fig:appendix_schedule_gantt}, and~\ref{fig:appendix_energy_dispatch}.}
\label{fig:MLLMOPT_Examples}
\end{figure}

As Figure~\ref{fig:MLLMOPT_Examples} illustrates, optimization-modeling inputs are often inherently multimodal. Across these examples, problem-defining information is distributed across textual instructions and structured visual carriers: a network diagram encodes topology and arc attributes in a minimum-cost-flow instance (Figure~\ref{fig:MLLMOPT_Examples}a), a spatial layout determines coverage and assignment relations in facility location (Figure~\ref{fig:MLLMOPT_Examples}b), an operation chart specifies precedence and machine requirements in job-shop scheduling (Figure~\ref{fig:MLLMOPT_Examples}c), and dispatch panels combine temporal demand with generator attributes in energy dispatch (Figure~\ref{fig:MLLMOPT_Examples}d). Thus, the visual artifacts are not auxiliary illustrations, but part of the model-facing specification from which variables, constraints, and objectives must be inferred and encoded.

Recent LLM-based optimization-modeling methods have made rapid progress, but they are still almost entirely built around the \emph{natural-language-to-optimization} (NL-to-Opt) paradigm. Multi-agent and workflow systems decompose text-only modeling into drafting, execution, and repair~\citep{xiao2023chain,ahmaditeshnizi2024opt}; supervised fine-tuning methods train on text-only problem--solution pairs, often checked by solver execution, to generate formulations or solver code~\citep{huang2025orlm,lu2025optmath}; and reinforcement-learning methods use solver outcomes as verifiable supervision~\citep{chen2025solver,xie2025murka}. The surrounding benchmark ecosystem is similarly text-centric, including NL4Opt~\citep{Ramamonjison2023NL4Opt}, IndustryOR~\citep{huang2025orlm}, ComplexOR~\citep{xiao2023chain}, MAMO~\citep{huang2024mamo}, ReSocratic~\citep{yang2024optibench}, OptMath~\citep{lu2025optmath}, LogiOR~\citep{yang2025automated}, Bench4Opt~\citep{wang2025orgeval}, and MIPLIB-NL~\citep{li2026constructing}. These resources have advanced automated optimization modeling, but leave out cases in which essential data and relations are carried by visual artifacts.

The broader multimodal literature does not close this gap. Existing datasets have advanced educational question answering (QA) and scientific-media understanding~\citep{lu2022learn,pramanick2024spiqa,tanaka2024instructdoc,mathew2021docvqa}, chart/table/graph reasoning~\citep{kahou2017figureqa,masry2022chartqa,wu2025tablebench,roberts2025grab}, and visual mathematics or expert-level multimodal evaluation~\citep{lu2024mathvista,zhang2024mathverse,yue2024mmmu,he2024olympiadbench}. However, their evaluation is usually answer-centric: models answer questions, solve posed problems, describe visual content, or generate generic artifacts. Even generation-oriented benchmarks such as ChartMimic and Plot2Code~\citep{yang2024chartmimic,wu2025plot2code} target generic code rather than optimization-specific formulations and solver implementations. Thus, the path from multimodal specifications to an optimization model---with decision variables, constraints, and objectives---and its executable solver code remains outside the scope of existing multimodal benchmarks.

We study this missing setting by introducing \emph{multimodal optimization modeling}: given a textual task description and one or more structured visual artifacts, a model must produce both a mathematical formulation and solver-executable code. We also propose a solver-grounded benchmark construction framework that generates structured optimization instances, verifies them with exact solvers, and derives aligned model-facing inputs, canonical formulations, reference solvers, verified solutions, and metadata from a common source of truth. This design makes evaluation auditable at both the symbolic and execution levels, while supporting scalable extension beyond a fixed handcrafted collection.

Using this framework, we build \textbf{MM-OptBench}, a controlled benchmark of 780 solver-verified instances with application-inspired, programmatically rendered visuals, spanning 6 optimization families, 26 subcategories, and 3 difficulty levels. The benchmark covers network optimization, location/covering/assignment, scheduling/sequencing, multi-period planning, routing/tour optimization, and combinatorial-logical models, with difficulty controlled primarily by structural complexity rather than raw numerical scale. Our first evaluation of frontier general-purpose and math-specialized MLLMs shows emerging but immature capability: GPT-5.4 and Gemini-3.1-Pro-Preview reach about half pass@1 overall, aggregate performance falls from $43.4\%$ on easy instances to $15.9\%$ on hard instances, and the tested math-specialized MLLMs solve no official instances. These results suggest that current MLLMs can write executable-looking optimization code, but still struggle to assemble globally correct decision models from distributed multimodal evidence.

In summary, our contributions are fourfold: (1) we introduce multimodal optimization modeling as a benchmark setting that is, to our knowledge, among the first to evaluate whether models can translate textual and visual specifications into mathematical formulations and solver-executable implementations (see Sec.~\ref{sec:setting}); (2) we develop a solver-grounded construction framework that uses structured generation, exact verification, and aligned artifact synthesis to produce auditable multimodal optimization instances (see Sec.~\ref{sec:benchmark} and App.~\ref{sec:appendix_construction}); (3) we release MM-OptBench, a 780-instance benchmark spanning 6 optimization families, 26 subcategories, and 3 difficulty levels (see Sec.~\ref{subsec:benchmark_design_taxonomy} and App.~\ref{sec:appendix_dev_plan}); and (4) we provide the first systematic evaluation of general-purpose and math-specialized MLLMs on multimodal optimization modeling, with aggregate, family-level, difficulty-level, and failure-mode analyses (see Sec.~\ref{sec:experiments} and App.~\ref{app:extended_experiment_results}).

\section{Related Work}
\label{sec:related_work}

MM-OptBench is positioned at the intersection of automated optimization modeling and multimodal mathematical reasoning. Prior work in these areas has advanced model construction from natural-language specifications and multimodal problem understanding, respectively, but their intersection remains largely unexplored. Existing benchmarks have not evaluated the full pipeline from multimodal optimization-modeling inputs to mathematical optimization models and the corresponding solver code. We therefore review the two lines of work separately and clarify the gap addressed by MM-OptBench; Appendix~\ref{app:extended_related_work_compact} provides a broader review and positioning within multimodal reasoning.

\textbf{Natural-language-to-optimization modeling.}
LLMs are increasingly used to turn natural-language problem descriptions into mathematical models and solver code~\citep{xiao2025survey}. Some work explores building models through dialogue with users~\citep{freuder2024conversational} or breaking the modeling task into smaller steps with separate prompts~\citep{tsouros2023holy}. Multi-agent systems assign formulation, coding, and debugging to different agents~\citep{xiao2023chain,ahmaditeshnizi2024opt}. Other work improves modeling ability through supervised fine-tuning~\citep{huang2025orlm,lu2025optmath,jiang2025llmopt} or reinforcement learning with solver feedback~\citep{chen2025solver,lu2026pearl,liu2026localizable}. In particular, other RL methods~\citep{yang2026modularized} may offer useful ideas for optimization modeling, such as designing rewards and training on progressively harder problems. Besides, note that a program may run successfully even when its model misrepresents the original problem. OptArgus~\citep{li2026optargus} uses specialized agents to detect such inconsistencies across the problem description, mathematical formulation, and code. OPT-Kraken~\citep{wu2026fix} is designed to repair these errors while keeping correct parts intact.

Progress has been closely tied to NL-to-Opt benchmarks such as NL4Opt~\citep{Ramamonjison2023NL4Opt}, IndustryOR~\citep{huang2025orlm}, ComplexOR~\citep{xiao2023chain}, MAMO~\citep{huang2024mamo}, NLP4LP~\citep{ahmaditeshnizi2024opt}, ReSocratic~\citep{yang2024optibench}, OptMath~\citep{lu2025optmath}, LogiOR~\citep{yang2025automated}, Bench4Opt~\citep{wang2025orgeval}, MIPLIB-NL~\citep{li2026constructing}, and EfficientOpt~\citep{li2026right}. These resources are valuable for studying formulation and code generation from text, but they omit settings where essential instance data are presented in tables, maps, schedules, graphs, or dashboard-like visual artifacts and must be used to construct the mathematical optimization model and corresponding solver code.

\textbf{Multimodal reasoning benchmarks.}
A separate line of work has advanced multimodal reasoning over educational QA and scientific documents~\citep{lu2022learn,pramanick2024spiqa,tanaka2024instructdoc,mathew2021docvqa}, structured visual carriers such as charts, tables, and graphs~\citep{kahou2017figureqa,masry2022chartqa,wu2025tablebench,roberts2025grab}, and visual mathematics or expert-level reasoning~\citep{lu2024mathvista,zhang2024mathverse,yue2024mmmu,he2024olympiadbench}. More recent generation-oriented or interleaved benchmarks, such as ChartMimic, Plot2Code, and MMIE~\citep{yang2024chartmimic,wu2025plot2code,xia2025mmie}, extend evaluation beyond short-answer QA toward visually grounded code generation or mixed comprehension-generation outputs. Their target, however, is still not optimization modeling: they evaluate answers, verification labels, explanations, or generic artifacts rather than a mathematical optimization model paired with executable solver code.

MM-OptBench is positioned precisely in this gap. Rather than evaluating whether a model can \emph{answer}, \emph{solve}, \emph{describe}, \emph{verify}, or \emph{generate} a generic artifact from multimodal inputs, we evaluate whether it can \emph{formulate} a correct optimization model and produce a solver-executable implementation from distributed textual and visual evidence. This positioning separates MM-OptBench from both text-only optimization-modeling benchmarks and existing multimodal reasoning benchmarks.

\section{Benchmark Setting and Solver-Grounded Evaluation}
\label{sec:setting}

Evaluating multimodal optimization modeling is different from checking a final answer: the model is asked to produce both a mathematical formulation and executable solver code, while official correctness is computed by running the generated code and checking task-specific objective agreement or logical feasibility. We therefore first define the task interface and expected outputs (Sec.~\ref{subsec:setting_task}), then describe the model-facing inputs and reference artifacts that constitute ground truth (Sec.~\ref{subsec:setting_artifacts}), formalize solver-grounded correctness and pass@$k$ scoring (Sec.~\ref{subsec:setting_protocol}), and finally separate official scoring from diagnostic failure attribution (Sec.~\ref{subsec:setting_scoring_pipeline}).

\subsection{Task Setting}
\label{subsec:setting_task}

\begin{definition}[Multimodal optimization specification]
A MM-OptBench instance is specified by $x=(T,V)$, where $T$ is a textual task description and $V=\{V_1,\dots,V_m\}$ is a set of structured visual artifacts. The text describes the modeling goal, semantic rules, and required output format, while the visuals carry information such as numerical parameters, topology, spatial relations, precedence, or temporal profiles. The specification is multimodal in the sense that the intended optimization model is determined by the joint evidence in $T$ and $V$; neither modality is assumed complete on its own.
\end{definition}

\begin{problem}[Multimodal optimization modeling]
Given a multimodal specification $x=(T,V)$, a model $f_\theta$ produces
$(M,S)=f_\theta(T,V)$, where $M$ is a mathematical formulation and $S$ is solver-executable code. The target is a complete modeling artifact: $M$ specifies indices, parameters, variables, constraints, and objective, and $S$ implements them for solver execution.
\end{problem}

This mirrors the standard optimization-modeling workflow, in which a mathematical model is specified before being translated into solver code. MM-OptBench therefore evaluates whether a model can produce this two-part modeling artifact, rather than only report a final optimum or decision.

\subsection{Benchmark Artifacts and Ground Truth}
\label{subsec:setting_artifacts}

\begin{definition}[MM-OptBench artifact package]
For each specification $x$, MM-OptBench provides a self-contained package $\mathcal{P}_x=(\mathcal{I}_x,\mathcal{R}_x)$ that separates model-facing inputs from ground-truth reference artifacts. The model-facing inputs $\mathcal{I}_x$ consist of \texttt{task\_input.txt} and visual files under \texttt{visuals/}. The ground-truth artifacts $\mathcal{R}_x$ consist of \texttt{instance\_data.json}, \texttt{math\_model.md}, \texttt{solver\_ref.py}, \texttt{solution\_ref.json}, and \texttt{meta.json}, recording the machine-readable public instance data, canonical formulation, reference implementation, verified optimum, and metadata.
\end{definition}

\subsection{Solver-Grounded Evaluation Protocol}
\label{subsec:setting_protocol}

\begin{definition}[Solver-grounded correctness]
Given a generated pair $(M,S)$ for instance $x$, MM-OptBench evaluates end-to-end correctness by executing $S$ under a fixed evaluation protocol. For optimization problems, the returned objective value must match the solver-verified optimum within the specified numerical tolerance, using the objective defined by the instance. For satisfiability problems, the returned assignment must be complete and satisfy all constraints of the instance.
\end{definition}

Scoring is based on instance-level execution outcomes, allowing different formulations and alternative optimal solutions. Let $f_i(x)\in\{0,1\}$ denote whether the $i$-th sampled generation for instance $x$ is solver-correct, and let $\mathcal{D}$ denote the benchmark set. We report
$
\text{pass@}k
=
\frac{1}{|\mathcal{D}|}
\sum_{x\in\mathcal{D}}
\mathbb{I}
\!\left(
\exists\, i\in\{1,\dots,k\}\ \text{s.t.}\ f_i(x)=1
\right),
$
so an instance is counted as solved if at least one of its $k$ sampled generations is solver-correct. We also report Valid Code Rate, the fraction of sampled outputs whose executable component runs successfully, to separate executability from solver-grounded correctness.

\subsection{Official Scoring Pipeline and Failure Attribution}
\label{subsec:setting_scoring_pipeline}

An incorrect generated artifact may fail because the model misread the multimodal instance, or because it read the instance correctly but made a formulation or coding error. MM-OptBench therefore separates scoring from diagnosis. The official submission contains only the formulation and solver artifact, and the score is computed by running that artifact. A separate diagnostic follow-up is used only after official failures to check whether the public instance data was read correctly; it does not change the official score. We define these two stages below (see App.~\ref{app:scoring_pipeline} for more details).

\begin{definition}[Stage 1: official scoring]
For instance $x$, the model receives the model-facing inputs $\mathcal{I}_x$ and returns a mathematical formulation and a solver code exposing \texttt{solve()}. The evaluation program executes \texttt{solve()} and applies the task-specific correctness criterion in Section~\ref{subsec:setting_protocol}. Valid Code Rate and pass@$k$ are computed only from Stage-1 outputs.
\end{definition}

\begin{definition}[Stage 2: diagnostic attribution]
Stage 2 is run only on Stage-1 failures. The model is asked to extract the public instance data from $\mathcal{I}_x$ without solving the problem, and the extraction is compared with ground-truth \texttt{instance\_data.json}. In this separate post-hoc check, a mismatch suggests difficulty reading the input, while a match points to possible formulation or coding errors. If the Stage-2 response cannot be parsed as structured instance data, we mark it as unresolved rather than assigning a cause. Stage 2 never changes the official score.
\end{definition}

We also test whether the model can generate a correct formulation and solver code when the problem's input data are provided as JSON-formatted text. We either provide the reference data directly (\textbf{oracle reading}) or use the model's own extracted data only when they match the reference (\textbf{verified extraction}). These tests are separate from official scoring. Figure~\ref{fig:evaluation_pipeline} (App.~\ref{app:scoring_pipeline}) gives the full protocol map, including the official scoring path, failure-attribution path, and auxiliary ablations.

\section{MM-OptBench: Benchmark Design, Taxonomy, and Construction}
\label{sec:benchmark}

This section describes how MM-OptBench is designed, constructed, and released as a benchmark for multimodal optimization modeling. We first discuss the design principles together with the taxonomy and coverage of the instances (Sec.~\ref{subsec:benchmark_design_taxonomy}), then describe the construction and validation pipeline (Sec.~\ref{subsec:benchmark_pipeline}), and the artifact contract and difficulty design used in the release (Sec.~\ref{subsec:benchmark_artifacts}).

\subsection{Design Principles, Taxonomy, and Coverage}
\label{subsec:benchmark_design_taxonomy}

MM-OptBench is designed around four connected requirements. First, instances are \emph{multimodal by construction}: the text specifies the modeling goal, semantic rules, and required output, while structured visual artifacts carry public instance data such as parameters, topology, spatial relations, precedence, and temporal profiles. Second, instances are \emph{solver-grounded}: each accepted case has a canonical formulation, executable reference solver, and verified optimum. Third, the benchmark is \emph{structurally diverse}: it probes different optimization-modeling primitives rather than many surface variants of one template. Fourth, it is \emph{information-complete, readable, and solver-safe}: the visuals contain the instance information needed for formulation, remain legible at benchmark scale, and avoid revealing solved decisions such as selected facilities, installed arcs, tours, schedules, or flows. The visual-design details behind these requirements are expanded in App.~\ref{subsec:visual_strategy}.

\begin{figure}[h]
    \centering
    \includegraphics[width=0.98\linewidth]{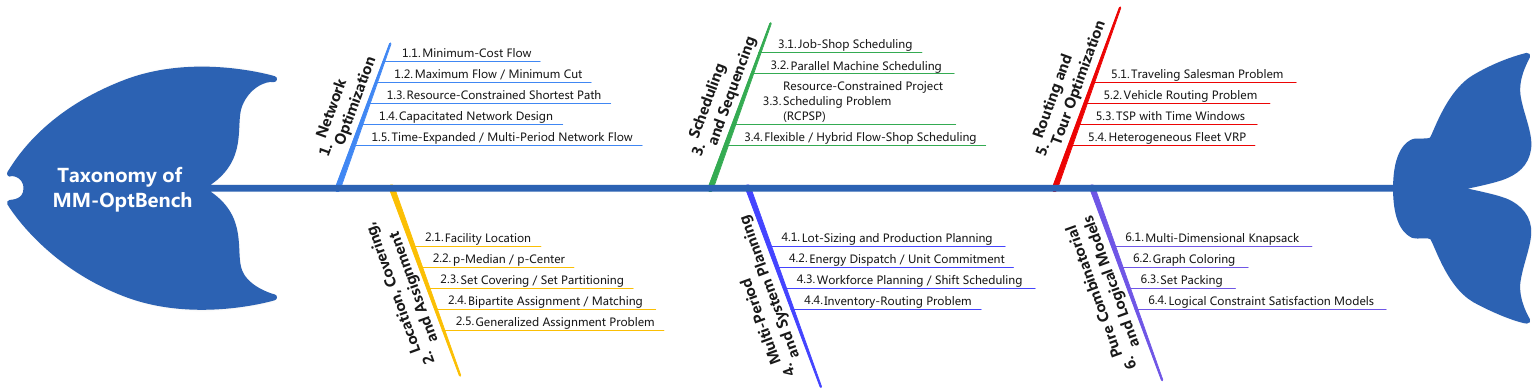}
    \caption{Taxonomy of MM-OptBench. The benchmark spans six major optimization families and 26 subcategories, organized by dominant mathematical structure rather than surface application domain.}
    \label{fig:mmoptbench_taxonomy}
\end{figure}

These requirements are instantiated through the taxonomy in Figure~\ref{fig:mmoptbench_taxonomy}. MM-OptBench is organized around six major optimization families defined primarily by \emph{mathematical structure} rather than surface application domain. These families expand into 26 subcategories, with the complete family/subcategory list and associated visual encodings reported in Appendix Table~\ref{tab:taxonomy_scope}. Each subcategory is instantiated at three difficulty levels (\emph{easy}, \emph{medium}, \emph{hard}) with 10 instances per level, yielding 780 solver-verified instances. The resulting coverage tests whether models can generalize across distinct formulation paradigms, including flow conservation, assignment and activation, temporal precedence, cross-period balance, global tour connectivity, and logical consistency.

The retained families and subcategories were selected by triangulating across optimization textbooks, course materials, open benchmark libraries, and open-source modeling repositories, with the source categories and family-level rationale detailed in App.~\ref{subsec:taxonomy_rationale}. We screened candidates along six axes: \emph{structural diversity}, so that the benchmark covers different constraint and objective patterns; \emph{canonicality}, so that tasks reflect established optimization models rather than custom puzzles; \emph{multimodal naturalness}, so that the visual artifact is a natural carrier of instance information; \emph{difficulty scalability}, so that each subcategory supports easy, medium, and hard regimes; \emph{solver verifiability}, so that reference solutions can be certified; and \emph{cross-family non-redundancy}, so that one family does not simply duplicate another under new names. The same logic determines the visual carriers described in App.~\ref{subsec:visual_strategy}: network problems use directed graph views; location, covering, and assignment problems use spatial, compatibility, or incidence views; scheduling problems use operation, precedence, stage, or resource views; multi-period planning uses timelines and planning dashboards; routing problems use map-like spatial views; and combinatorial/logical problems use tables and matrices.

\subsection{Solver-Grounded Construction and Validation}
\label{subsec:benchmark_pipeline}

\begin{figure}[h]
    \centering
    \includegraphics[width=1.0\linewidth]{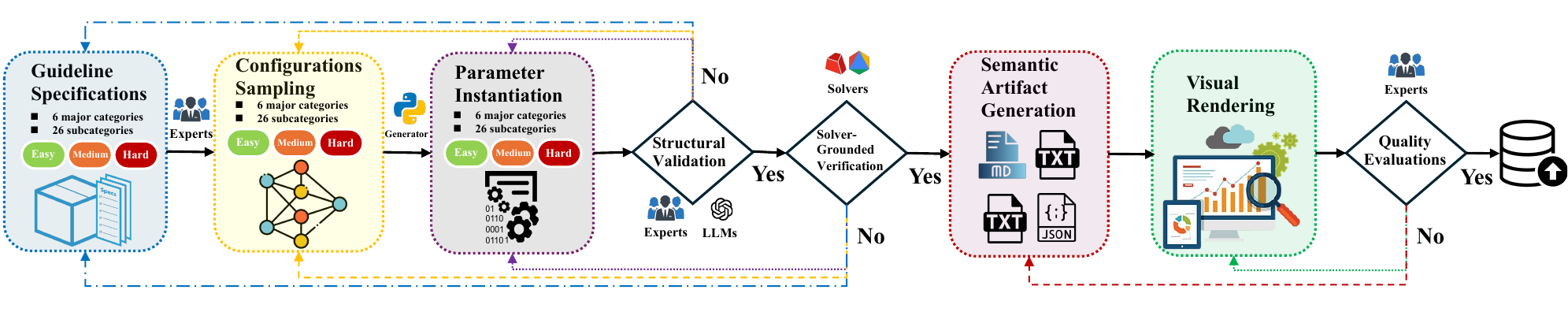}
    \caption{MM-OptBench construction and validation pipeline. Family-specific guidelines drive configuration sampling and parameter instantiation. Candidate instances are filtered by structural validation and solver-grounded verification, then converted into aligned semantic artifacts and rendered visual inputs. Final quality evaluation rejects inconsistent, unreadable, or solution-leaking cases before release.}
    \label{fig:benchmark_pipeline_overview}
\end{figure}

All instances follow the shared pipeline in Figure~\ref{fig:benchmark_pipeline_overview}, with the full stage-by-stage description in App.~\ref{subsec:appendix_pipeline}. Expert-written family guidelines first specify admissible scales, structural motifs, difficulty regimes, visual carriers, and readability constraints; App.~\ref{subsec:appendix_sampling} details how these guidelines become sampling and difficulty rules. The generator then samples a discrete configuration and instantiates numerical or categorical parameters. The sampled candidate enters structural validation, where experts, aided by LLM-assisted checks, inspect family semantics, nontriviality, dimensional consistency, and cross-field parameter bindings. Structurally valid candidates proceed to solver-grounded verification: a reference model or exact solving procedure certifies an optimal solution and records the benchmark objective and any family-specific solution object. Only after verification do we construct semantic artifacts, render visual inputs, and run final quality evaluations for readability, cross-modal consistency, artifact completeness, and solution leakage; solver/artifact alignment and quality-control details are given in Apps.~\ref{subsec:appendix_solver_templates} and~\ref{subsec:appendix_qa}.

The key construction invariant is \emph{single-source alignment}. Once a candidate is verified, its model-facing text, rendered visual artifact(s), canonical formulation, reference solver, solution record, and metadata are derived from the same structured instance record rather than authored independently. This prevents drift between text, visuals, formulation, and executable reference behavior, and makes each released package auditable. Representative family-specific pipelines are provided in Apps.~\ref{subsec:data_pipeline_network}--\ref{subsec:data_pipeline_combinatorial}; complete subcategory-level descriptions are provided in the online supplementary material.

\subsection{Artifact Contract and Difficulty Design}
\label{subsec:benchmark_artifacts}

Each released instance is organized as a two-sided package. The \emph{model-facing} side contains \texttt{task\_input.txt} and one or more rendered files under \texttt{visuals/}. The hidden \emph{reference} side contains the structured instance data, canonical formulation, executable reference solver, verified solution, and metadata used for auditing and scoring. This separation defines the evaluation contract: models receive only the public multimodal specification, while the benchmark retains the artifacts needed to verify execution, inspect formulations, and audit failures. App.~\ref{subsec:appendix_format} gives the file-level organization; Apps.~\ref{subsec:appendix_solver_templates} and~\ref{subsec:appendix_qa} describe reference-solver alignment and validation.

\begin{wrapfigure}{r}{0.4\linewidth}
\begingroup
\centering
\includegraphics[width=\linewidth]{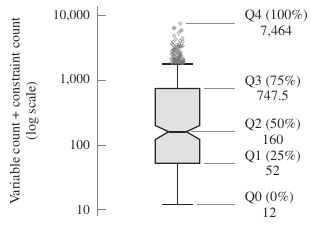}
\setlength{\abovecaptionskip}{2pt}
\setlength{\belowcaptionskip}{0pt}
\caption[Formulation-size distribution across 780 instances.]{\raggedright Distribution of formulation sizes across all 780 instances. Formulation size is measured by the total number of decision variables and constraints. The vertical axis uses a logarithmic scale. Q0--Q4 indicate the minimum, quartiles, and maximum (\mbox{median 160}; \mbox{mean 617}).}
\label{fig:formulation_size_boxplot}
\par\endgroup
\end{wrapfigure}

Difficulty is defined at the subcategory level through \emph{structural complexity}, rather than by a single global size rule. Depending on the family, moving from easy to hard may increase the number or dimensionality of decisions, coupling density, integrality structure, temporal or routing dependencies, visual density, or the amount of information that must be integrated across text and visuals. To complement these structural criteria, we characterize instance scale using formulation size, defined as the number of decision variables plus the number of constraints following MIPLIB-NL~\citep{li2026constructing}. Figure~\ref{fig:formulation_size_boxplot} shows the distribution across all 780 instances: sizes range from 12 to 7,464, with a median of 160. The distribution has a long upper tail: 274 instances (35.1\%) have size at least 500, 154 (19.7\%) at least 1,000, and 83 (10.6\%) at least 1,800. The full sampling and difficulty rules are described in App.~\ref{subsec:appendix_sampling}, and worked examples of the released artifact format appear in App.~\ref{subsec:appendix_examples}. Together, the artifact contract and difficulty design specify what models see, what the benchmark withholds for verification, and how structural complexity is varied across the 780 released instances.

\WFclear

\section{Experiments}
\label{sec:experiments}

Our experiments ask whether current MLLMs can turn MM-OptBench inputs into correct optimization formulations and executable solver artifacts, how robust this capability is across structures and difficulty levels, and where failures arise. We organize the study around four questions: \textbf{RQ1.} How capable are frontier general-purpose MLLMs at multimodal optimization modeling? \textbf{RQ2.} How well do MLLMs trained for mathematical reasoning perform on multimodal optimization modeling? \textbf{RQ3.} Do models perform consistently across optimization families, and how much does performance drop from easy to hard instances? \textbf{RQ4.} When models fail, do they mainly misread the text and visuals, or fail later in formulation, solving, and code?

\noindent\textbf{Experimental Setup.}
Our main benchmark comparison evaluates two model groups without MM-OptBench-specific fine-tuning, retrieval augmentation, or external agentic repair. The general-purpose group contains GPT-5.4, Gemini 3.1 Pro Preview, Claude Sonnet 4.6, Qwen3-VL-Plus, Qwen-VL-Max, and GLM-4.5V. The math-specialized group contains MathCoder-VL-8B~\citep{wang2025mathcoder}, MM-Eureka~\citep{meng2025mm}, and MM-PRM~\citep{du2025mm}. Each model receives \texttt{task\_input.txt} plus visual artifact(s), and returns a mathematical formulation and a solver-executable artifact. We report the Stage-1 metrics defined in Section~\ref{subsec:setting_protocol}: Valid Code Rate, pass@1, and pass@4, with five runs per model. Auxiliary experiments are in App.~\ref{subsec:app_controlled_evaluations}. The main text emphasizes Valid Code Rate and pass@1; pass@4 and full numeric tables are reported in Apps.~\ref{subsubsec:app_overall_metrics}--\ref{subsubsec:app_pass4_heatmap}. For RQ4, we use the Stage-2 and oracle-reading diagnostics from Section~\ref{subsec:setting_scoring_pipeline}; complete prompts, environment details, tolerances, and timeout analysis are in Apps.~\ref{subsec:app_exp_setup_details} and~\ref{subsec:app_runtime_analysis}. The dataset, reference solvers, evaluation code, and test prompts are publicly available under CC BY-SA 4.0 at \url{https://github.com/ZhongLIFR/MM-OptBench}.

\noindent\textbf{Section Organization.}
The rest of this section follows these questions directly. Section~\ref{subsec:exp_overall} addresses RQ1--RQ2 through overall pass@k, Valid Code Rate, and math-specialized transfer. Section~\ref{subsec:exp_family} addresses RQ3 with family-level and difficulty-level breakdowns. Section~\ref{subsec:exp_failures} addresses RQ4 through Stage-2 and oracle-reading failure attribution. Appendix~\ref{app:extended_experiment_results} provides the full setup (\S~\ref{subsec:app_exp_setup_details}), extended tables (\S\S~\ref{subsubsec:app_overall_metrics}--\ref{subsubsec:app_difficulty_results}), pass@4 heatmaps (\S~\ref{subsubsec:app_pass4_heatmap}), diagnostic counts (\S\S~\ref{subsec:app_general_purpose_failure_analysis}--\ref{subsec:app_math_specialized_failure_analysis}), runtime checks (\S~\ref{subsec:app_runtime_analysis}), and controlled repair and robustness evaluations (\S~\ref{subsec:app_controlled_evaluations}).

\subsection{Overall Capability and Executability Gap (RQ1 and RQ2)}
\label{subsec:exp_overall}

\begin{figure*}[h]
\centering
\includegraphics[width=1\textwidth]{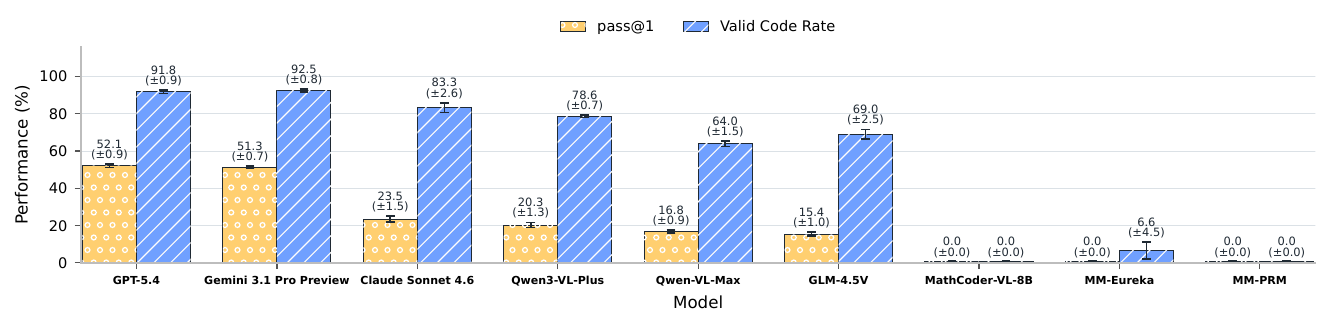}
\caption{Overall MM-OptBench performance for all nine evaluated MLLMs. For each model, the left bar reports pass@1 and the right bar reports Valid Code Rate, both as mean$\pm$standard deviation over five runs; pass@4 is reported in Appendix Table~\ref{tab:app_exp_overall}.}
\label{fig:exp_overall_summary}
\end{figure*}

Figure~\ref{fig:exp_overall_summary} summarizes pass@1 and Valid Code Rate for all nine evaluated models. The complete model-level results, including pass@4, are reported in Appendix Table~\ref{tab:app_exp_overall}. The paired bars compare solver-grounded correctness, measured by pass@1, with executability, measured by Valid Code Rate: a model may produce code that runs, yet still encode the wrong optimization problem.

Figure~\ref{fig:exp_overall_summary} reveals three model-level patterns. We first examine how much of the benchmark is solved by general-purpose MLLMs, then compare runnable code with solver-correct code, and finally test whether math-specialized MLLMs transfer to this setting.

\noindent\emph{(1) General-purpose MLLMs show emerging but limited capability.} GPT-5.4 reaches 52.1\% pass@1, narrowly ahead of Gemini 3.1 Pro Preview at 51.3\%; the other general-purpose models remain below 25\%. Averaged over the six general-purpose MLLMs, pass@1 is only 29.9\%, and pass@4 reaches 42.2\% (Appendix Table~\ref{tab:app_exp_overall}). This creates a clear two-model leading group, but even the strongest single-attempt results solve only about half of the benchmark. Multi-sample generation helps, yet the best pass@4 value, 68.3\% for GPT-5.4, still leaves a large unsolved fraction.

\noindent\emph{(2) Executability substantially overstates modeling correctness.} GPT-5.4 and Gemini produce runnable code on 91.8\% and 92.5\% of samples, respectively, but roughly half of the benchmark still fails solver-grounded verification. Similar gaps appear for all weaker systems, so Valid Code Rate is a weak proxy for correct optimization modeling.

\noindent\emph{(3) The three models trained for mathematical reasoning perform poorly on MM-OptBench.} MathCoder-VL-8B, MM-Eureka, and MM-PRM obtain 0.0\% pass@1 and 0.0\% pass@4; only MM-Eureka has nonzero Valid Code Rate, at 6.6\%. Oracle-reading does not rescue these models: after supplying the ground-truth public instance record, MathCoder-VL-8B solves $0/780$ diagnostic cases, MM-PRM solves $0/780$, and MM-Eureka solves $1/780$. Appendix~\ref{subsec:app_math_specialized_failure_analysis} shows that the failures extend beyond visual reading to response-format compliance, execution reliability, solver-code construction, and optimization-model synthesis; even MM-Eureka's Qwen2.5-VL-7B-Instruct base model solves $0/260$ easy oracle-reading cases. Their low scores cannot be attributed to training for mathematical reasoning alone; model size (7--8B parameters) and the need to write solver code may also affect performance.

\subsection{Structure and Difficulty Scaling (RQ3)}
\label{subsec:exp_family}
\label{subsec:exp_difficulty}

\begin{figure*}[h]
\centering
\includegraphics[width=0.98\textwidth]{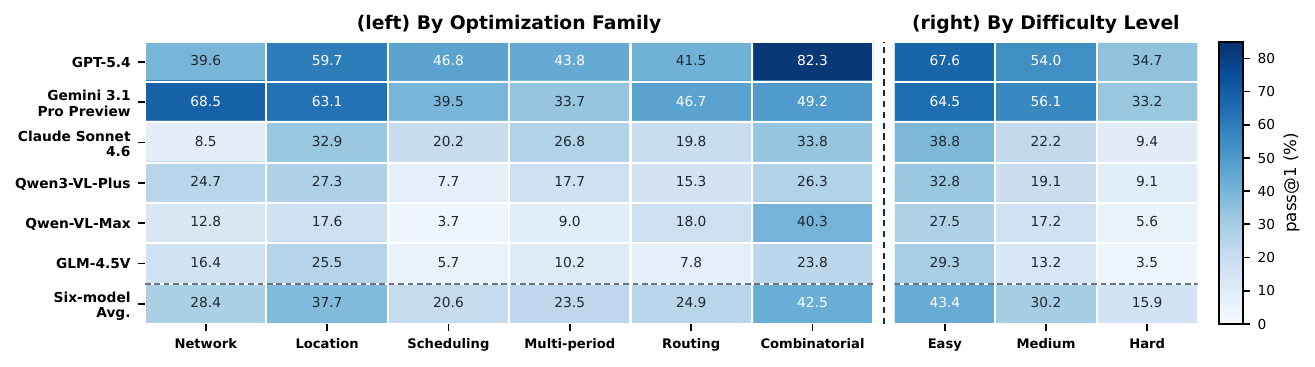}
\caption{pass@1 mean for the six general-purpose MLLMs across major optimization families (left) and released difficulty levels (right), using a shared model axis and color scale; the bottom row reports the average over these six models. Darker cells indicate higher solver-grounded correctness; the underlying family and difficulty values are reported in Appendix Tables~\ref{tab:app_exp_family} and~\ref{tab:app_exp_difficulty}, and the corresponding pass@4 heatmap is shown in Appendix Figure~\ref{fig:app_exp_family_heatmap_pass4}.}
\label{fig:exp_family_heatmap}
\end{figure*}

RQ3 asks whether performance generalizes across optimization structure and released difficulty tiers. Because the three math-specialized models obtain zero official success in every structural slice, Figure~\ref{fig:exp_family_heatmap} focuses on the six general-purpose MLLMs; aggregate rows average these six models only. 

\noindent\emph{(1) Performance is strongly family-dependent.} GPT-5.4 leads on Combinatorial / Logical Models (82.3\%) and Scheduling / Sequencing (46.8\%), while Gemini is strongest on Network Optimization (68.5\%), Location / Covering / Assignment (63.1\%), Multi-Period / System Planning (33.7\%), and Routing / Tour Optimization (46.7\%). No single model dominates every structural regime.

\noindent\emph{(2) The family averages reveal where the benchmark is hardest.} Combinatorial / Logical Models (42.5\%) and Location / Covering / Assignment (37.7\%) are the strongest families under the six-model general-purpose average, while Scheduling / Sequencing (20.6\%) and Multi-Period / System Planning (23.5\%) are weakest. pass@4 preserves the same qualitative ordering (Appendix~\ref{subsubsec:app_pass4_heatmap}), so repeated sampling improves absolute rates but does not remove family-specific bottlenecks.

\noindent\emph{(3) Difficulty produces a monotone degradation.} Averaged over the six general-purpose models, pass@1 falls from 43.4\% on easy instances to 30.2\% on medium and 15.9\% on hard. pass@4 follows the same trend, dropping from 59.4\% to 42.6\% and then 24.7\% (Appendix~\ref{subsubsec:app_pass4_heatmap}). Even the strongest models fall to roughly one-third pass@1 on hard instances, supporting the intended design that harder cases require more global consistency across indices, constraints, temporal coupling, resource interactions, and solver logic.

\subsection{Failure Attribution (RQ4)}
\label{subsec:exp_failures}

\begin{figure*}[h]
\centering
\includegraphics[width=0.98\textwidth]{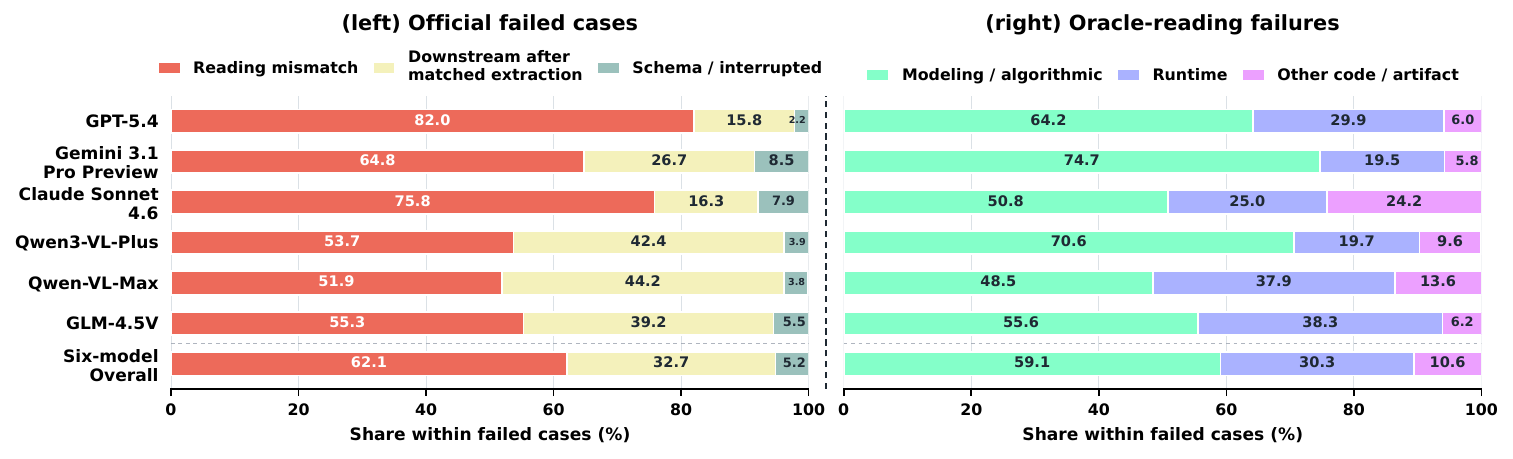}
\caption{Per-model post-hoc failure diagnostics on a typical run. The left panel shows, among official failures, the percentage assigned to each Stage-2 attribution category. The right panel shows, among oracle-reading failures, the percentage assigned to each downstream failure category. The bottom row pools failures over the six general-purpose models. Failure-category definitions are given in  Table~\ref{tab:failure_figure_label_definitions}.}
\label{fig:exp_failure_breakdown}
\end{figure*}

Finally, we analyze failure modes of evaluated MLLMs. Figure~\ref{fig:exp_failure_breakdown} uses a typical run and separates two diagnostic views. The left panel starts from \emph{official failed cases}: Stage-1 submissions whose solver artifact does not produce a correct solution. In Stage 2, we ask the model to extract the instance data from the same text and images without solving the problem, then compare the extracted data with the reference data. The right panel considers \emph{oracle-reading failures}: cases that still fail after the ground-truth public instance record is supplied directly. Figure-label definitions are given in Appendix Table~\ref{tab:failure_figure_label_definitions}, and exact left- and right-panel counts are reported in Tables~\ref{tab:app_exp_failure_breakdown_official} and~\ref{tab:app_exp_failure_breakdown_oracle}.

\noindent\emph{(1) Reading text-plus-visual instance data is a major bottleneck.} In the left panel, we examine 3,302 cases in which the six general-purpose models failed to solve the problem in Stage 1. When asked to extract the input data in Stage 2, the models return incorrect data in 62.1\% of these cases and data matching the reference in 32.7\%. The remaining 5.2\% are unresolved. GPT-5.4 has the largest reading-mismatch share among its failures (82.0\%), whereas Qwen3-VL-Plus and Qwen-VL-Max have larger downstream-after-matched-extraction shares (42.4\% and 44.2\%).

\noindent\emph{(2) Correctly reading the instance is still not enough.} In the right panel, under oracle-reading where the ground-truth public instance data are supplied exactly, failed cases are mostly modeling or algorithmic mismatches (59.1\%), followed by runtime failures (30.3\%) and other code/artifact failures (10.6\%). From the results in  Tables~\ref{tab:app_exp_failure_breakdown_official}-\ref{tab:app_exp_failure_breakdown_oracle}, the solved share increases from 29.4\% with the original text and images (Stage 1) to 51.7\% when the reference data are provided directly (oracle reading), but substantial downstream failures remain. This improvement suggests that reading information from text and images is an important source of difficulty.

\noindent\emph{(3) Models must read the input correctly and produce correct optimization models and solver code.} Models must first read the multimodal instance accurately, and then turn that instance into a globally coherent formulation and executable solving procedure. The math-specialized failure breakdown is reported separately in Appendix~\ref{subsec:app_math_specialized_failure_analysis}.

\section{Conclusion}
\label{sec:conclusion}

We introduced multimodal optimization modeling, a benchmark setting in which models translate textual and visual specifications into both mathematical formulations and executable solver code. MM-OptBench instantiates this setting with a solver-grounded construction framework and 780 verified instances spanning 6 optimization families, 26 subcategories, and three difficulty levels. Our evaluation shows that frontier MLLMs exhibit emerging capability but remain far from reliable: performance drops sharply with structural difficulty, executable code often fails solver-grounded correctness, and the tested math-specialized MLLMs obtain zero official solves. Models struggle to extract data from text and images. Even when given the correct input data, they still make errors in optimization modeling and solver code. These results suggest that multimodal optimization modeling is a stringent test of whether foundation models can move beyond local visual understanding toward formal optimization-model construction; limitations and broader impact are discussed in Appendix~\ref{app:limitations_impact}.

\section*{Acknowledgements and Disclosures}
The computational resources were supported by the Center for Intelligent Computing and Song-Shan Lake HPC Center (SSL-HPC) in Great Bay University, Dongguan, China. This work was supported in part by National Key Research and Development Program of China under the grant number 2025YFA1017800 and the XAIPre project funded by the Dutch Research Council (NWO) under grant number 19455. We also thank the anonymous reviewers for their valuable feedback. The authors declare no competing interests.

\begingroup
\color{black}
\bibliographystyle{ACM-Reference-Format}
\bibliography{References}

@inproceedings{li2026constructing,
  title={Constructing Industrial-Scale Optimization Modeling Benchmark},
  author={Li, Zhong and Lu, Hongliang and Wei, Tao and Chen, Yuxuan and Liu, Wenyu and Lan, Yuan and Zhang, Fan and Wen, Zaiwen},
  booktitle={Proceedings of the 43rd International Conference on Machine Learning},
  volume={306},
  series={Proceedings of Machine Learning Research},
  pages={68993--69036},
  year={2026},
  publisher={PMLR},
  url={https://proceedings.mlr.press/v306/li26dz.html}
}

@misc{lu2026pearl,
  title={{PEARL}: Solver-in-the-Loop Interactive Optimization Modeling from Natural Language},
  author={Lu, Hongliang and Li, Zhong and Chen, Yuxuan and Lan, Yuan and Zhang, Fan and Wen, Zaiwen},
  year={2026},
  eprint={2607.18256},
  archivePrefix={arXiv},
  primaryClass={cs.AI},
  url={https://arxiv.org/abs/2607.18256}
}

@misc{li2026right,
  title={Right Answers, Costly Models: The Efficiency Gap in {LLM}-based Optimization Modeling},
  author={Li, Zhong and Huang, Xin and Wan, Jinhui and Wang, Xiangyi and Zhang, Shenkai and Chen, Ruiqi and Liu, Wenyu and Wen, Zaiwen and Luo, Ziyan},
  year={2026},
  eprint={2609.38884},
  archivePrefix={arXiv},
  primaryClass={cs.LG},
  url={https://arxiv.org/abs/2609.38884}
}

@misc{liu2026localizable,
  title={Automated Optimization Modeling via a Localizable Error-Driven Perspective},
  author={Liu, Weiting and Wu, Han and Kuang, Yufei and Han, Xiongwei and Zhong, Tao and Feng, Jianfeng and Lu, Wenlian},
  year={2026},
  eprint={2602.11164},
  archivePrefix={arXiv},
  primaryClass={cs.LG},
  url={https://arxiv.org/abs/2602.11164}
}

@inproceedings{xiao2025survey,
  title={A survey of optimization modeling meets LLMs: progress and future directions},
  author={Xiao, Ziyang and Xie, Jingrong and Xu, Lilin and Guan, Shisi and Zhu, Jingyan and Han, Xiongwei and Fu, Xiaojin and Yu, WingYin and Wu, Han and Shi, Wei and others},
  booktitle={Proceedings of the Thirty-Fourth International Joint Conference on Artificial Intelligence},
  pages={10742--10750},
  year={2025}
}

@article{anand2017comparative,
  title={A comparative analysis of optimization solvers},
  author={Anand, Rimmi and Aggarwal, Divya and Kumar, Vijay},
  journal={Journal of Statistics and Management Systems},
  volume={20},
  number={4},
  pages={623--635},
  year={2017},
  publisher={Taylor \& Francis}
}

@book{antoniou2007practical,
  title={Practical optimization: algorithms and engineering applications},
  author={Antoniou, Andreas and Lu, Wu-Sheng},
  year={2007},
  publisher={Springer}
}

@article{huang2025orlm,
  title={Orlm: A customizable framework in training large models for automated optimization modeling},
  author={Huang, Chenyu and Tang, Zhengyang and Hu, Shixi and Jiang, Ruoqing and Zheng, Xin and Ge, Dongdong and Wang, Benyou and Wang, Zizhuo},
  journal={Operations Research},
  year={2025},
  publisher={INFORMS}
}

@inproceedings{jiang2025llmopt,
  title={LLMOPT: Learning to Define and Solve General Optimization Problems from Scratch},
  author={Jiang, Caigao and Shu, Xiang and Qian, Hong and Lu, Xingyu and ZHOU, JUN and Zhou, Aimin and Yu, Yang},
  booktitle={The Thirteenth International Conference on Learning Representations},
year ={2025}
}

@article{fan2025artificial,
  title={Artificial intelligence for optimization: Unleashing the potential of parameter generation, model formulation, and solution methods},
  author={Fan, Zhenan and Ghaddar, Bissan and Wang, Xinglu and Xing, Linzi and Zhang, Yong and Zhou, Zirui},
  journal={European Journal of Operational Research},
  year={2025},
  publisher={Elsevier}
}

@inproceedings{xiao2023chain,
  title={Chain-of-experts: When llms meet complex operations research problems},
  author={Xiao, Ziyang and Zhang, Dongxiang and Wu, Yangjun and Xu, Lilin and Wang, Yuan Jessica and Han, Xiongwei and Fu, Xiaojin and Zhong, Tao and Zeng, Jia and Song, Mingli and others},
  booktitle={The twelfth international conference on learning representations},
  year={2024}
}

@misc{ahmaditeshnizi2024opt,
      title={OptiMUS: Scalable Optimization Modeling with (MI)LP Solvers and Large Language Models}, 
      author={Ali AhmadiTeshnizi and Wenzhi Gao and Madeleine Udell},
      year={2024},
      eprint={2402.10172},
      archivePrefix={arXiv},
      primaryClass={cs.AI},
      url={https://arxiv.org/abs/2402.10172}, 
}

@inproceedings{lu2025optmath,
  title={OptMATH: A Scalable Bidirectional Data Synthesis Framework for Optimization Modeling},
  author={Lu, Hongliang and Xie, Zhonglin and Wu, Yaoyu and Ren, Can and Chen, Yuxuan and Wen, Zaiwen},
  booktitle={Forty-second International Conference on Machine Learning},
year={2025}
}

@InProceedings{Ramamonjison2023NL4Opt,
  title = 	 {NL4Opt Competition: Formulating Optimization Problems Based on Their Natural Language Descriptions},
  author =       {Ramamonjison, Rindranirina and Yu, Timothy and Li, Raymond and Li, Haley and Carenini, Giuseppe and Ghaddar, Bissan and He, Shiqi and Mostajabdaveh, Mahdi and Banitalebi-Dehkordi, Amin and Zhou, Zirui and Zhang, Yong},
  booktitle = 	 {Proceedings of the NeurIPS 2022 Competitions Track},
  pages = 	 {189--203},
  year = 	 {2022},
  editor = 	 {Ciccone, Marco and Stolovitzky, Gustavo and Albrecht, Jacob},
  volume = 	 {220},
  series = 	 {Proceedings of Machine Learning Research},
  month = 	 {28 Nov--09 Dec},
  publisher =    {PMLR},
  url = 	 {https://proceedings.mlr.press/v220/ramamonjison23a.html},
}

@article{huang2024mamo,
  title={Mamo: a mathematical modeling benchmark with solvers},
  author={Huang, Xuhan and Shen, Qingning and Hu, Yan and Gao, Anningzhe and Wang, Benyou},
  journal={arXiv preprints arXiv:2405.13144},
  year={2024}
}

@article{yang2025automated,
  title={Automated Optimization Modeling through Expert-Guided Large Language Model Reasoning},
  author={Yang, Beinuo and Zhou, Qishen and Li, Junyi and Su, Chenxing and Hu, Simon},
  journal={arXiv preprint arXiv:2508.14410},
  year={2025}
}

@article{wang2025orgeval,
  title={ORGEval: Graph-Theoretic Evaluation of LLMs in Optimization Modeling},
  author={Wang, Zhuohan and Zhu, Ziwei and Li, Ziniu and Chen, Congliang and Han, Yizhou and Lin, Yufeng and Lin, Zhihang and Gu, Angyang and Hu, Xinglin and Sun, Ruoyu and  Tian, Ding},
  journal={arXiv preprint arXiv:2510.27610},
  year={2025}
}

@inproceedings{yang2024optibench,
  title={OptiBench Meets ReSocratic: Measure and Improve LLMs for Optimization Modeling},
  author={Yang, Zhicheng and Wang, Yiwei and Huang, Yinya and Guo, Zhijiang and Shi, Wei and Han, Xiongwei and Feng, Liang and Song, Linqi and Liang, Xiaodan and Tang, Jing},
  booktitle={The Thirteenth International Conference on Learning Representations},
  year ={2025},
  url={https://openreview.net/forum?id=fsDZwS49uY}
}

@inproceedings{xie2025murka,
  title={MURKA: Multi-Reward Reinforcement Learning with Knowledge Alignment for Optimization Tasks},
  author={Xie, Wantong and Hu, Yi-Xiang and Xu, Jieyang and Wu, Feng and Li, Xiangyang},
  booktitle={The Thirty-ninth Annual Conference on Neural Information Processing Systems},
  year ={2025},
  url={https://openreview.net/forum?id=f4pvPNf9ox}
}

@article{chen2025solver,
  title={Solver-Informed RL: Grounding Large Language Models for Authentic Optimization Modeling},
  author={Chen, Yitian and Xia, Jingfan and Shao, Siyu and Ge, Dongdong and Ye, Yinyu},
  journal={arXiv preprint arXiv:2505.11792},
  year={2025}
}

@article{wang2025large,
  title={Large Language Models in Operations Research: Methods, Applications, and Challenges},
  author={Wang, Yang and Li, Kai},
  journal={arXiv preprint arXiv:2509.18180},
  year={2025}
}

@article{singh2012overview,
  title={An overview of the optimization modelling applications},
  author={Singh, Ajay},
  journal={Journal of Hydrology},
  volume={466},
  pages={167--182},
  year={2012},
  publisher={Elsevier}
}

@article{tsouros2023holy,
  title={Holy grail 2.0: From natural language to constraint models},
  author={Tsouros, Dimos and Verhaeghe, H{\'e}l{\`e}ne and Kad{\i}o{\u{g}}lu, Serdar and Guns, Tias},
  journal={arXiv preprint arXiv:2308.01589},
  year={2023}
}

@inproceedings{freuder2024conversational,
  title={Conversational modeling for constraint satisfaction},
  author={Freuder, Eugene C},
  booktitle={Proceedings of the AAAI Conference on Artificial Intelligence},
  volume={38},
  number={20},
  pages={22592--22597},
  year={2024}
}

@inproceedings{zhang2024geoeval,
  title={Geoeval: benchmark for evaluating llms and multi-modal models on geometry problem-solving},
  author={Zhang, Jiaxin and Li, Zhong-Zhi and Zhang, Ming-Liang and Yin, Fei and Liu, Cheng-Lin and Moshfeghi, Yashar},
  booktitle={Findings of the Association for Computational Linguistics: ACL 2024},
  pages={1258--1276},
  year={2024}
}

@inproceedings{lu2021inter,
  title={Inter-gps: Interpretable geometry problem solving with formal language and symbolic reasoning},
  author={Lu, Pan and Gong, Ran and Jiang, Shibiao and Qiu, Liang and Huang, Siyuan and Liang, Xiaodan and Zhu, Song-Chun},
  booktitle={Proceedings of the 59th Annual Meeting of the Association for Computational Linguistics and the 11th International Joint Conference on Natural Language Processing (Volume 1: Long Papers)},
  pages={6774--6786},
  year={2021}
}

@inproceedings{seo2015solving,
  title={Solving geometry problems: Combining text and diagram interpretation},
  author={Seo, Minjoon and Hajishirzi, Hannaneh and Farhadi, Ali and Etzioni, Oren and Malcolm, Clint},
  booktitle={Proceedings of the 2015 conference on empirical methods in natural language processing},
  pages={1466--1476},
  year={2015}
}

@inproceedings{chen2021geoqa,
  title={Geoqa: A geometric question answering benchmark towards multimodal numerical reasoning},
  author={Chen, Jiaqi and Tang, Jianheng and Qin, Jinghui and Liang, Xiaodan and Liu, Lingbo and Xing, Eric and Lin, Liang},
  booktitle={Findings of the Association for Computational Linguistics: ACL-IJCNLP 2021},
  pages={513--523},
  year={2021}
}

@inproceedings{
lu2024mathvista,
title={MathVista: Evaluating Mathematical Reasoning of Foundation Models in Visual Contexts},
author={Pan Lu and Hritik Bansal and Tony Xia and Jiacheng Liu and Chunyuan Li and Hannaneh Hajishirzi and Hao Cheng and Kai-Wei Chang and Michel Galley and Jianfeng Gao},
booktitle={The Twelfth International Conference on Learning Representations},
year={2024},
url={https://openreview.net/forum?id=KUNzEQMWU7}
}

@inproceedings{he2024olympiadbench,
  title={Olympiadbench: A challenging benchmark for promoting agi with olympiad-level bilingual multimodal scientific problems},
  author={He, Chaoqun and Luo, Renjie and Bai, Yuzhuo and Hu, Shengding and Thai, Zhen and Shen, Junhao and Hu, Jinyi and Han, Xu and Huang, Yujie and Zhang, Yuxiang and others},
  booktitle={Proceedings of the 62nd Annual Meeting of the Association for Computational Linguistics (Volume 1: Long Papers)},
  pages={3828--3850},
  year={2024}
}

@inproceedings{wang2024scibench,
  title={SCIBENCH: evaluating college-level scientific problem-solving abilities of large language models},
  author={Wang, Xiaoxuan and Hu, Ziniu and Lu, Pan and Zhu, Yanqiao and Zhang, Jieyu and Subramaniam, Satyen and Loomba, Arjun R and Zhang, Shichang and Sun, Yizhou and Wang, Wei},
  booktitle={Proceedings of the 41st International Conference on Machine Learning},
  pages={50622--50649},
  year={2024}
}

@inproceedings{yue2024mmmu,
  title={Mmmu: A massive multi-discipline multimodal understanding and reasoning benchmark for expert agi},
  author={Yue, Xiang and Ni, Yuansheng and Zhang, Kai and Zheng, Tianyu and Liu, Ruoqi and Zhang, Ge and Stevens, Samuel and Jiang, Dongfu and Ren, Weiming and Sun, Yuxuan and others},
  booktitle={Proceedings of the IEEE/CVF conference on computer vision and pattern recognition},
  pages={9556--9567},
  year={2024}
}

@article{zhang2024cmmmu,
  title={Cmmmu: A chinese massive multi-discipline multimodal understanding benchmark},
  author={Zhang, Ge and Du, Xinrun and Chen, Bei and Liang, Yiming and Luo, Tongxu and Zheng, Tianyu and Zhu, Kang and Cheng, Yuyang and Xu, Chunpu and Guo, Shuyue and others},
  journal={arXiv preprint arXiv:2401.11944},
  year={2024}
}

@article{zhang2023m3exam,
  title={M3exam: A multilingual, multimodal, multilevel benchmark for examining large language models},
  author={Zhang, Wenxuan and Aljunied, Mahani and Gao, Chang and Chia, Yew Ken and Bing, Lidong},
  journal={Advances in Neural Information Processing Systems},
  volume={36},
  pages={5484--5505},
  year={2023}
}

@inproceedings{zhang2024mathverse,
  title={Mathverse: Does your multi-modal llm truly see the diagrams in visual math problems?},
  author={Zhang, Renrui and Jiang, Dongzhi and Zhang, Yichi and Lin, Haokun and Guo, Ziyu and Qiu, Pengshuo and Zhou, Aojun and Lu, Pan and Chang, Kai-Wei and Qiao, Yu and others},
  booktitle={European Conference on Computer Vision},
  pages={169--186},
  year={2024},
  organization={Springer}
}

@article{wang2024measuring,
  title={Measuring multimodal mathematical reasoning with math-vision dataset},
  author={Wang, Ke and Pan, Junting and Shi, Weikang and Lu, Zimu and Ren, Houxing and Zhou, Aojun and Zhan, Mingjie and Li, Hongsheng},
  journal={Advances in Neural Information Processing Systems},
  volume={37},
  pages={95095--95169},
  year={2024}
}

@inproceedings{kembhavi2016diagram,
  title={A diagram is worth a dozen images},
  author={Kembhavi, Aniruddha and Salvato, Mike and Kolve, Eric and Seo, Minjoon and Hajishirzi, Hannaneh and Farhadi, Ali},
  booktitle={European conference on computer vision},
  pages={235--251},
  year={2016},
  organization={Springer}
}

@inproceedings{
lu2021iconqa,
title={Icon{QA}: A New Benchmark for Abstract Diagram Understanding and Visual Language Reasoning},
author={Pan Lu and Liang Qiu and Jiaqi Chen and Tony Xia and Yizhou Zhao and Wei Zhang and Zhou Yu and Xiaodan Liang and Song-Chun Zhu},
booktitle={Thirty-fifth Conference on Neural Information Processing Systems Datasets and Benchmarks Track (Round 2)},
year={2021},
url={https://openreview.net/forum?id=uXa9oBDZ9V1}
}

@inproceedings{kembhavi2017you,
  title={Are you smarter than a sixth grader? textbook question answering for multimodal machine comprehension},
  author={Kembhavi, Aniruddha and Seo, Minjoon and Schwenk, Dustin and Choi, Jonghyun and Farhadi, Ali and Hajishirzi, Hannaneh},
  booktitle={Proceedings of the IEEE Conference on Computer Vision and Pattern recognition},
  pages={4999--5007},
  year={2017}
}

@article{lu2022learn,
  title={Learn to explain: Multimodal reasoning via thought chains for science question answering},
  author={Lu, Pan and Mishra, Swaroop and Xia, Tanglin and Qiu, Liang and Chang, Kai-Wei and Zhu, Song-Chun and Tafjord, Oyvind and Clark, Peter and Kalyan, Ashwin},
  journal={Advances in neural information processing systems},
  volume={35},
  pages={2507--2521},
  year={2022}
}

@inproceedings{masry2022chartqa,
  title={Chartqa: A benchmark for question answering about charts with visual and logical reasoning},
  author={Masry, Ahmed and Do, Xuan Long and Tan, Jia Qing and Joty, Shafiq and Hoque, Enamul},
  booktitle={Findings of the association for computational linguistics: ACL 2022},
  pages={2263--2279},
  year={2022}
}

@article{xu2023chartbench,
  title={Chartbench: A benchmark for complex visual reasoning in charts},
  author={Xu, Zhengzhuo and Du, Sinan and Qi, Yiyan and Xu, Chengjin and Yuan, Chun and Guo, Jian},
  journal={arXiv preprint arXiv:2312.15915},
  year={2023}
}

@inproceedings{mathew2021docvqa,
  title={Docvqa: A dataset for vqa on document images},
  author={Mathew, Minesh and Karatzas, Dimosthenis and Jawahar, CV},
  booktitle={Proceedings of the IEEE/CVF winter conference on applications of computer vision},
  pages={2200--2209},
  year={2021}
}

@inproceedings{sun2024mm,
  title={Mm-math: Advancing multimodal math evaluation with process evaluation and fine-grained classification},
  author={Sun, Kai and Bai, Yushi and Qi, Ji and Hou, Lei and Li, Juanzi},
  booktitle={Findings of the Association for Computational Linguistics: EMNLP 2024},
  pages={1358--1375},
  year={2024}
}

@article{yang2024chartmimic,
  title={Chartmimic: Evaluating lmm's cross-modal reasoning capability via chart-to-code generation},
  author={Yang, Cheng and Shi, Chufan and Liu, Yaxin and Shui, Bo and Wang, Junjie and Jing, Mohan and Xu, Linran and Zhu, Xinyu and Li, Siheng and Zhang, Yuxiang and others},
  journal={arXiv preprint arXiv:2406.09961},
  year={2024}
}

@inproceedings{wu2025plot2code,
  title={Plot2code: A comprehensive benchmark for evaluating multi-modal large language models in code generation from scientific plots},
  author={Wu, Chengyue and Liang, Zhixuan and Ge, Yixiao and Guo, Qiushan and Lu, Zeyu and Wang, Jiahao and Shan, Ying and Luo, Ping},
  booktitle={Findings of the Association for Computational Linguistics: NAACL 2025},
  pages={3006--3028},
  year={2025}
}

@inproceedings{qiao2025we,
  title={We-math: Does your large multimodal model achieve human-like mathematical reasoning?},
  author={Qiao, Runqi and Tan, Qiuna and Dong, Guanting and MinhuiWu, MinhuiWu and Sun, Chong and Song, Xiaoshuai and Wang, Jiapeng and Gongque, Zhuoma and Lei, Shanglin and Zhang, Yifan and others},
  booktitle={Proceedings of the 63rd Annual Meeting of the Association for Computational Linguistics (Volume 1: Long Papers)},
  pages={20023--20070},
  year={2025}
}

@inproceedings{wu2025tablebench,
  title={Tablebench: A comprehensive and complex benchmark for table question answering},
  author={Wu, Xianjie and Yang, Jian and Chai, Linzheng and Zhang, Ge and Liu, Jiaheng and Du, Xeron and Liang, Di and Shu, Daixin and Cheng, Xianfu and Sun, Tianzhen and others},
  booktitle={Proceedings of the AAAI Conference on Artificial Intelligence},
  volume={39},
  number={24},
  pages={25497--25506},
  year={2025}
}

@inproceedings{roberts2025grab,
  title={GRAB: A challenging graph analysis benchmark for large multimodal models},
  author={Roberts, Jonathan and Han, Kai and Albanie, Samuel},
  booktitle={Proceedings of the IEEE/CVF International Conference on Computer Vision},
  pages={1644--1654},
  year={2025}
}

@inproceedings{liu2025cmm,
  title={Cmm-math: A chinese multimodal math dataset to evaluate and enhance the mathematics reasoning of large multimodal models},
  author={Liu, Wentao and Pan, Qianjun and Zhang, Yi and Liu, Zhuo and Wu, Ji and Zhou, Jie and Zhou, Aimin and Chen, Qin and Jiang, Bo and He, Liang},
  booktitle={Proceedings of the 33rd ACM International Conference on Multimedia},
  pages={12585--12591},
  year={2025}
}

@inproceedings{
xia2025mmie,
title={{MMIE}: Massive Multimodal Interleaved Comprehension Benchmark for Large Vision-Language Models},
author={Peng Xia and Siwei Han and Shi Qiu and Yiyang Zhou and Zhaoyang Wang and Wenhao Zheng and Zhaorun Chen and Chenhang Cui and Mingyu Ding and Linjie Li and Lijuan Wang and Huaxiu Yao},
booktitle={The Thirteenth International Conference on Learning Representations},
year={2025},
url={https://openreview.net/forum?id=HnhNRrLPwm}
}

@inproceedings{li2025sketch2code,
  title={Sketch2code: Evaluating vision-language models for interactive web design prototyping},
  author={Li, Ryan and Zhang, Yanzhe and Yang, Diyi},
  booktitle={Proceedings of the 2025 Conference of the Nations of the Americas Chapter of the Association for Computational Linguistics: Human Language Technologies (Volume 1: Long Papers)},
  pages={3921--3955},
  year={2025}
}

@article{sun2025fullfront,
  title={FullFront: Benchmarking MLLMs Across the Full Front-End Engineering Workflow},
  author={Sun, Haoyu and Wang, Huichen Will and Gu, Jiawei and Li, Linjie and Cheng, Yu},
  journal={arXiv preprint arXiv:2505.17399},
  year={2025}
}

@inproceedings{wang2025sciver,
  title={SciVer: Evaluating Foundation Models for Multimodal Scientific Claim Verification},
  author={Wang, Chengye and Shen, Yifei and Kuang, Zexi and Cohan, Arman and Zhao, Yilun},
  booktitle={Proceedings of the 63rd Annual Meeting of the Association for Computational Linguistics (Volume 1: Long Papers)},
  pages={8562--8579},
  year={2025}
}

@article{liu2024comet,
  title={COMET:“cone of experience” enhanced large multimodal model for mathematical problem generation},
  author={Liu, Sannyuya and Feng, Jintian and Yang, Zongkai and Luo, Yawei and Wan, Qian and Shen, Xiaoxuan and Sun, Jianwen},
  journal={Science China Information Sciences},
  volume={67},
  number={12},
  pages={220108},
  year={2024},
  publisher={Springer}
}

@article{pramanick2024spiqa,
  title={Spiqa: A dataset for multimodal question answering on scientific papers},
  author={Pramanick, Shraman and Chellappa, Rama and Venugopalan, Subhashini},
  journal={Advances in Neural Information Processing Systems},
  volume={37},
  pages={118807--118833},
  year={2024}
}

@inproceedings{tanaka2024instructdoc,
  title={Instructdoc: A dataset for zero-shot generalization of visual document understanding with instructions},
  author={Tanaka, Ryota and Iki, Taichi and Nishida, Kyosuke and Saito, Kuniko and Suzuki, Jun},
  booktitle={Proceedings of the AAAI conference on artificial intelligence},
  volume={38},
  number={17},
  pages={19071--19079},
  year={2024}
}

@article{xia2024docgenome,
  title={Docgenome: An open large-scale scientific document benchmark for training and testing multi-modal large language models},
  author={Xia, Renqiu and Mao, Song and Yan, Xiangchao and Zhou, Hongbin and Zhang, Bo and Peng, Haoyang and Pi, Jiahao and Fu, Daocheng and Wu, Wenjie and Ye, Hancheng and others},
  journal={arXiv preprint arXiv:2406.11633},
  year={2024}
}

@inproceedings{li2024multimodal,
  title={Multimodal arxiv: A dataset for improving scientific comprehension of large vision-language models},
  author={Li, Lei and Wang, Yuqi and Xu, Runxin and Wang, Peiyi and Feng, Xiachong and Kong, Lingpeng and Liu, Qi},
  booktitle={Proceedings of the 62nd Annual Meeting of the Association for Computational Linguistics (Volume 1: Long Papers)},
  pages={14369--14387},
  year={2024}
}

@article{kahou2017figureqa,
  title={Figureqa: An annotated figure dataset for visual reasoning},
  author={Kahou, Samira Ebrahimi and Michalski, Vincent and Atkinson, Adam and K{\'a}d{\'a}r, {\'A}kos and Trischler, Adam and Bengio, Yoshua},
  journal={arXiv preprint arXiv:1710.07300},
  year={2017}
}

@inproceedings{kafle2018dvqa,
  title={Dvqa: Understanding data visualizations via question answering},
  author={Kafle, Kushal and Price, Brian and Cohen, Scott and Kanan, Christopher},
  booktitle={Proceedings of the IEEE conference on computer vision and pattern recognition},
  pages={5648--5656},
  year={2018}
}

@inproceedings{hsu2021scicap,
  title={SciCap: Generating captions for scientific figures},
  author={Hsu, Ting-Yao and Giles, C Lee and Huang, Ting-Hao},
  booktitle={Findings of the Association for Computational Linguistics: EMNLP 2021},
  pages={3258--3264},
  year={2021}
}

@inproceedings{masry2023unichart,
  title={Unichart: A universal vision-language pretrained model for chart comprehension and reasoning},
  author={Masry, Ahmed and Kavehzadeh, Parsa and Do, Xuan Long and Hoque, Enamul and Joty, Shafiq},
  booktitle={Proceedings of the 2023 conference on empirical methods in natural language processing},
  pages={14662--14684},
  year={2023}
}

@article{chang2022mapqa,
  title={Mapqa: A dataset for question answering on choropleth maps},
  author={Chang, Shuaichen and Palzer, David and Li, Jialin and Fosler-Lussier, Eric and Xiao, Ningchuan},
  journal={arXiv preprint arXiv:2211.08545},
  year={2022}
}

@article{lu2022dynamic,
  title={Dynamic prompt learning via policy gradient for semi-structured mathematical reasoning},
  author={Lu, Pan and Qiu, Liang and Chang, Kai-Wei and Wu, Ying Nian and Zhu, Song-Chun and Rajpurohit, Tanmay and Clark, Peter and Kalyan, Ashwin},
  journal={The Eleventh International Conference on Learning Representations},
  year={2022}
}

@article{wang2024pin,
  title={Pin: A knowledge-intensive dataset for paired and interleaved multimodal documents},
  author={Wang, Junjie and Zhang, Yuxiang and Liu, Minghao and Zhang, Yin and Ji, Yatai and Xuan, Weihao and Lin, Nie and Zhu, Kang and Lin, Zhiqiang and Ren, Yiming and others},
  journal={arXiv preprint arXiv:2406.13923},
  year={2024}
}

@inproceedings{shi2024math,
  title={Math-llava: Bootstrapping mathematical reasoning for multimodal large language models},
  author={Shi, Wenhao and Hu, Zhiqiang and Bin, Yi and Liu, Junhua and Yang, Yang and Ng, See Kiong and Bing, Lidong and Lee, Roy Ka-Wei},
  booktitle={Findings of the Association for Computational Linguistics: EMNLP 2024},
  pages={4663--4680},
  year={2024}
}

@inproceedings{li2024mmsci,
  title={Mmsci: A multimodal multi-discipline dataset for phd-level scientific comprehension},
  author={Li, Zekun and Yang, Xianjun and Choi, Kyuri and Zhu, Wanrong and Hsieh, Ryan and Kim, HyeonJung and Lim, Jin Hyuk and Ji, Sungyoung and Lee, Byungju and Yan, Xifeng and others},
  booktitle={AI for Accelerated Materials Design-Vienna 2024},
  year={2024}
}

@article{zhang2024mavis,
  title={Mavis: Mathematical visual instruction tuning with an automatic data engine},
  author={Zhang, Renrui and Wei, Xinyu and Jiang, Dongzhi and Guo, Ziyu and Li, Shicheng and Zhang, Yichi and Tong, Chengzhuo and Liu, Jiaming and Zhou, Aojun and Wei, Bin and others},
  journal={arXiv preprint arXiv:2407.08739},
  year={2024}
}

@article{zeng2024advancing,
  title={Advancing multimodal large language models in chart question answering with visualization-referenced instruction tuning},
  author={Zeng, Xingchen and Lin, Haichuan and Ye, Yilin and Zeng, Wei},
  journal={IEEE Transactions on Visualization and Computer Graphics},
  volume={31},
  number={1},
  pages={525--535},
  year={2024},
  publisher={IEEE}
}

@inproceedings{cai2024geogpt4v,
  title={Geogpt4v: Towards geometric multi-modal large language models with geometric image generation},
  author={Cai, Shihao and Bao, Keqin and Guo, Hangyu and Zhang, Jizhi and Song, Jun and Zheng, Bo},
  booktitle={Proceedings of the 2024 Conference on Empirical Methods in Natural Language Processing},
  pages={750--766},
  year={2024}
}

@inproceedings{han2409infimm,
  title={{InfiMM-WebMath-40B}: Advancing Multimodal Pre-Training for Enhanced Mathematical Reasoning},
  author={Han, Xiaotian and Jian, Yiren and Hu, Xuefeng and Liu, Haogeng and Wang, Yiqi and Fan, Qihang and Ai, Yuang and Huang, Huaibo and He, Ran and Yang, Zhenheng and You, Quanzeng},
  booktitle={Findings of the Association for Computational Linguistics: EMNLP 2025},
  year={2025},
  publisher={Association for Computational Linguistics},
  address={Suzhou, China},
  pages={14221--14231},
  doi={10.18653/v1/2025.findings-emnlp.766},
  url={https://aclanthology.org/2025.findings-emnlp.766/}
}

@book{hart2017pyomo,
  title={{Pyomo}---Optimization Modeling in {Python}},
  author={Hart, William E. and Laird, Carl D. and Watson, Jean-Paul and Woodruff, David L. and Hackebeil, Gabriel A. and Nicholson, Bethany L. and Siirola, John D.},
  edition={2},
  series={Springer Optimization and Its Applications},
  volume={67},
  year={2017},
  publisher={Springer},
  address={Cham},
  doi={10.1007/978-3-319-58821-6}
}

@article{peng2024multimath,
  title={Multimath: Bridging visual and mathematical reasoning for large language models},
  author={Peng, Shuai and Fu, Di and Gao, Liangcai and Zhong, Xiuqin and Fu, Hongguang and Tang, Zhi},
  journal={arXiv preprint arXiv:2409.00147},
  year={2024}
}

@article{yang2024mathglm,
  title={Mathglm-vision: solving mathematical problems with multi-modal large language model},
  author={Yang, Zhen and Chen, Jinhao and Du, Zhengxiao and Yu, Wenmeng and Wang, Weihan and Hong, Wenyi and Jiang, Zhihuan and Xu, Bin and Tang, Jie},
  journal={arXiv preprint arXiv:2409.13729},
  year={2024}
}

@inproceedings{wang2025mathcoder,
  title={Mathcoder-vl: Bridging vision and code for enhanced multimodal mathematical reasoning},
  author={Wang, Ke and Pan, Junting and Wei, Linda and Zhou, Aojun and Shi, Weikang and Lu, Zimu and Xiao, Han and Yang, Yunqiao and Ren, Houxing and Zhan, Mingjie and others},
  booktitle={Findings of the Association for Computational Linguistics: ACL 2025},
  pages={2505--2534},
  year={2025}
}

@article{meng2025mm,
  title={Mm-eureka: Exploring the frontiers of multimodal reasoning with rule-based reinforcement learning},
  author={Meng, Fanqing and Du, Lingxiao and Liu, Zongkai and Zhou, Zhixiang and Lu, Quanfeng and Fu, Daocheng and Han, Tiancheng and Shi, Botian and Wang, Wenhai and He, Junjun and others},
  journal={arXiv preprint arXiv:2503.07365},
  year={2025}
}

@article{du2025mm,
  title={Mm-prm: Enhancing multimodal mathematical reasoning with scalable step-level supervision},
  author={Du, Lingxiao and Meng, Fanqing and Liu, Zongkai and Zhou, Zhixiang and Luo, Ping and Zhang, Qiaosheng and Shao, Wenqi},
  journal={arXiv preprint arXiv:2505.13427},
  year={2025}
}

@article{li2025zebra,
  title={Zebra-cot: A dataset for interleaved vision language reasoning},
  author={Li, Ang and Wang, Charles and Fu, Deqing and Yue, Kaiyu and Cai, Zikui and Zhu, Wang Bill and Liu, Ollie and Guo, Peng and Neiswanger, Willie and Huang, Furong and others},
  journal={The Fourteenth International Conference on Learning Representations},
  year={2025}
}

@misc{li2026optargus,
  title={{OptArgus}: A Multi-Agent System to Detect Hallucinations in {LLM}-based Optimization Modeling},
  author={Li, Zhong and Guo, Zihan and Lu, Xiaohan and Wang, Juntao and Song, Jie and Shen, Chao and Wu, Jiageng and Sun, Mingyang},
  year={2026},
  eprint={2605.11738},
  archivePrefix={arXiv},
  url={https://arxiv.org/abs/2605.11738}
}

@misc{wu2026fix,
  title={Fix What's Wrong, Keep What's Right: Hallucination Repair for {LLM}-Based Optimization Modeling},
  author={Wu, Jiageng and Shen, Chao and Guo, Zihan and Shao, Junyan and Zhang, Zongyan and Song, Jie and Li, Zhong and Sun, Mingyang},
  year={2026},
  eprint={2610.12529},
  archivePrefix={arXiv},
  url={https://arxiv.org/abs/2610.12529}
}

@misc{yang2026modularized,
  title={Modularized Reinforcement Learning on {LLMs}: From {MDP} Creation to Exploration and Learning},
  author={Yang, Zhao and Jiang, Yuxuan and Chen, Ting-Chih and Yang, Lincen and Wong, Annie and Gao, Chao and Kooi, Jacob E. and Li, Zhong and Shi, Jiayang and Qiu, Kevin and Huang, Qi and Zu, Xinrui and Yang, Shiping and Zhang, Hengyuan and Wong, Ngai and Ilievski, Filip and Yu, Shujian and Plaat, Aske and Ren, Zhaochun and Hoogendoorn, Mark and Fran{\c{c}}ois-Lavet, Vincent},
  year={2026},
  eprint={2606.21943},
  archivePrefix={arXiv},
  url={https://arxiv.org/abs/2606.21943}
}
\endgroup


\newpage
\appendix

{\large{\textbf{Appendix for \textit{MM-OptBench: A Solver-Grounded Benchmark for Multimodal Optimization Modeling}}}}

\section*{\large Table of Contents}
{\footnotesize
\begin{itemize}
    \item[A] \hyperref[app:extended_related_work_compact]{Extended Related Work}
    \item[B] \hyperref[app:scoring_pipeline]{Official Scoring Pipeline and Failure Attribution}
    \item[C] \hyperref[sec:appendix_dev_plan]{Benchmark Taxonomy and Design Principles}
    \begin{itemize}
        \item[C.1] \hyperref[subsec:target_families]{Benchmark Taxonomy and Coverage}
        \item[C.2] \hyperref[subsec:taxonomy_rationale]{Rationale for the Taxonomy Design}
        \item[C.3] \hyperref[subsec:visual_strategy]{Visual Design Principles Across Families}
    \end{itemize}
    \item[D] \hyperref[sec:appendix_construction]{Shared Benchmark Construction and Validation Protocol}
    \begin{itemize}
        \item[D.1] \hyperref[subsec:appendix_format]{Ground-Truth Artifacts and File Organization}
        \item[D.2] \hyperref[subsec:appendix_pipeline]{Shared Construction Protocol}
        \item[D.3] \hyperref[subsec:appendix_sampling]{Configuration Sampling and Difficulty Regimes}
        \item[D.4] \hyperref[subsec:appendix_solver_templates]{Reference Solver Implementations and Artifact Alignment}
        \item[D.5] \hyperref[subsec:appendix_qa]{Instance Validation and Quality Control}
        \item[D.6] \hyperref[subsec:expert_roles]{Expert-Guided Release Auditing and Residual Risks}
        \item[D.7] \hyperref[subsec:appendix_examples]{Four Illustrative Benchmark Instances}
    \end{itemize}
    \item[E] \hyperref[app:data_generation_pipeline]{Representative Family-Specific Data Generation Pipelines}
    \begin{itemize}
        \item[E.1] \hyperref[subsec:data_pipeline_network]{Network Optimization}
        \item[E.2] \hyperref[subsec:data_pipeline_location]{Location, Covering, and Assignment}
        \item[E.3] \hyperref[subsec:data_pipeline_scheduling]{Scheduling and Sequencing}
        \item[E.4] \hyperref[subsec:data_pipeline_multiperiod]{Multi-Period and System Planning}
        \item[E.5] \hyperref[subsec:data_pipeline_routing]{Routing and Tour Optimization}
        \item[E.6] \hyperref[subsec:data_pipeline_combinatorial]{Pure Combinatorial and Logical Models}
    \end{itemize}
    \item[F] \hyperref[app:extended_experiment_results]{Experimental Protocol, Extended Results, and Diagnostics}
    \begin{itemize}
        \item[F.1] \hyperref[subsec:app_exp_setup_details]{Experimental Setup Details}
        \item[F.2] \hyperref[subsec:app_extended_performance_tables]{Extended Performance Tables and pass@4 Heatmap}
        \begin{itemize}
            \item[F.2.1] \hyperref[subsubsec:app_overall_metrics]{Overall Metrics and Model-Level Performance}
            \item[F.2.2] \hyperref[subsubsec:app_family_results]{Family-Level Performance}
            \item[F.2.3] \hyperref[subsubsec:app_difficulty_results]{Difficulty-Level Performance}
            \item[F.2.4] \hyperref[subsubsec:app_pass4_heatmap]{pass@4 Structural Heatmap}
        \end{itemize}
        \item[F.3] \hyperref[app:failure_taxonomy_details]{Detailed Failure Taxonomy}
        \item[F.4] \hyperref[subsec:app_general_purpose_failure_analysis]{Failure Analysis for General-Purpose MLLMs}
        \item[F.5] \hyperref[subsec:app_math_specialized_failure_analysis]{Failure Analysis for Math-Specialized MLLMs}
        \item[F.6] \hyperref[subsec:app_runtime_analysis]{Reference Runtime and Evaluation Timeout Analysis}
        \item[F.7] \hyperref[subsec:app_controlled_evaluations]{Revision and Robustness Experiments}
    \end{itemize}
    \item[G] \hyperref[app:limitations_impact]{Limitations and Impact Statement}
\end{itemize}
}

\section{Extended Related Work}
\label{app:extended_related_work_compact}

Research on multimodal mathematical and scientific reasoning has broadened rapidly, expanding from educational question answering to scientific document understanding, structured-visual reasoning, visual mathematics, and interleaved multimodal generation. For the purpose of positioning MM-OptBench, the most relevant prior work can be organized into two layers: datasets used for multimodal mathematical or scientific training, and benchmarks used to evaluate multimodal reasoning performance. Tables~\ref{tab:rw-datasets-compact} and~\ref{tab:rw-benchmarks-compact} summarize representative works rather than attempting exhaustive coverage. Across these lines, prior work overwhelmingly studies whether a model can \emph{answer}, \emph{solve}, \emph{describe}, \emph{verify}, or \emph{generate} from multimodal inputs. By contrast, MM-OptBench evaluates whether a model can \emph{formulate} a formal optimization model from multimodal specifications and produce a solver-executable implementation. To the best of our knowledge, multimodal optimization-model synthesis has not been directly studied in prior MLLM mathematical and scientific datasets or benchmarks.

\begin{table*}[h]
\centering
\small
\setlength{\tabcolsep}{4pt}
\renewcommand{\arraystretch}{1.12}
\resizebox{\textwidth}{!}{%
\begin{tabular}{
>{\raggedright\arraybackslash}m{4cm}
>{\raggedright\arraybackslash}m{5.6cm}
>{\raggedright\arraybackslash}m{3.0cm}
>{\raggedright\arraybackslash}m{4cm}
>{\raggedright\arraybackslash}m{4cm}}
\hline
\textbf{Dataset Group} & \textbf{Representative Resources} & \textbf{Typical Input} & \textbf{Primary Supervision} & \textbf{Difference from MM-OptBench} \\
\hline
Educational multimodal QA &
ScienceQA~\citep{lu2022learn} &
Questions with images and textual context &
Answer prediction with explanations and supporting context &
Focuses on answer accuracy and rationale generation, not symbolic model construction. \\
\hline
Scientific documents and figures &
SPIQA~\citep{pramanick2024spiqa}, InstructDoc~\citep{tanaka2024instructdoc}, DocGenome~\citep{xia2024docgenome}, Multimodal ArXiv~\citep{li2024multimodal}, MMSci~\citep{li2024mmsci}, SciCap~\citep{hsu2021scicap} &
Scientific papers, document pages, figures, plots &
Paper QA, document understanding, scientific comprehension, captioning &
Requires scientific-media interpretation, but not optimization formulation or solver execution. \\
\hline
Structured visual carriers &
FigureQA~\citep{kahou2017figureqa}, DVQA~\citep{kafle2018dvqa}, UniChart~\citep{masry2023unichart}, MapQA~\citep{chang2022mapqa}, ChartInstructionData~\citep{zeng2024advancing} &
Charts, maps, figures, infographic-like visuals &
Visual reasoning, chart QA, chart-oriented instruction tuning &
Closest in modality to our setting, yet still centered on perception and QA rather than formulation synthesis. \\
\hline
Math-specific multimodal alignment &
COMET~\citep{liu2024comet}, TabMWP~\citep{lu2022dynamic}, GeoGPT4V~\citep{cai2024geogpt4v}, MultiMath~\citep{peng2024multimath}, Math-LLaVA~\citep{shi2024math}, MathGLM-Vision~\citep{yang2024mathglm}, MAVIS~\citep{zhang2024mavis} &
Visual math problems, tables, diagrams, mixed mathematical imagery &
Answer generation, step-wise solutions, visual-math alignment &
Improves multimodal math solving, but the target is still problem solving rather than formal model construction. \\
\hline
Large-scale pre-training and process supervision &
PIN~\citep{wang2024pin}, InfiMM-WebMath-40B~\citep{han2409infimm}, MM-PRM~\citep{du2025mm}, Zebra-CoT~\citep{li2025zebra} &
Interleaved documents, math web pages, multimodal reasoning traces &
Pre-training, interleaved document modeling, process reward modeling, visual chain-of-thought supervision &
Strengthens general reasoning or step-level supervision, but does not define optimization-model outputs. \\
\hline
\end{tabular}
}
\caption{Representative multimodal mathematical and scientific datasets and training resources. The table intentionally emphasizes representative works instead of exhaustive coverage.}
\label{tab:rw-datasets-compact}
\end{table*}

\begin{table*}[h]
\centering
\scriptsize
\setlength{\tabcolsep}{4pt}
\renewcommand{\arraystretch}{1.12}
\resizebox{\textwidth}{!}{%
\begin{tabular}{
>{\raggedright\arraybackslash}m{4.2cm}
>{\raggedright\arraybackslash}m{5.6cm}
>{\raggedright\arraybackslash}m{3.0cm}
>{\raggedright\arraybackslash}m{4cm}
>{\raggedright\arraybackslash}m{4cm}}
\hline
\textbf{Benchmark Group} & \textbf{Representative Benchmarks} & \textbf{Typical Modality} & \textbf{Target Output} & \textbf{Gap to MM-OptBench} \\
\hline
Geometry reasoning &
GEOS~\citep{seo2015solving}, Geometry3K / Inter-GPS~\citep{lu2021inter}, GeoQA~\citep{chen2021geoqa}, GeoEval~\citep{zhang2024geoeval} &
Geometry text and diagrams &
Answer, formal language, or proof-like reasoning trace &
Evaluates diagram-grounded geometry solving, not optimization-model synthesis. \\
\hline
Visual mathematical reasoning &
MathVista~\citep{lu2024mathvista}, MathVerse~\citep{zhang2024mathverse}, MATH-Vision~\citep{wang2024measuring}, MM-MATH~\citep{sun2024mm}, WE-MATH~\citep{qiao2025we}, CMM-Math~\citep{liu2025cmm} &
Math problems with diagrams, figures, or mixed visual contexts &
Answer prediction, process-aware evaluation, error analysis &
Measures whether models can solve visual math problems, rather than formulate solver-ready mathematical programs. \\
\hline
Broad expert-level evaluation &
MMMU~\citep{yue2024mmmu}, CMMMU~\citep{zhang2024cmmmu}, M3Exam~\citep{zhang2023m3exam}, OlympiadBench~\citep{he2024olympiadbench}, SciBench~\citep{wang2024scibench} &
Heterogeneous expert visuals, textbook images, charts, exam figures &
Answer prediction across subjects and difficulty levels &
Broader in subject coverage, but still answer-centric rather than formulation-centric. \\
\hline
Structured visual and document reasoning &
AI2D~\citep{kembhavi2016diagram}, IconQA~\citep{lu2021iconqa}, TQA~\citep{kembhavi2017you}, ChartQA~\citep{masry2022chartqa}, ChartBench~\citep{xu2023chartbench}, DocVQA~\citep{mathew2021docvqa}, TableBench~\citep{wu2025tablebench}, GRAB~\citep{roberts2025grab}, SCIVER~\citep{wang2025sciver} &
Diagrams, charts, tables, documents, graphs, scientific evidence &
QA, fact verification, semantic parsing, claim verification &
Most relevant in visual carrier type, but still evaluates extraction, verification, or reasoning over fixed questions. \\
\hline
Generation-oriented multimodal evaluation &
ChartMimic~\citep{yang2024chartmimic}, Plot2Code~\citep{wu2025plot2code}, Sketch2Code~\citep{li2025sketch2code}, FullFront~\citep{sun2025fullfront}, MMIE~\citep{xia2025mmie} &
Charts, plots, sketches, webpages, interleaved image-text sequences &
Code, UI artifacts, summaries, mixed understanding-generation outputs &
Extends beyond QA, but the outputs are still generic code or generation artifacts rather than formal optimization models. \\
\hline
\end{tabular}
}
\caption{Representative multimodal mathematical and scientific benchmarks grouped by evaluation focus. Prior work studies multimodal reasoning, but not multimodal optimization-model formulation.}
\label{tab:rw-benchmarks-compact}
\end{table*}

\paragraph{Training Resources for Multimodal Mathematical and Scientific Reasoning.}
Existing training-oriented resources for multimodal mathematical and scientific reasoning fall into several recurring patterns. One line develops educational QA corpora with explanation supervision, of which ScienceQA~\citep{lu2022learn} is the canonical example. A second line moves from textbook-style questions to scientific media. SPIQA~\citep{pramanick2024spiqa}, InstructDoc~\citep{tanaka2024instructdoc}, DocGenome~\citep{xia2024docgenome}, and Multimodal ArXiv~\citep{li2024multimodal} study question answering and comprehension over scientific papers, multi-page documents, and abstract figures; MMSci~\citep{li2024mmsci} extends this agenda to expert scientific comprehension; and SciCap~\citep{hsu2021scicap} focuses on scientific figure captioning. These resources are relevant to MM-OptBench because real optimization problems are often communicated through reports, tables, figures, or scientific documents rather than through a single clean textual prompt. Nevertheless, their supervision remains document-centric: models are trained to answer, retrieve, classify, or describe, not to synthesize a formal mathematical program.

Another important dataset line focuses on structured visual carriers and multimodal mathematical alignment. FigureQA~\citep{kahou2017figureqa}, DVQA~\citep{kafle2018dvqa}, UniChart~\citep{masry2023unichart}, MapQA~\citep{chang2022mapqa}, and ChartInstructionData~\citep{zeng2024advancing} treat charts, maps, and scientific-style graphics as first-class inputs for visual reasoning and instruction tuning. More directly relevant to mathematical reasoning are COMET~\citep{liu2024comet}, TabMWP~\citep{lu2022dynamic}, GeoGPT4V~\citep{cai2024geogpt4v}, MultiMath~\citep{peng2024multimath}, Math-LLaVA~\citep{shi2024math}, MathGLM-Vision~\citep{yang2024mathglm}, and MAVIS~\citep{zhang2024mavis}, which improve visual-math alignment, step-wise solution generation, and mathematical instruction tuning. At a larger scale, PIN~\citep{wang2024pin} and InfiMM-WebMath-40B~\citep{han2409infimm} provide large interleaved multimodal corpora, while MM-PRM~\citep{du2025mm} and Zebra-CoT~\citep{li2025zebra} move toward process-level supervision and interleaved visual reasoning traces. These developments substantially strengthen multimodal mathematical reasoning, but they remain solving-oriented rather than formulation-oriented.

\paragraph{Benchmarks for Geometry, Visual Math, and Expert Reasoning.}
On the evaluation side, geometry is the most mature multimodal mathematical subfield. GEOS~\citep{seo2015solving} first combined text understanding and diagram interpretation for geometry problem solving; Geometry3K / Inter-GPS~\citep{lu2021inter} introduced larger benchmarks with formal-language annotation and symbolic reasoning; GeoQA~\citep{chen2021geoqa} emphasized geometric question answering grounded in textual and visual information; and GeoEval~\citep{zhang2024geoeval} consolidated this line with standardized hard and backward-reasoning subsets for contemporary LLMs and MLLMs. This literature is important because it establishes that multimodal mathematical reasoning demands structured grounding rather than generic image understanding. Yet the target artifact remains an answer or a problem-specific reasoning trace, not a reusable optimization model constructed from multimodal specifications.

Recent benchmarks broaden this agenda from geometry to general visual mathematics and expert-level reasoning. MathVista~\citep{lu2024mathvista}, MathVerse~\citep{zhang2024mathverse}, MATH-Vision~\citep{wang2024measuring}, MM-MATH~\citep{sun2024mm}, WE-MATH~\citep{qiao2025we}, and CMM-Math~\citep{liu2025cmm} evaluate multimodal mathematical reasoning under diagrams, plots, and other visual contexts, often with finer error taxonomies or stronger control over textual shortcutting. In parallel, MMMU~\citep{yue2024mmmu}, CMMMU~\citep{zhang2024cmmmu}, M3Exam~\citep{zhang2023m3exam}, OlympiadBench~\citep{he2024olympiadbench}, and SciBench~\citep{wang2024scibench} test college-level, multilingual, or olympiad-level expert reasoning across disciplines. These benchmarks are broader than MM-OptBench in subject coverage, but they still evaluate whether a model can solve a posed problem rather than whether it can construct the underlying formal program.

\paragraph{Structured Visual Reasoning and Generation-Oriented Evaluation.}
The benchmarks most closely related to MM-OptBench are those built around structured visual media and richer multimodal outputs. AI2D~\citep{kembhavi2016diagram}, IconQA~\citep{lu2021iconqa}, and TQA~\citep{kembhavi2017you} study diagram and textbook reasoning, while ChartQA~\citep{masry2022chartqa}, ChartBench~\citep{xu2023chartbench}, DocVQA~\citep{mathew2021docvqa}, TableBench~\citep{wu2025tablebench}, GRAB~\citep{roberts2025grab}, and SCIVER~\citep{wang2025sciver} evaluate charts, documents, tables, graphs, and multimodal scientific evidence. These settings are especially relevant because optimization problems in practice are frequently specified through tables, schedules, layouts, figures, or long-form technical material rather than a single paragraph. Even so, their outputs remain answers, labels, summaries, semantic parses, or verification decisions over fixed prompts.

A more recent generation of benchmarks moves beyond pure question answering toward cross-modal generation and interleaved multimodal understanding. ChartMimic~\citep{yang2024chartmimic} and Plot2Code~\citep{wu2025plot2code} evaluate visually grounded code generation from charts and scientific plots; Sketch2Code~\citep{li2025sketch2code} and FullFront~\citep{sun2025fullfront} benchmark UI and front-end generation from visual inputs; and MMIE~\citep{xia2025mmie} evaluates large vision-language models on interleaved multimodal comprehension. These works show that multimodal evaluation is expanding beyond short-answer QA. However, the generated artifacts remain generic code, interface outputs, or interleaved understanding-generation products rather than formal optimization models whose semantic validity can be checked by solver execution.

\paragraph{Positioning MM-OptBench.}
The distinction above motivates MM-OptBench as a benchmark for \emph{multimodal formulation} rather than \emph{multimodal solving}. Existing datasets and benchmarks primarily ask whether a model can answer a question, solve a mathematical problem, summarize a chart, verify a claim, or generate a piece of code from a visual input. MM-OptBench instead asks whether the model can identify the symbolic scaffold of an optimization task from multimodal evidence and translate it into a correct optimization formulation and executable solver implementation. This shift alters both the output space and the evaluation protocol: the target is no longer a scalar answer or free-form explanation, but a formally specified optimization model whose correctness can be assessed through solver-grounded verification. In this sense, prior multimodal mathematical and scientific resources provide important perceptual and reasoning prerequisites for our setting, but they do not directly evaluate the optimization-modeling capability itself.
\section{Official Scoring Pipeline and Failure Attribution}
\label{app:scoring_pipeline}

\begin{figure*}[h]
    \centering
    \begin{tikzpicture}[
        font=\footnotesize,
        box/.style={draw, rounded corners=2pt, align=center, text width=2.55cm, minimum height=1.02cm, fill=blue!5},
        auxbox/.style={draw, rounded corners=2pt, align=center, text width=2.50cm, minimum height=1.02cm, fill=gray!12},
        outcome/.style={draw, rounded corners=2pt, align=center, text width=2.50cm, minimum height=1.02cm, fill=green!6},
        warn/.style={draw, rounded corners=2pt, align=center, text width=2.50cm, minimum height=1.02cm, fill=red!6},
        decision/.style={draw, diamond, aspect=1.85, align=center, text width=1.55cm, inner sep=1.4pt, fill=orange!10},
        stageonegroup/.style={draw=blue!45, dashed, rounded corners=4pt, line width=0.6pt},
        stagetwogroup/.style={draw=orange!65!black, dashed, rounded corners=4pt, line width=0.6pt},
        oraclegroup/.style={draw=gray!70, dashed, rounded corners=4pt, line width=0.6pt},
        verifiedgroup/.style={draw=gray!70, dashed, rounded corners=4pt, line width=0.6pt},
        auxsupergroup/.style={draw=gray!55, dashed, rounded corners=5pt, line width=0.6pt},
        wideaux/.style={draw, rounded corners=2pt, align=center, text width=2.95cm, minimum height=1.02cm, fill=gray!12},
        wideoutcome/.style={draw, rounded corners=2pt, align=center, text width=2.95cm, minimum height=1.02cm, fill=green!6},
        line/.style={-{Latex[length=2.4mm,width=1.7mm]}, thick, shorten >=1pt, shorten <=1pt},
        auxline/.style={-{Latex[length=2.4mm,width=1.7mm]}, thick, dashed, shorten >=1pt, shorten <=1pt}
    ]
        \draw[stageonegroup] (-0.15, 3.70) rectangle (6.25, 11.15);
        \draw[stagetwogroup] (-0.15, -1.05) rectangle (13.1, 3.15);
        \draw[auxsupergroup] (6.38, 4.10) rectangle (13.4, 11.15);
        \draw[oraclegroup] (9.9, 4.45) rectangle (13.2, 10.55);
        \draw[verifiedgroup] (6.45, 4.45) rectangle (9.85, 10.55);

        \node[align=center, font=\footnotesize\bfseries] at (6.48, 11.55) {Official scoring and failure attribution};
        \node[align=center, font=\footnotesize\bfseries] at (9.82, 10.85) {Auxiliary ablations (not scored)};

        \node[box] (input) at (1.35, 10.35) {Model-facing input\\\texttt{task\_input.txt}\\+ \texttt{visuals/}};
        \node[box] (stageone) at (1.35, 8.45) {Stage 1:\\formulation $M$\\+ \texttt{solve()} artifact};
        \node[box] (exec) at (1.35, 6.55) {Evaluation program:\\run \texttt{solve()}\\check task criterion};
        \node[outcome] (validcode) at (4.75, 6.55) {Record\\Valid Code\\Rate};
        \node[decision] (successq) at (1.35, 4.55) {Solver-correct?};
        \node[outcome] (score) at (4.75, 4.55) {Solved for\\pass@$k$};

        \node[box] (stagetwo) at (1.35, 2.25) {Stage 2:\\extraction-only\\follow-up};
        \node[decision] (parseq) at (4.75, 2.25) {Parseable\\extraction?};
        \node[decision] (extractq) at (8.15, 2.25) {Matches\\reference?};
        \node[warn] (downstream) at (11.55, 2.25) {Stage-2 match;\\Stage-1 artifact\\incorrect};
        \node[warn] (unresolved) at (4.75, -0.25) {Unresolved:\\extraction not\\parseable};
        \node[warn] (reading) at (8.15, -0.25) {Stage-2\\extraction\\mismatch};

        \node[wideoutcome] (oraclemetric) at (11.55, 9.85) {Oracle\\downstream\\accuracy};
        \node[wideaux] (oracle) at (11.55, 7.55) {Oracle-reading:\\formulation and\\code generation};
        \node[wideaux] (oracledata) at (11.55, 5.25) {Ground-truth\\\texttt{instance\_data.json}};

        \node[wideoutcome] (verifiedmetric) at (8.15, 9.85) {Verified-subset\\downstream\\accuracy};
        \node[wideaux] (verified) at (8.15, 7.55) {Verified-extraction:\\formulation and\\code generation};
        \node[wideaux] (verifiedsrc) at (8.15, 5.25) {Verified\\Stage-2\\extractions};

        \draw[line] (input.south) -- (stageone.north);
        \draw[line] (stageone.south) -- (exec.north);
        \draw[line] (exec.east) -- (validcode.west);
        \draw[line] (exec.south) -- (successq.north);
        \draw[line] (successq.east) -- node[above]{yes} (score.west);
        \draw[line] (successq.south) -- node[right]{no} (stagetwo.north);

        \draw[line] (stagetwo.east) -- (parseq.west);
        \draw[line] (parseq.east) -- node[above]{yes} (extractq.west);
        \draw[line] (parseq.south) -- node[right]{no} (unresolved.north);
        \draw[line] (extractq.east) -- node[above]{yes} (downstream.west);
        \draw[line] (extractq.south) -- node[right]{no} (reading.north);

        \draw[auxline] (oracledata.north) -- (oracle.south);
        \draw[auxline] (oracle.north) -- (oraclemetric.south);
        \draw[auxline] (extractq.north) -- (verifiedsrc.south);
        \draw[auxline] (verifiedsrc.north) -- (verified.south);
        \draw[auxline] (verified.north) -- (verifiedmetric.south);
    \end{tikzpicture}
    \caption{Official scoring pipeline and failure-attribution flow. Solid arrows show the official Stage-1 scoring path and the Stage-2 attribution path for failed samples. Valid Code Rate measures whether the generated code runs successfully, while pass@$k$ requires the solver-correct branch. Dashed arrows show two separate auxiliary ablations used to measure downstream formulation-and-code capability after public-instance reading is bypassed or independently verified.}
    \label{fig:evaluation_pipeline}
\end{figure*}
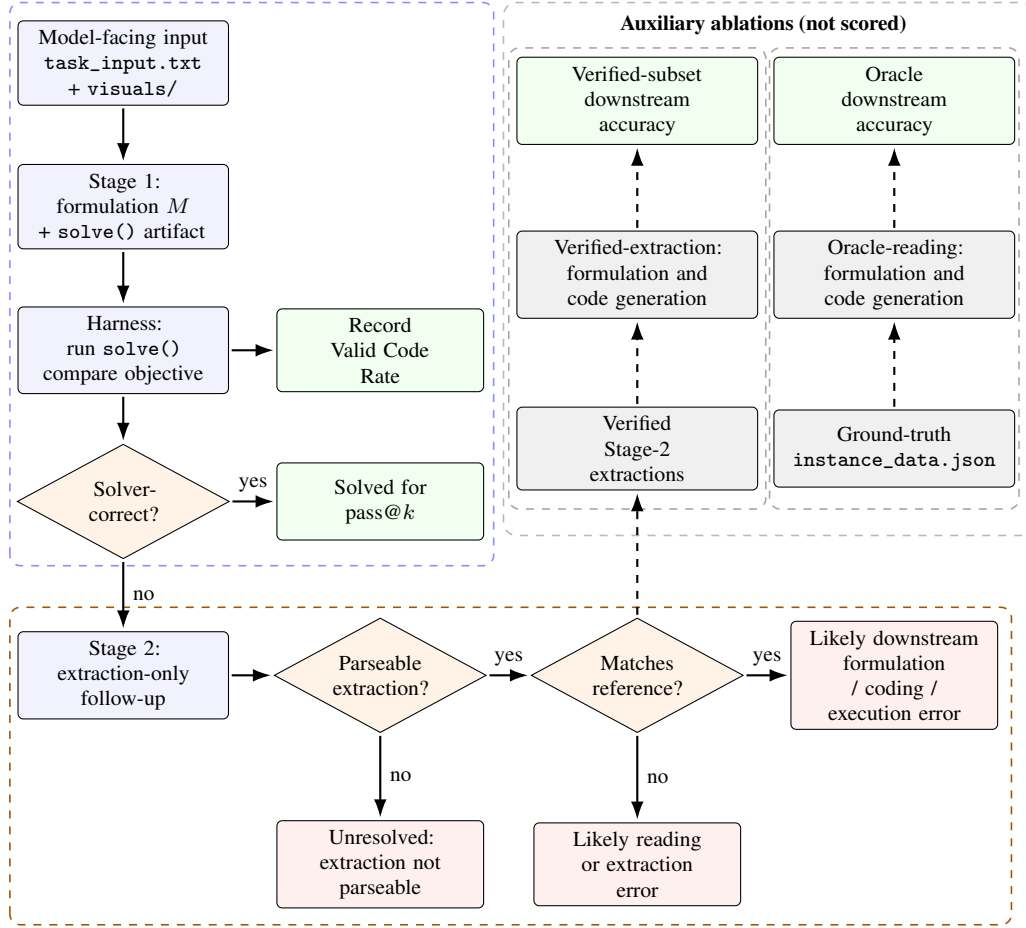

Figure~\ref{fig:evaluation_pipeline} should be read as separating three questions: whether the submitted solver artifact solves the original benchmark instance, whether a separate extraction follow-up recovers the public instance data correctly, and how much downstream formulation-and-code capability remains once public-instance reading is removed or verified. The solid arrows answer the first two questions and define the official scoring and attribution path. The dashed arrows answer the third question through auxiliary ablations. Colors encode function rather than model family: blue nodes are model-facing or execution steps, orange diamonds are checks, green nodes are metric outcomes, red nodes are failure categories, and gray nodes are diagnostic ablation inputs or generation steps; blue, orange, and gray dashed frames mark Stage 1, Stage 2, and auxiliary scopes.

The separation between scoring and diagnosis is methodological. Official scoring only needs to know whether the generated \texttt{solve()} artifact satisfies the task-specific criterion in Section~\ref{subsec:setting_protocol}. Failure diagnosis needs a different object: a structured extraction of the public instance data that can be compared with \texttt{instance\_data.json}. If both requirements were bundled into one response, an error in the diagnostic extraction format could block or confuse evaluation of the solver artifact itself. MM-OptBench therefore runs the solver artifact first for the official score, and asks the extraction question only afterward, only for attribution.

\textbf{Stage 1 is the only scoring stage.}
The model receives only the model-facing inputs $\mathcal{I}_x$: the released \texttt{task\_input.txt} and visual artifact(s). It then generates a mathematical formulation and a solver-executable artifact exposing \texttt{solve()}. The evaluation program runs \texttt{solve()} in an isolated solver environment and checks the returned result against the task-specific criterion in Section~\ref{subsec:setting_protocol}. Samples that fail to expose a runnable solver, fail during execution, or fail that criterion are incorrect for pass@$k$; Valid Code Rate records executability separately. No diagnostic branch can turn a failed Stage-1 sample into a solved official sample.

\textbf{Stage 2 attributes failures rather than rescoring them.}
Stage 2 is triggered only after a Stage-1 failure. It asks the model to extract the public instance data from the same multimodal input without solving the optimization problem. The extracted record is compared against \texttt{instance\_data.json}, the machine-readable public instance record used to derive the text, visuals, reference formulation, solver, and verified optimum. A mismatch means the model extracted incorrect data in Stage 2. A match means it extracted the data correctly. Because Stage 2 is a separate request, these results do not tell us what data the model used in Stage 1. If the response cannot be parsed as structured instance data, the case is left unresolved instead of being forced into either category.

\textbf{The auxiliary ablations isolate downstream capability.}
The right panel of Figure~\ref{fig:evaluation_pipeline} contains two non-scoring ablations, both drawn bottom-to-top from structured instance data to downstream accuracy. \emph{Oracle-reading} bypasses multimodal reading entirely by giving the model \texttt{instance\_data.json} and asking only for formulation and solver generation. \emph{Verified-extraction} is less idealized: it starts from the original multimodal input, but forwards only Stage-2 extractions that match \texttt{instance\_data.json} to downstream formulation and code generation. Comparing these diagnostic accuracies with Stage-1 pass@$k$ characterizes performance when instance reading is bypassed or independently verified.

\section{Benchmark Taxonomy and Design Principles}
\label{sec:appendix_dev_plan}
\begin{table*}[h]
\centering
\footnotesize
\caption{Taxonomy of MM-OptBench. The table summarizes the major problem families, the benchmark subcategories, their core modeling characteristics, and their primary visual encodings.}
\label{tab:taxonomy_scope}
\renewcommand{\arraystretch}{1.08}
\resizebox{\textwidth}{!}{%
\begin{tabular}{
>{\raggedright\arraybackslash}m{6cm}
>{\raggedright\arraybackslash}m{6cm}
>{\raggedright\arraybackslash}m{6.8cm}
>{\raggedright\arraybackslash}m{6cm}
}
\toprule
\textbf{Major Category} & \textbf{Subcategory} & \textbf{Core Characteristics} & \textbf{Primary Visual Encoding} \\
\midrule

\multirow{5}{6cm}{\raggedright Network Optimization\\(flow balance, path, and design structure)}
& Minimum-Cost Flow
& Supply-demand balance, arc capacities, and route-cost trade-offs in directed networks.
& Directed layered network with node balances and arc $(\mathrm{capacity}, \mathrm{cost})$ labels. \\

& Maximum Flow / Minimum Cut
& Source-sink throughput, bottleneck capacities, and cut-structure reasoning.
& Directed network with marked source/sink and arc-capacity labels. \\

& Resource-Constrained Shortest Path
& Path selection under additive arc costs and multiple resource limits, where feasibility depends jointly on topology and path-wise resource accumulation.
& Directed network with compact arc coefficient labels for easy cases; topology-plus-parameter-table packaging for denser medium/hard settings. \\

& Capacitated Network Design
& Binary arc-installation decisions coupled with routed flow, fixed-versus-variable cost trade-offs, and capacity-linked design choices.
& Directed network design diagram with source/sink, admissible arcs, and arc-level cost/capacity annotations supporting both routing and installation decisions. \\

& Time-Expanded / Multi-Period Network Flow
& Time-indexed flow conservation, carry-over or inventory arcs, and inter-period balance under temporally varying capacities or demand.
& Time-layered network view or coordinated network-plus-planning panel showing physical topology, time structure, and carry-over relations. \\
\midrule

\multirow{5}{6cm}{\raggedright Location, Covering, and Assignment\\(selection, coverage, and allocation coupling)}
& Facility Location
& Facility opening, geometric coverage feasibility, and demand-weighted customer assignment.
& Spatial coverage diagram with facilities, customers, and coverage regions. \\

& $p$-Median / $p$-Center
& Selection of exactly $p$ sites under total-distance or max-distance service objectives.
& Map-style site-demand layout with labeled coordinates and local attributes. \\

& Set Covering / Set Partitioning
& Binary subset selection under covering or exact-partitioning semantics with cost trade-offs.
& Subset-incidence table with aligned cost bars; split panels for wider instances. \\

& Bipartite Assignment / Matching
& Sparse compatibility relations and assignment costs over bipartite pairs.
& Bipartite compatibility graph plus aligned cost table. \\

& Generalized Assignment Problem
& Agent capacities, heterogeneous costs/resources, and sparse feasible task-agent pairs.
& Dual exact tables for assignment costs/forbidden pairs and resource/capacity data. \\
\midrule

\multirow{4}{6cm}{\raggedright Scheduling and Sequencing\\(precedence and resource-conflict reasoning)}
& Job-Shop Scheduling
& Job-specific machine routes, precedence chains, and machine non-overlap.
& Job-operation layout diagram; not a solved Gantt chart. \\

& Parallel Machine Scheduling
& Machine eligibility or relatedness together with job durations and load-balance structure.
& Job-machine eligibility matrix with processing-duration bars. \\

& Resource-Constrained Project Scheduling Problem (RCPSP)
& Precedence DAGs with renewable-resource capacities and activity-resource demands.
& Two-view package: precedence network plus resource/data matrix-chart view. \\

& Flexible / Hybrid Flow-Shop Scheduling
& Stage-based processing with parallel machines within stages and bottleneck-stage effects.
& Combined stage-architecture diagram plus job-stage processing-time matrix. \\
\midrule

\multirow{4}{6cm}{\raggedright Multi-Period and System Planning\\(time-indexed balance and cross-period coupling)}
& Lot-Sizing and Production Planning
& Inventory balance, setup-linking, and shared period-capacity constraints.
& Integrated planning dashboard with capacity timeline and demand/cost tables. \\

& Energy Dispatch / Unit Commitment
& Dispatch levels, startup decisions, and generator-capacity trade-offs over time.
& Generator economics table, demand timeline, and capacity-range chart. \\

& Workforce Planning / Shift Scheduling
& Staffing coverage, worker availability, overtime, and temporal rest constraints.
& Complementary demand heatmap plus worker profile/availability table(s). \\

& Inventory-Routing Problem
& Joint replenishment and routing decisions across periods with inventory carry-over.
& Two-view package: routing map plus planning-data dashboard. \\
\midrule

\multirow{4}{6cm}{\raggedright Routing and Tour Optimization\\(tour connectivity and route partitioning in Euclidean space)}
& Traveling Salesman Problem
& Hamiltonian tour construction over city coordinates.
& City map with identifiers and coordinates; complete graph omitted. \\

& Vehicle Routing Problem
& Capacity-constrained route partitioning from a depot to customers.
& Depot-customer map with demand labels; sparse edges only in smaller cases. \\

& TSP with Time Windows
& Tour connectivity with service times and time-window feasibility.
& Spatial node map with coordinates, service times, and time-window labels. \\

& Heterogeneous Fleet VRP
& Routing coupled with fleet categories, capacities, fixed costs, and distance multipliers.
& Depot-customer map plus fleet-attribute planning sheet. \\
\midrule

\multirow{4}{6cm}{\raggedright Pure Combinatorial and Logical Models\\(binary selection and logical consistency)}
& Multi-Dimensional Knapsack
& Item selection under multiple resource capacities and value-weight trade-offs.
& Item-value sheet with segmented resource-consumption bars and capacity bars. \\

& Graph Coloring
& Adjacency conflicts and chromatic-number optimization.
& Adjacency matrix. \\

& Set Packing
& Selection of mutually compatible subsets under weighted objectives.
& Incidence matrix with subset value bars. \\

& Logical Constraint Satisfaction Models
& Clause, implication, and at-most-one logic as constrained binary optimization.
& Clause-variable matrix with signed literals and row-family shading. \\

\bottomrule
\end{tabular}%
}
\end{table*}

MM-OptBench must be broad enough to cover distinct optimization structures, yet controlled enough that each instance remains visually legible, solver-verifiable, and non-leaking. We therefore first summarize the released taxonomy and family coverage (Sec.~\ref{subsec:target_families}), then explain the source triangulation and selection criteria behind the taxonomy (Sec.~\ref{subsec:taxonomy_rationale}), and finally describe visual design principles across families (Sec.~\ref{subsec:visual_strategy}).

\subsection{Benchmark Taxonomy and Coverage}
\label{subsec:target_families}

MM-OptBench is organized around six major problem families defined primarily by \emph{mathematical structure} rather than application semantics. Table~\ref{tab:taxonomy_scope} summarizes the full benchmark taxonomy, which comprises 26 subcategories. In the finalized benchmark design, each subcategory is instantiated at three difficulty levels (easy/medium/hard) with 10 instances per level, yielding 780 instances in total.

This scale is intended to balance breadth and controllability: it is large enough to cover diverse optimization paradigms and multimodal encodings, while still supporting family-level analysis, difficulty-stratified evaluation, and systematic ablation across subcategories.

\subsection{Rationale for the Taxonomy Design}
\label{subsec:taxonomy_rationale}

\textbf{Source triangulation.} The taxonomy in Table~\ref{tab:taxonomy_scope} was not chosen ad hoc. Instead, we triangulated across three source categories when identifying candidate problem families and variants: \textit{(i) textbooks and academic teaching materials}, including classical operations research texts, university course repositories, and constraint-programming examples such as MiniZinc models; \textit{(ii) open benchmark libraries}, including OR-Library, CVRPLIB, PSPLIB, MIPLIB (used only for structural inspiration), and the MiniZinc benchmark suite; and \textit{(iii) open-source modeling repositories}, including representative examples from Pyomo, Gurobi, and OR-Tools. These sources were used strictly for structural inspiration rather than direct instance reuse: all benchmark instances are regenerated through our own configuration-sampling and verification pipeline to ensure reproducibility and to avoid licensing or artifact-consistency issues.

\textbf{Selection criteria.} Within this broad candidate pool, we selected the retained families and subcategories using six criteria. \textit{Structural diversity} requires the benchmark to cover different modeling primitives, such as flow conservation, assignment, activation, precedence, inventory balance, routing connectivity, and logical consistency, rather than many surface variants of one formulation template. \textit{Canonicality and popularity in the optimization literature} favors problems that are standard in OR textbooks, courses, benchmark suites, or modeling examples, so that the benchmark tests established modeling knowledge rather than obscure custom puzzles. \textit{Multimodal naturalness} requires the problem to have a plausible visual carrier, such as a network diagram, spatial layout, schedule chart, dashboard, matrix, or table, in which instance-defining information is naturally communicated. \textit{Difficulty scalability} requires each subcategory to support controlled easy/medium/hard regimes through changes in size, coupling, temporal depth, sparsity, or integrality burden. \textit{Solver verifiability} requires generated instances to have exact or otherwise reliable reference solutions under the released evaluation protocol. \textit{Cross-family non-redundancy} avoids retaining two families that mainly test the same formulation operation under different names.

\textbf{Family-level rationale.} These criteria determine both the major families and the subcategories within them. \textit{Network Optimization} includes minimum-cost flow, maximum flow, RCSP, capacitated network design, and time-expanded multi-period flow because these subcategories span conservation, cut, path-resource accumulation, binary design coupling, and temporal expansion rather than merely repeating one graph template. \textit{Location, Covering, and Assignment} tests selection-and-allocation reasoning through spatial coverage, \(p\)-median/\(p\)-center placement, set covering/partitioning, bipartite matching, and generalized assignment, which differ in whether feasibility is driven by geometry, incidence, compatibility, or capacity. \textit{Scheduling and Sequencing} is diversified across job-shop, parallel-machine, RCPSP, and flexible flow-shop settings so that temporal reasoning is tested under job-specific routing, eligibility constraints, renewable-resource coupling, and stage-based architecture. \textit{Multi-Period and System Planning} covers time-indexed balance and cross-period coupling through lot-sizing, energy dispatch/unit commitment, workforce planning, and inventory routing, where visual inputs naturally combine tables, timelines, heatmaps, and planning dashboards. \textit{Routing and Tour Optimization} focuses on global connectivity and route partitioning in Euclidean settings, ranging from TSP to capacitated VRP, time-window routing, and heterogeneous-fleet routing. \textit{Pure Combinatorial and Logical Models} includes multi-dimensional knapsack, graph coloring, set packing, and logical constraint satisfaction because these problems share binary decision structure but differ sharply in whether the dominant reasoning is capacity-based, adjacency-based, incidence-based, or logical.

\textbf{Breadth and coherence.} The resulting taxonomy balances breadth and coherence. It is broad enough to cover major modeling motifs that recur in both benchmark libraries and real decision-support settings, yet coherent enough that each retained family corresponds to a distinctive formulation paradigm and a natural multimodal encoding. This balance is important for MM-OptBench: we aim to evaluate genuine optimization-modeling capability rather than narrow template recall within a single problem style.

\textbf{Structural Separation of Families.} Table~\ref{tab:taxonomy_structure} complements Table~\ref{tab:taxonomy_scope} by summarizing the \emph{dominant} modeling motifs of the six major families. We emphasize dominant motifs rather than all secondary features: some subcategories introduce additional extensions, such as path-resource accumulation in resource-constrained shortest path, binary design decisions in capacitated network design, temporal expansion in multi-period network flow, or time windows in routing. However, these extensions do not change the primary structural identity of the family to which the subcategory belongs.

\begin{table}[h]
\centering
\small
\caption{Dominant structural motifs across the six major MM-OptBench families. A checkmark indicates a family-defining modeling primitive; some subcategories may additionally contain secondary motifs not shown in this summary table.}
\label{tab:taxonomy_structure}
\resizebox{\textwidth}{!}{
\begin{tabular}{lcccccc}
\toprule
Family & Flow & Time & Connectivity & Assignment & Sequencing & Logic \\
\midrule
Network Optimization & \checkmark &  &  &  &  &  \\
Location, Covering, and Assignment &  &  &  & \checkmark &  &  \\
Scheduling and Sequencing &  & \checkmark &  &  & \checkmark &  \\
Multi-Period and System Planning & \checkmark & \checkmark &  &  &  &  \\
Routing and Tour Optimization &  &  & \checkmark &  &  &  \\
Pure Combinatorial and Logical Models &  &  &  &  &  & \checkmark \\
\bottomrule
\end{tabular}
}
\end{table}

This structural separation helps ensure that strong performance in one family does not simply reflect reuse of a near-identical modeling template in another. For example, flow-balance reasoning in network models does not by itself resolve global tour-connectivity constraints in routing problems, and precedence-based reasoning in scheduling does not directly transfer to incidence- or logic-dominated combinatorial models. By organizing MM-OptBench around these family-defining structural motifs, we aim to evaluate whether a model can generalize across genuinely different optimization paradigms rather than only across superficially varied instances of the same template.

\subsection{Visual Design Principles Across Families}
\label{subsec:visual_strategy}

This subsection explains how MM-OptBench turns structured optimization instances into multimodal inputs. The design goal is not to make decorative figures, but to choose visual carriers that are faithful to the mathematical structure of each family, expose the public data needed for formulation, avoid revealing solved decisions, and remain readable at benchmark scale.

\textbf{Structure--visual alignment.} The visual encodings in Table~\ref{tab:taxonomy_scope} were chosen to match the dominant structural language of each optimization family, rather than to impose a single generic image format. \textit{Network Optimization} is represented through directed network diagrams because its core structure is topological. \textit{Location, Covering, and Assignment} uses spatial, compatibility, or incidence views because the main modeling question is which entities can be selected, covered, matched, or assigned. \textit{Scheduling and Sequencing} uses operation, precedence, stage, and resource views because the central structure is temporal order under shared resources. \textit{Multi-Period and System Planning} uses timelines and coordinated planning dashboards because decisions are coupled across periods. \textit{Routing and Tour Optimization} uses map-like spatial views because connectivity and route partitioning are defined over locations. \textit{Pure Combinatorial and Logical Models} uses tables and matrices because the relevant structure is discrete incidence, conflict, capacity, or logical relation.

\textbf{Exact instance information.} Within these family-specific carriers, the visuals encode instance-defining data rather than illustrative context. In \textit{Network Optimization}, node labels, arc directions, and arc annotations specify supplies/demands, capacities, costs, and resource coefficients that determine conservation and flow feasibility. In \textit{Location, Covering, and Assignment}, coordinates, coverage radii, compatibility edges, incidence entries, opening costs, demands, and capacities determine which assignments or selections are feasible and how they are priced. In \textit{Scheduling and Sequencing}, operation blocks, precedence arcs, machine labels, processing times, eligibility entries, and resource-capacity panels define the temporal and resource constraints of the scheduling model. In \textit{Multi-Period and System Planning}, timelines, demand profiles, cost tables, capacity ranges, inventory parameters, availability matrices, and planning dashboards specify the period-indexed data that drive balance and coupling constraints. In \textit{Routing and Tour Optimization}, maps provide depot/customer/city coordinates, demand labels, time windows, service times, fleet attributes, and distance rules needed to instantiate tour or vehicle-routing models. In \textit{Pure Combinatorial and Logical Models}, item tables, resource bars, adjacency matrices, incidence matrices, and clause-variable matrices specify binary-choice structure, conflicts, capacities, profits, and logical relations. A model that ignores these visuals therefore misses part of the formal instance specification.

\textbf{Dense and multi-view cases.} When a single figure would become overloaded, MM-OptBench uses exact tables, matrices, or coordinated multi-panel views rather than decorative simplifications. In \textit{Network Optimization}, most instances can be shown as one annotated graph, but denser resource-constrained or time-expanded variants may require auxiliary coefficient tables or layered views. In \textit{Location, Covering, and Assignment}, spatial layouts are sufficient for geometric cases, whereas incidence-heavy or capacity-heavy cases such as set covering and generalized assignment use matrix or table views. In \textit{Scheduling and Sequencing}, job-shop layouts can be compact, but RCPSP and flexible flow-shop instances often need separate precedence, stage, eligibility, or resource panels. In \textit{Multi-Period and System Planning}, coordinated dashboards are the default because demand, capacity, cost, availability, inventory, and routing information are distributed across periods. In \textit{Routing and Tour Optimization}, map views are paired with demand, time-window, or fleet tables when node geometry alone is insufficient. In \textit{Pure Combinatorial and Logical Models}, matrices and tables are the natural primary view because the instance is defined by adjacency, incidence, resource, value, or clause structure.

\textbf{Solver-safe disclosure.} Across all families, visuals expose only the public instance specification, not the optimizer's decision output. Thus, a network diagram may show capacities and costs but not optimal arc flows; a location layout may show coverage and opening costs but not selected facilities; a scheduling view may show operation requirements but not start times or machine sequences; a routing map may show customers and demands but not vehicle routes; and a combinatorial matrix may show conflicts or clauses but not the selected subset, coloring, or assignment. This policy makes the visual artifact information-complete for formulation while preventing shortcut evaluation through leaked solutions.

\textbf{Readability controls.} Visual readability is treated as part of benchmark validity rather than as a cosmetic detail. In \textit{Network Optimization}, graph density, edge crossings, node spacing, and arc-label placement are controlled so that topology and coefficients remain legible. In \textit{Location, Covering, and Assignment}, marker overlap, coverage-circle clutter, coordinate labels, and incidence-table width are controlled to keep feasible relations inspectable. In \textit{Scheduling and Sequencing}, operation blocks, precedence layers, stage diagrams, and resource panels are sized so that temporal order and resource requirements can be read without inferring a solved schedule. In \textit{Multi-Period and System Planning}, timelines, heatmaps, tables, and dashboards are aligned by period and split across panels when needed, since cross-period consistency is central to the formulation. In \textit{Routing and Tour Optimization}, complete graphs are usually omitted and map annotations are kept sparse, with demand, time-window, or fleet data moved to companion tables when geometry alone would become crowded. In \textit{Pure Combinatorial and Logical Models}, dense matrices are split or resized while preserving exact entries, row/column labels, and value or capacity bars. These controls reduce the risk that model failure reflects rendering noise rather than multimodal optimization-modeling capability.

Together, these principles make the visual modality a controlled part of the benchmark specification. Each figure or panel is intended to be structurally natural, information-complete for modeling, solver-safe with respect to the hidden solution, and legible enough that performance reflects multimodal optimization modeling rather than accidental rendering artifacts.


\section{Shared Benchmark Construction and Validation Protocol}
\label{sec:appendix_construction}

This appendix collects the \emph{shared} construction contract underlying all MM-OptBench instances. Unlike App.~\ref{app:data_generation_pipeline}, which gives family-specific generation recipes, this appendix focuses on the common guarantees that every released instance must satisfy: a clear artifact boundary, a single verified source of truth, solver-aligned references, controlled difficulty regimes, and validation against structural and multimodal inconsistencies. We start with the ground-truth artifacts and file organization (App.~\ref{subsec:appendix_format}), then give the shared construction protocol that produces them (App.~\ref{subsec:appendix_pipeline}). We next describe configuration sampling and difficulty regimes (App.~\ref{subsec:appendix_sampling}), reference solver alignment (App.~\ref{subsec:appendix_solver_templates}), validation and quality-control checks (App.~\ref{subsec:appendix_qa}), and expert-guided release auditing (App.~\ref{subsec:expert_roles}). We close with four full-size illustrative instances corresponding to Figure~\ref{fig:MLLMOPT_Examples}, showing how textual and visual evidence are paired in the released benchmark (App.~\ref{subsec:appendix_examples}).

\subsection{Ground-Truth Artifacts and File Organization}
\label{subsec:appendix_format}

This subsection defines the released file contract for each MM-OptBench instance. We organize the discussion around three boundaries: the files visible to evaluated models, the files reserved for verification and auditing, and the metadata and provenance records that make the package reproducible.

\subsubsection{Package Layout}

Each MM-OptBench instance is released as a self-contained directory. The official model-facing input consists of \texttt{task\_input.txt} together with the rendered visual file(s) under \texttt{visuals/}. The remaining files are hidden during evaluation and are used for auditing, reference solving, reproducibility checks, and post-hoc analysis.

\begin{center}
\fbox{
\begin{minipage}{0.90\linewidth}
\small\ttfamily
instance\_id/\\
\hspace*{1em}canonical\_specification.txt\\
\hspace*{1em}task\_input.txt\\
\hspace*{1em}visuals/\\
\hspace*{2em}generate\_visual.py\\
\hspace*{2em}visual\_0.png\\
\hspace*{2em}visual\_1.png\ \ (optional)\\
\hspace*{1em}ground\_truth/\\
\hspace*{2em}instance\_data.json\\
\hspace*{2em}math\_model.md\\
\hspace*{2em}solver\_ref.py\\
\hspace*{2em}solution\_ref.json\\
\hspace*{1em}meta.json
\end{minipage}}
\end{center}

\subsubsection{Model-Facing Inputs}

The file \texttt{task\_input.txt} contains the textual component presented to the evaluated model. It states the modeling objective, structural rules, and expected outputs, while leaving essential instance-specific information to be read from the accompanying visual artifact(s). The folder \texttt{visuals/} stores the rendered image inputs, such as graph diagrams, spatial layouts, coverage regions, operation charts, dashboards, or matrices, depending on the problem family. These images encode public instance information needed for modeling, including quantities such as topology, capacities, costs, coordinates, coverage relations, machine assignments, processing times, compatibility relations, or demand profiles. The rendering script \texttt{generate\_visual.py} is retained for provenance, but the official evaluation input is the rendered visual artifact rather than the script.

\subsubsection{Reference and Audit Artifacts}

The file \texttt{canonical\_specification.txt} contains a complete, human-readable audit specification of the fully instantiated problem. It records the public entities, numerical parameters, objective, variant-specific structural rules, and, when useful, notes on how the visual should be interpreted. It is not the model-facing prompt; instead, it provides an authoritative record for human inspection.

The hidden \texttt{ground\_truth/} directory contains the machine-readable and executable references. The file \texttt{instance\_data.json} stores the public instance data in structured form, including basic sets, indices, numerical parameters, and family-specific derived structures such as feasible assignment relations, adjacency matrices, routing patterns, operation-machine mappings, or cost tables when these are part of the public instance definition. The file \texttt{math\_model.md} documents the canonical mathematical formulation, typically including sets, parameters, decision variables, objective, constraints, and instance-specific notes. The file \texttt{solver\_ref.py} provides a verified reference implementation for the exact instance, and \texttt{solution\_ref.json} stores the verified optimum and family-specific reference outputs such as objective values, flows, cuts, assignments, opened facilities, tours, colorings, or schedules.

\subsubsection{Reproducibility and Artifact Alignment}

The file \texttt{meta.json} records provenance and benchmark descriptors for the instance, such as instance identifier, family or variant, difficulty label, structural regime tags, formulation-size statistics, random seed, solver/runtime information, and visual checksums when available. These fields support reproducibility, stratified analysis, and dataset auditing.

All files in an instance package are derived from the same verified instance specification rather than authored independently. This shared source of truth keeps the model-facing text, visual artifact(s), canonical formulation, executable reference, recorded solution, and metadata mutually aligned. It also supports two complementary forms of inspection: formulation-level comparison through \texttt{math\_model.md} and execution-level verification through \texttt{solver\_ref.py} and \texttt{solution\_ref.json}. Detailed validation and quality-control checks used to retain only benchmark-consistent instances are described in App.~\ref{subsec:appendix_qa}.

Overall, the artifact package is designed to separate the model-facing problem specification from the hidden reference record while keeping both derived from the same source of truth. This separation supports fair evaluation, execution-based scoring, and post-hoc inspection of modeling failures.

\subsection{Shared Construction Protocol}
\label{subsec:appendix_pipeline}

This subsection specifies the construction protocol common to all MM-OptBench families. The family-specific generators differ in their optimization structure, parameterization, and visual carrier, but they share the same acceptance contract: an instance is released only after its structure, solver behavior, multimodal artifacts, and metadata are mutually consistent. Representative family-specific generators are described in App.~\ref{app:data_generation_pipeline}, and the complete subcategory-level generator descriptions are provided in the online supplementary material.

\textbf{Role of LLM assistance.} After configuration sampling and parameter instantiation, LLM-assisted checks support expert structural validation. They help inspect whether a candidate instance satisfies the intended family semantics, nontriviality conditions, and cross-field consistency requirements before it proceeds to solver verification. These checks can flag suspicious structures, missing relations, or inconsistent parameter bindings for expert adjudication.

\textbf{Shared pipeline stages.} Figure~\ref{fig:benchmark_pipeline_overview} summarizes the unified solver-grounded pipeline. The pipeline should be read as an acceptance protocol rather than as a single monolithic generator: each family implements the stages below with its own structural rules and visual design.
\begin{enumerate}
    \item \textbf{Benchmark guideline specification.} The process begins with a family-level design specification that fixes the target difficulty regime, admissible scale, structural motifs, visual carrier, and readability constraints.
    \item \textbf{Instance configuration sampling.} The generator samples the discrete structure of a candidate instance, such as graph topology, spatial layout, temporal horizon, routing regime, resource pattern, or logical sparsity pattern.
    \item \textbf{Parameter instantiation.} Numerical and categorical parameters are then assigned to the sampled structure, including costs, capacities, demands, processing times, coordinates, compatibility relations, or logical coefficients.
    \item \textbf{Structural validation.} Before solving, the candidate is checked for family-specific validity and nontriviality, such as connectivity, feasible coverage, admissible density, dimensional consistency, or meaningful resource conflict. This is the stage where LLM-assisted review is used to help experts surface possible semantic or structural inconsistencies; any flagged issue is resolved by expert adjudication, rejection, or regeneration before solver verification.
    \item \textbf{Solver-grounded verification.} Structurally valid candidates are translated into a reference optimization model or exact solving procedure. A candidate is retained only if the solver certifies an optimal solution and records a benchmark-consistent objective value together with any family-specific solution object.
    \item \textbf{Semantic artifact construction.} For each verified candidate, the canonical specification, model-facing task text, mathematical formulation, reference solver, verified solution, and metadata are derived from the same verified instance data.
    \item \textbf{Visual rendering and consistency check.} Visual inputs are rendered from the same parameter source used by the reference model. Rendering checks verify label readability, correct annotation, cross-modal consistency, and absence of solution leakage.
    \item \textbf{Quality evaluation and rejection--regeneration.} Final checks assess difficulty alignment, structural diversity, solver consistency, artifact completeness, and multimodal consistency. Candidates that fail any check are rejected and regenerated until the target number of valid instances is collected.
\end{enumerate}

This shared protocol is what makes MM-OptBench scalable without giving up auditability. Each family can instantiate its own mathematical structure, parameter regime, and visual carrier, while the release criteria remain fixed: structural validity, solver verification, aligned artifacts, readable visuals, and explicit rejection--regeneration of failed candidates. The next subsections unpack the main parts of this protocol: configuration sampling and difficulty design (Stages~1--3, with Stage~8 checks), solver and artifact alignment (Stages~5--6), validation and quality control (Stages~4--8), and expert-guided release auditing around the full acceptance process.

\subsection{Configuration Sampling and Difficulty Regimes}
\label{subsec:appendix_sampling}

This subsection explains how MM-OptBench controls instance diversity and difficulty before solver verification. In the shared protocol of App.~\ref{subsec:appendix_pipeline}, these choices occupy the front of the pipeline: Stage~1 fixes the family-level guidelines, Stage~2 samples the structural configuration, Stage~3 instantiates numerical and categorical parameters, and Stage~8 later checks whether the realized instance still matches the intended difficulty regime. The key idea is that difficulty is sampled through family-specific structural regimes rather than through a single global size rule or arbitrary numerical scaling. Full subcategory-level sampling ranges, parameter grids, and difficulty rules are provided in the online supplementary material.

\textbf{Structural difficulty (Stages~1--3).} MM-OptBench is constructed from family-specific benchmark design guidelines rather than from a small set of hand-crafted instances. For each problem family, the generator first samples a structural configuration that determines the target difficulty level, admissible scale, and instance regime, and then instantiates the numerical parameters consistent with that configuration. Accordingly, difficulty is defined primarily by \emph{structural} complexity rather than by numerical magnitude alone. Across families, the main control knobs include problem size, variable dimensionality, coupling patterns across entities, temporal or routing structure, integrality requirements, and interaction density. In the released benchmark, these abstract knobs are realized through family-specific regimes. For example, in the minimum-cost-flow family, released instances range from compact layered networks with 6 nodes and 7 arcs to substantially larger instances with 20 nodes and 52 arcs, where additional bypass and coupling arcs create richer interactions. In facility location, difficulty may increase from a small uncapacitated instance with 4 candidate facilities and 5 customers to a capacitated instance with 8 facilities and 17 customers under tighter coverage and capacity margins. In job-shop scheduling, the generator varies the number of jobs and machines together with the routing regime, ranging from compact 3-job/3-machine instances to larger 10-job instances with 6 or 8 machines and pronounced bottleneck-machine effects. In graph coloring, difficulty depends jointly on graph size, density, topology regime, and target chromatic structure; for instance, one released medium instance contains 15 vertices and 51 edges and has a verified chromatic number of 5, yielding a nontrivial color-minimization case.

\textbf{Family-local regimes (Stages~1--3 and Stage~8).} Rather than applying a single global scoring rule, MM-OptBench organizes instances into easy/medium/hard regimes defined within each family’s configuration space. Easy instances typically involve smaller scales and simpler interaction patterns, whereas medium and hard instances introduce denser couplings, richer structural dependencies, tighter feasibility margins, or additional formulation features depending on the family. For example, an easy facility-location instance may only require open-or-not decisions and sparse feasible assignments, while a harder capacitated instance must additionally coordinate opening decisions, assignment feasibility, and facility-capacity utilization. Likewise, a minimum-cost-flow instance becomes harder not merely because capacities or costs are numerically larger, but because the network contains more interacting corridors and cut structures that must be identified correctly from the figure. Formulation-size statistics are recorded as a useful diagnostic signal, but they are not the sole determinant of difficulty. Instead, the difficulty labels are intended to reflect the complexity of constructing the correct optimization model from the multimodal input.

\textbf{Rejection--regeneration (Stages~4--5 and Stage~8).} Sampling is coupled with an iterative rejection--regeneration loop. After configuration sampling and parameter instantiation, each candidate instance undergoes family-specific structural validation, such as connectivity or flow-balance checks for network problems, coverage and capacity screening for location problems, conflict sufficiency checks for scheduling problems, or density and nontriviality checks for graph problems. Structurally valid candidates are then verified by solving the corresponding reference model under a prescribed runtime budget. Only instances whose verified optimum and family-specific invariants match the intended task semantics are retained. For example, a graph-coloring instance targeted at a particular chromatic-number regime is rejected if solver verification shows that the sampled graph does not match that regime; similarly, a facility-location instance is discarded if the sampled geometry appears visually plausible but leads to uncovered customers, invalid feasible-assignment structure, or trivial capacity slack. Final quality checks further ensure visual readability, artifact consistency, and alignment with the target difficulty regime, preventing invalid, trivial, or noisy instances from entering the benchmark.

\textbf{Illustrative configuration (Stages~2--5 and Stage~8).} A representative medium graph-coloring configuration may specify 15 vertices, an overlapping-clique topology regime, and an edge density around 0.49. After the candidate graph is generated, the reference solver computes its chromatic number to verify that the instance matches the intended difficulty regime. In one released medium instance, solver verification certifies that the graph has chromatic number 5; as a result, the instance is retained as a nontrivial color-minimization case that remains structurally interpretable through its adjacency-matrix visual. If solver verification had instead shown a different chromatic number, or if the realized graph density had drifted outside the intended regime, the instance would have been rejected and regenerated. This example illustrates how MM-OptBench controls difficulty through structural configuration and solver-grounded validation jointly, rather than through scale alone.

In summary, the easy/medium/hard labels are produced by controlled structural sampling and then checked by solver-grounded validation. This design keeps difficulty meaningful within each family while still enabling aggregate analysis across the full benchmark.

\subsection{Reference Solver Implementations and Artifact Alignment}
\label{subsec:appendix_solver_templates}

This subsection explains how executable references are used to anchor the benchmark. Within the shared protocol in App.~\ref{subsec:appendix_pipeline}, this is the role of Stage~5, which certifies the candidate by solving it, and Stage~6, which turns the verified instance data into aligned semantic artifacts. The reference solvers are not merely answer generators; they connect the structured instance data, canonical mathematical formulation, verified objective value, and family-specific solution object. Detailed solver-template choices for all subcategories are deferred to the online supplementary material.

\textbf{Executable references (Stage~5).} Each MM-OptBench instance is paired with a verified reference solver implementation that serves as the executable ground-truth counterpart of the canonical formulation. Although the concrete implementation is family-specific, the role of \texttt{solver\_ref.py} is consistent across the benchmark: it reads \texttt{instance\_data.json}, constructs the exact optimization model or equivalent exact solving procedure for the instantiated problem, solves it with the designated reference method, and exports the result to \texttt{solution\_ref.json}. In practice, these reference solvers do not all share a single surface form. For example, medium-level TSP instances in the routing category use a direct instance-level script that solves an MTZ formulation with Gurobi and falls back to Held--Karp dynamic programming when needed; graph-coloring instances in the combinatorial category may use either Gurobi or a SciPy-based exact MILP fallback to compute the chromatic number; and medium-level job-shop instances in the scheduling category use a thin \texttt{solver\_ref.py} wrapper that calls a shared family-level \texttt{solve\_reference} routine defined in the generator.

\textbf{Shared derivation (Stage~6).} The canonical formulation artifact \texttt{math\_model.md} and the executable artifact \texttt{solver\_ref.py} are generated from the same verified instance specification rather than being authored independently. Concretely, both artifacts are derived from the same family-level generation logic together with the instantiated public data stored in \texttt{instance\_data.json}. This shared derivation helps ensure that the mathematical description, the executable solver, and the recorded solution remain mutually consistent. For example, in lot-sizing and production-planning instances from the multi-period category, the same product-period data generate both the structured formulation with inventory-balance, setup-linking, and shared-capacity constraints and the reference MILP solver that returns production, inventory, and setup decisions. In graph-coloring instances, the same graph determines both the formulation written in \texttt{math\_model.md} and the reference solver that computes the chromatic number. In job-shop scheduling, the same operation-machine assignments and processing times are used to generate both the disjunctive scheduling formulation and the corresponding executable reference solver.

\textbf{Alignment role (Stages~5--6).} This alignment is important because MM-OptBench evaluates both symbolic modeling and executable realization. The file \texttt{math\_model.md} provides a structured reference for the intended sets, variables, objective, and constraints, while \texttt{solver\_ref.py} and \texttt{solution\_ref.json} provide an execution-level reference for the verified optimum and family-specific outputs such as tours, colorings, production plans, or schedules. By coupling these artifacts through the same verified instance data and family-specific generation logic, MM-OptBench reduces drift between textual formulations and executable ground truth.

Together, these solver and formulation artifacts make the benchmark auditable at two levels. They provide a symbolic reference for inspecting modeling choices and an execution-level reference for computing official correctness through verified solver behavior.

\subsection{Instance Validation and Quality Control}
\label{subsec:appendix_qa}

This subsection details the checks used to decide whether a generated candidate can enter the released benchmark. In the shared protocol of App.~\ref{subsec:appendix_pipeline}, validation begins with Stage~4 structural checks, continues through Stage~5 solver verification, checks the Stage~6 artifacts for consistency, inspects the Stage~7 visual renderings, and culminates in the Stage~8 accept-or-regenerate decision. The validation stack combines problem-specific structural checks, solver-based verification, and cross-artifact consistency audits. The online supplementary material gives the more granular subcategory-level validation rules that are too detailed to enumerate here.

\textbf{Structural validity (Stage~4).} Every released MM-OptBench instance is retained only after passing a multi-stage validation procedure. First, the instantiated problem itself must satisfy family-specific structural checks before optimization is attempted. Depending on the family, these checks may include connectivity or balance conditions, admissible density ranges, coverage feasibility, dimensional consistency, bounded scale, and nontriviality conditions. For example, in medium-level TSP instances, the generator rejects instances whose city coordinates are too crowded or whose node count falls outside the intended difficulty range; in lot-sizing and production-planning instances, candidates may be discarded before solver execution if prefix-capacity feasibility fails; and in scheduling families, sampled structures are rejected when they do not induce sufficient machine or resource conflicts to make precedence and non-overlap constraints substantively meaningful.

\textbf{Solver verification (Stage~5).} Second, the reference solver must reproduce a benchmark-consistent ground truth. The generated \texttt{solver\_ref.py} is executed automatically, and the resulting \texttt{solution\_ref.json} is checked against both the intended task semantics and family-specific invariants. These checks go beyond verifying that the script runs without errors. Routing instances must return a valid closed tour or route set; for example, a TSP solution is rejected if the tour is not closed or does not visit each city exactly once. Scheduling instances must satisfy precedence and non-overlap relations; in released job-shop instances, the verified schedule is additionally checked to ensure that machine conflicts are genuinely active rather than vacuous. Multi-period planning instances must satisfy inventory-balance, setup-linking, and shared-capacity constraints period by period. In several families, we also enforce nontriviality indicators derived from the verified solution, such as requiring binding resources in bottleneck RCPSP instances, binding capacity periods in inventory-routing or lot-sizing settings, or meaningful constraint activity in combinatorial instances. Only solver-verified instances are retained.

\textbf{Artifact consistency (Stages~6--8).} Third, the released artifact package must be internally consistent. Programmatic checks verify file completeness, including \texttt{task\_input.txt}, \texttt{canonical\_specification.txt}, \texttt{instance\_data.json}, \texttt{math\_model.md}, \texttt{solver\_ref.py}, \texttt{solution\_ref.json}, and the expected visual files. Visuals are rendered from the same parameter source used by the reference model, and their checksums are stored in \texttt{meta.json} to guard against accidental drift; in families with multiple visual artifacts, such as RCPSP or multi-period planning, each released image is checked and recorded separately. Other consistency properties require both automated screening and human inspection. In particular, visual readability, correct annotation, natural parameter presentation, and absence of solution leakage are reviewed because they can depend on layout and semantic judgment rather than only on file schemas. Instances are retained only when the textual specification, visual artifact(s), canonical formulation, and solver outputs agree under this combined automated and expert-audited process.

These validation checks ensure that a released instance is not merely solvable, but also structurally meaningful, visually faithful, and internally consistent. They are the main safeguard against treating rendering artifacts, stale files, or degenerate optimization problems as valid benchmark cases.

\subsection{Expert-Guided Release Auditing and Residual Risks}
\label{subsec:expert_roles}

This subsection clarifies the role of expert judgment in a benchmark that is otherwise generated and checked automatically. Expert review is woven into the shared protocol of App.~\ref{subsec:appendix_pipeline} rather than added as a ninth stage: it shapes the Stage~1 design guidelines, audits recurring issues exposed by Stages~4--7, and informs the Stage~8 accept-or-regenerate decision. The goal is not to replace automation, but to identify the parts of benchmark quality that cannot be reliably certified by solver status or schema checks alone. Additional subcategory-specific generator and audit notes are included in the online supplementary material.

\paragraph{Contributors and resources.}
A small group of authors and research colleagues with optimization experience developed the taxonomy, generators, reference solvers, visual templates, and validation rules. After automated generation and rejection, experts checked visual, cross-modal, and solver consistency. We retrospectively estimate approximately 800--900 person-hours, mainly for generator and validator development and visual and cross-modal quality checks. There were no paid crowd-annotation costs, and no model training was required. Exact reference solving for the 780 retained instances required approximately 0.6 CPU-hours in total. Including rejected candidates, rendering, and regression checks, the construction pipeline remained workstation-scale. Separately, commercial-MLLM evaluation, including pilot experiments, debugging, and reruns, cost approximately US\$2,000 in API credits.

\textbf{Why automation is not enough (cross-stage failure modes).} The family-specific pipelines in Apps.~\ref{subsec:data_pipeline_network}--\ref{subsec:data_pipeline_combinatorial} provide scale, but high-quality benchmark construction still requires expert guidance. A generator can produce instances that are syntactically valid and solver-feasible yet weak as benchmark examples: the visual may be hard to read, the parameter regime may be unnatural, the optimal solution may be nearly forced, or the realized split may overuse one structural template. These failures are not always visible from solver status alone.

\textbf{Expert-defined checks (Stage~1, Stage~8, and audits of Stages~4--7).} Expert involvement enters before and after generation. Before generation, experts define the family-level modeling contract, difficulty regimes, admissible visual carriers, nontriviality requirements, and solution-leakage boundaries. After generation, experts inspect failure cases from solver verification and rendering, review sampled releases for visual and semantic plausibility, and revise generator rules when an automated check misses a recurring issue. This is why MM-OptBench is best understood as \emph{expert-guided automation}: automation supplies breadth and reproducibility, while expert review supplies validity and interpretability.

\textbf{Residual risks (Stages~4--8).} The main remaining risks are infeasible or degenerate candidates that evade early filters, solver-runtime instability, stale files or cross-artifact drift, visual-text mismatches, hidden solution leakage, and insufficient structural diversity within a family. The validation stack in App.~\ref{subsec:appendix_qa} reduces these risks through structural checks, exact solving, artifact consistency checks, visual-readability screening, and manual audits, but it cannot prove that every visual design or parameter regime is optimal. The release process therefore treats quality control as iterative: when audits identify weak layouts, unnatural parameters, or overconcentrated templates, the corresponding generator rules are revised and affected instances are regenerated.

The resulting construction process is therefore neither purely manual nor purely automatic. MM-OptBench uses automation for coverage and reproducibility, while expert-guided release auditing helps preserve semantic quality, visual interpretability, and structural diversity.

\subsection{Four Illustrative Benchmark Instances}
\label{subsec:appendix_examples}

This subsection gives four full-size examples of the multimodal inputs summarized in Figure~\ref{fig:MLLMOPT_Examples}: minimum-cost flow, facility location, job-shop scheduling, and energy dispatch/unit commitment. Each example includes a rendered visual input, the original model-facing task text, and notes on the corresponding canonical formulation.

\paragraph{Example A: Minimum-Cost Flow (Directed Network Visual).}
\label{subsec:appendix_ex_flow}

\begin{figure}[h]
    \centering
    \includegraphics[width=0.77\linewidth]{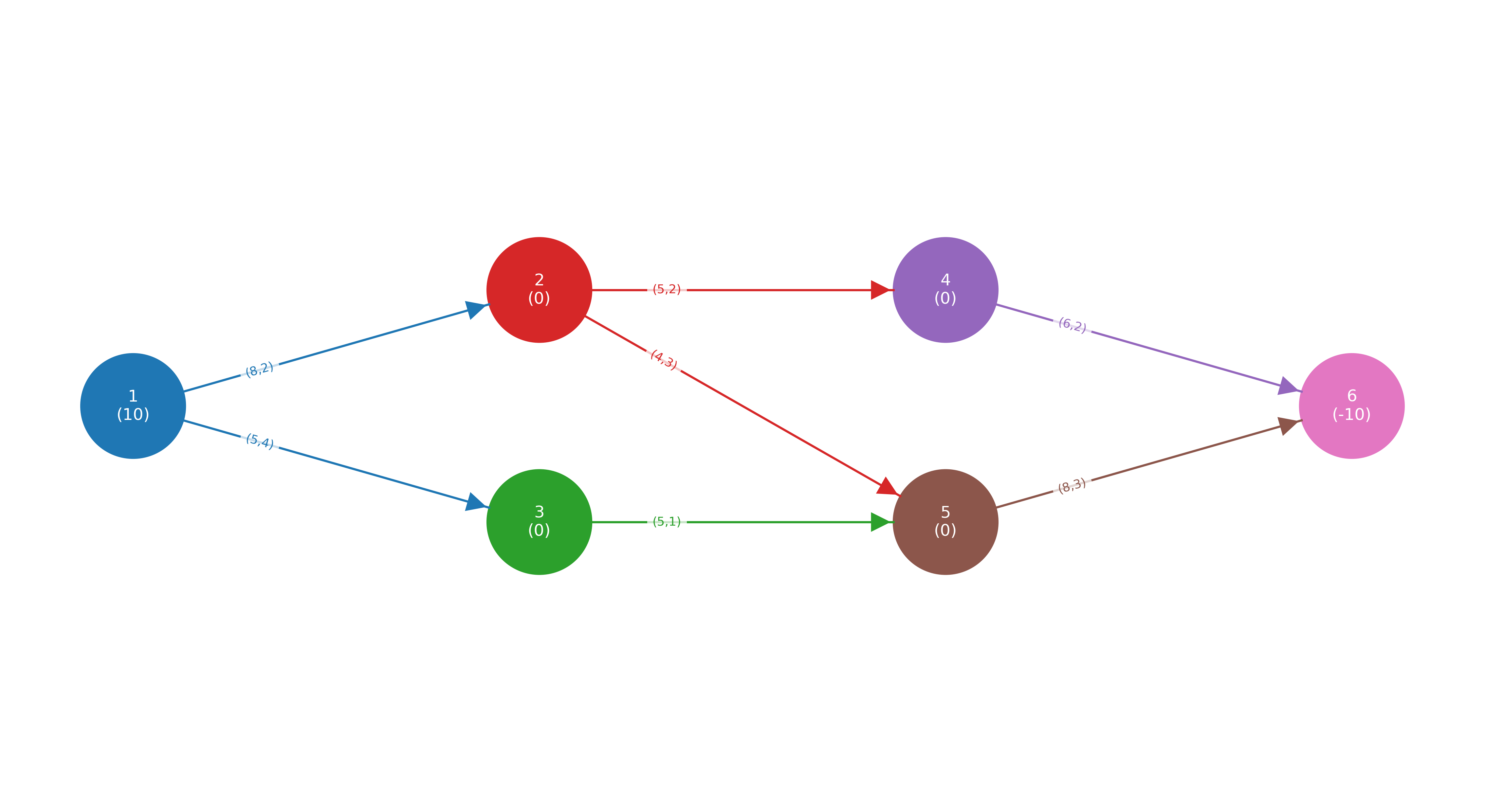}
    \caption{(Example A) Minimum-cost-flow visual input. Node labels encode supplies/demands, and arc labels encode capacities and costs.}
    \label{fig:appendix_flow_graph}
\end{figure}

Figure~\ref{fig:appendix_flow_graph} shows a directed transportation network. Each node is labeled with its supply (positive) or demand (negative), and each directed arc is labeled with a pair \((\text{capacity}, \text{cost})\). Consistent with the network-family construction described in App.~\ref{subsec:data_pipeline_network}, the visual encodes the full public network structure needed to formulate the optimization model, while the text specifies the modeling objective without restating the instance parameters.

\noindent\textbf{Original task text (\texttt{task\_input.txt}).}
\begin{quote}
\small
A directed transportation network is shown in the figure.

Each node is labeled with its supply (positive) or demand (negative).
Each directed arc is labeled with (capacity, cost).

Construct:
(i) the mathematical transportation-flow model, and
(ii) solver-executable code implementing the model.
\end{quote}

\noindent\textbf{Ground-truth notes.}
The corresponding \texttt{math\_model.md} defines a single-commodity minimum-cost flow model. The sets are the node set \(\mathcal{N}\) and directed-arc set \(\mathcal{A}\subseteq\mathcal{N}\times\mathcal{N}\). The only decision variables are nonnegative arc flows \(x_{ij}\), with parameters \(b_i\) for node supply/demand satisfying \(\sum_{i\in\mathcal{N}} b_i=0\), \(u_{ij}\) for arc capacity, and \(c_{ij}\) for unit flow cost. The complete model is
\[
\begin{aligned}
\min \quad & \sum_{(i,j)\in\mathcal{A}} c_{ij}x_{ij} \\
\text{s.t.}\quad
& \sum_{j:(i,j)\in\mathcal{A}} x_{ij} - \sum_{j:(j,i)\in\mathcal{A}} x_{ji} = b_i && \forall i\in\mathcal{N}, \\
& 0 \le x_{ij} \le u_{ij} && \forall (i,j)\in\mathcal{A}.
\end{aligned}
\]
Thus the figure supplies exactly the public data used by the reference formulation: node balances \(b_i\), arc capacities \(u_{ij}\), and arc costs \(c_{ij}\), with all numeric values also recorded in \texttt{instance\_data.json}.

\paragraph{Example B: Facility Location (Spatial Coverage Diagram).}
\label{subsec:appendix_ex_facility}

\begin{figure}[h]
    \centering
    \includegraphics[width=0.78\linewidth]{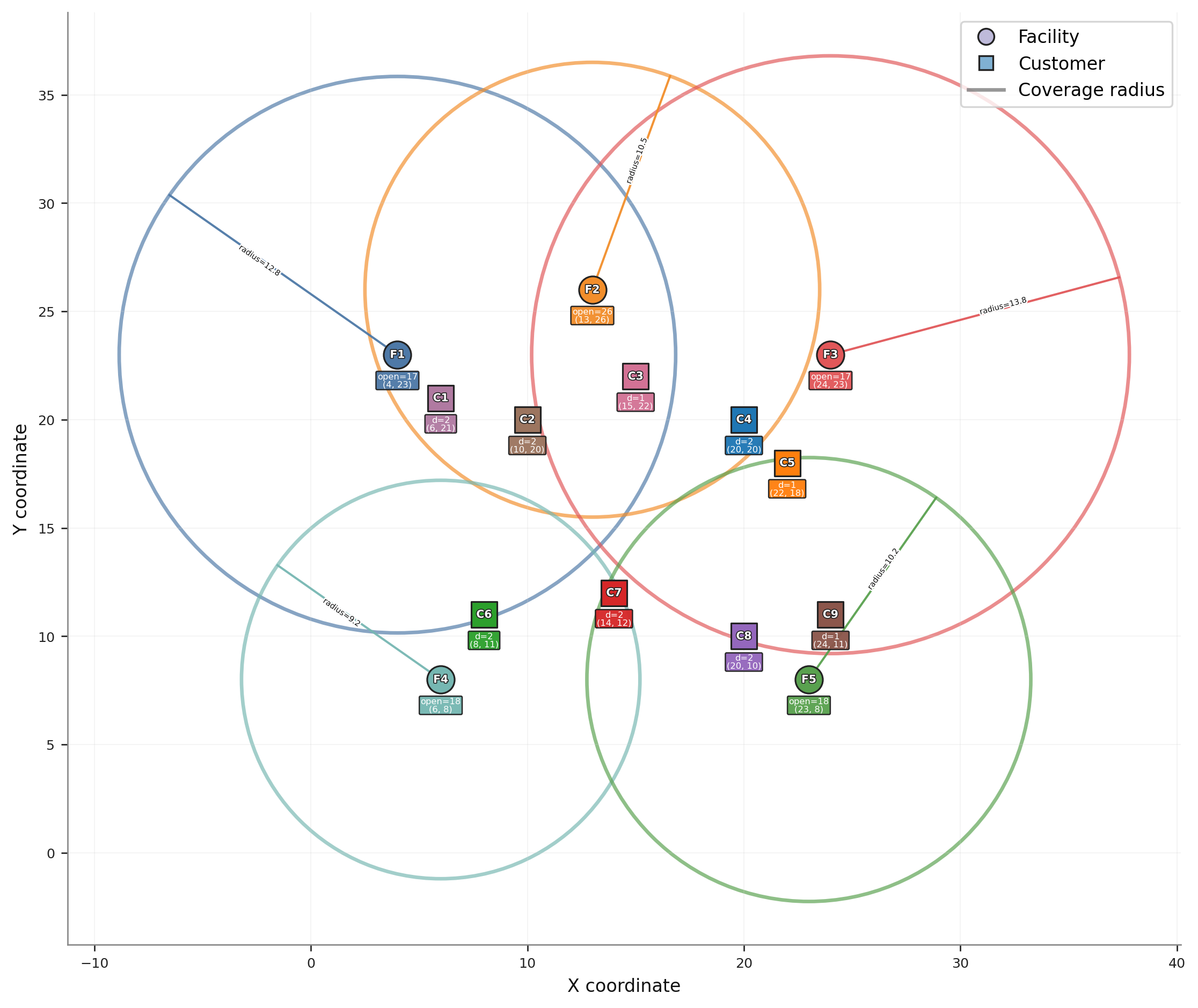}
    \caption{(Example B) Facility-location visual input. Coverage regions and displayed geometry determine feasible assignments and distance-based assignment costs.}
    \label{fig:appendix_facility_diagram}
\end{figure}

Figure~\ref{fig:appendix_facility_diagram} illustrates a facility-location planning instance. Candidate facilities are shown as circles with labels and opening costs, customers are shown as squares with demands, and each facility has a coverage region that determines which assignments are feasible. In the uncapacitated variant illustrated here, assignment cost is induced by the displayed geometry, matching the Euclidean-distance interpretation used in the released benchmark instances.

\noindent\textbf{Original task text (\texttt{task\_input.txt}).}
\begin{quote}
\small
A planning diagram is provided.

- Candidate facilities are shown as circles with their labels and opening costs.

- Customers are shown as squares with their labels and demands.

- Each facility has a coverage circle. A customer may only be assigned to a facility if the customer lies inside that facility's coverage circle.

- Each facility's coverage radius is also shown by a same-colored line segment from the facility center to the coverage circle, labeled with the radius value.

- Assignment cost per unit demand is the Euclidean distance between the customer and the facility, based on the displayed coordinates.

Task:
1) Write a mathematical optimization model that decides which facilities to open and assigns each customer to exactly one open facility that covers it.

2) Provide solver-executable code implementing the model.

Objective: minimize total facility opening cost plus demand-weighted assignment cost.
\end{quote}

\noindent\textbf{Ground-truth notes.}
The corresponding \texttt{math\_model.md} defines this example as a single-period uncapacitated facility-location problem with coverage-based assignment feasibility. The sets are \(F=\{F1,F2,F3,F4\}\) for candidate facilities and \(C=\{C1,C2,C3,C4,C5\}\) for customers. The formulation uses binary opening variables \(y_f\) and binary assignment variables \(x_{cf}\), with opening costs \(f_f\), customer demands \(d_c\), assignment costs \(c_{cf}\), and coverage indicators \(a_{cf}\). The complete model is
\[
\begin{aligned}
\min \quad
& \sum_{f\in F} f_f y_f + \sum_{c\in C}\sum_{f\in F} d_c c_{cf} x_{cf} \\
\text{s.t.}\quad
& \sum_{f\in F} x_{cf} = 1 && \forall c\in C, \\
& x_{cf} \le y_f && \forall c\in C,\; f\in F, \\
& x_{cf} \le a_{cf} && \forall c\in C,\; f\in F, \\
& y_f\in\{0,1\} && \forall f\in F, \\
& x_{cf}\in\{0,1\} && \forall c\in C,\; f\in F.
\end{aligned}
\]
No capacity constraints appear in this UFL instance. Feasible assignments are determined by the coverage geometry encoded in \texttt{instance\_data.json} and reflected in the visual.

\paragraph{Example C: Job-Shop Scheduling (Operation-Layout Visual).}
\label{subsec:appendix_ex_schedule}

\begin{figure}[h]
    \centering
    \includegraphics[width=0.93\linewidth]{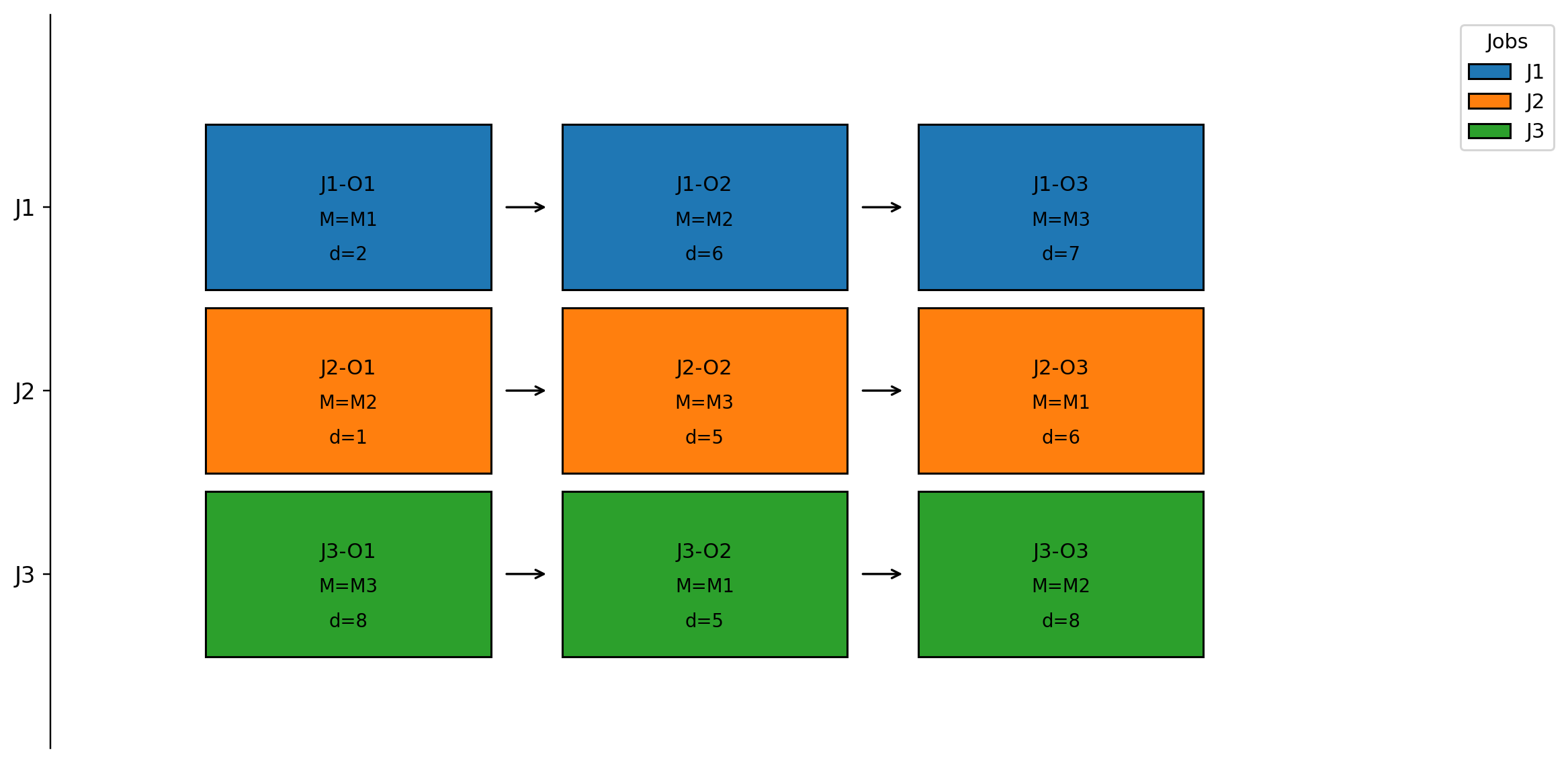}
    \caption{(Example C) Job-shop scheduling visual input. The diagram encodes operation order, machine assignments, and durations, but does not reveal a feasible schedule.}
    \label{fig:appendix_schedule_gantt}
\end{figure}

Figure~\ref{fig:appendix_schedule_gantt} presents a classical job-shop scheduling instance in the released operation-layout visual format. Each row corresponds to a job, and each block is labeled with an operation identifier, its required machine, and its processing duration. The left-to-right order of blocks within a row gives the precedence structure for that job. Importantly, as also emphasized in App.~\ref{subsec:data_pipeline_scheduling}, this visualization is not a traditional Gantt chart: it does not encode a feasible schedule or start times, but rather the structural information from which the schedule must be modeled.

\noindent\textbf{Original task text (\texttt{task\_input.txt}).}
\begin{quote}
\small
A scheduling diagram is provided for a machine-processing scheduling instance.

The visual shows each job as an ordered sequence of operations.
Each operation block is labeled by its job-operation identifier,
required machine, and processing duration.
The left-to-right order of blocks within a row gives the precedence
order for that job.

Interpretation:
- Each job consists of an ordered sequence of operations that must be processed in the listed order.

- Each operation requires the machine shown in its block.

- The duration shown in the block is the processing time of that operation.

- Each machine can process at most one operation at a time.

- Operations are non-preemptive.

Task:
1) Write a mathematical optimization model for this instance.

2) Provide solver-executable code implementing the model.

Objective:
Minimize the makespan.
\end{quote}

\noindent\textbf{Ground-truth notes.}
The corresponding \texttt{math\_model.md} defines a classical job-shop scheduling model with \(J=\{J1,J2,J3\}\), \(M=\{M1,M2,M3\}\), and ordered operation sets \(O_j\). The instance-specific operations are
\[
\begin{aligned}
J1 &: O1(M1,2)\rightarrow O2(M2,6)\rightarrow O3(M3,7),\\
J2 &: O1(M2,1)\rightarrow O2(M3,5)\rightarrow O3(M1,6),\\
J3 &: O1(M3,8)\rightarrow O2(M1,5)\rightarrow O3(M2,8).
\end{aligned}
\]
The formulation uses start-time variables \(s_{j,k}\ge 0\), a makespan variable \(C_{\max}\ge 0\), and binary ordering variables \(y_{(j,k),(j',k')}\) for pairs of distinct operations assigned to the same machine. The complete model is
\[
\begin{aligned}
\min \quad & C_{\max} \\
\text{s.t.}\quad
& s_{j,k+1} \ge s_{j,k} + p_{j,k} && \forall j\in J,\; k=1,\dots,|O_j|-1, \\
& s_{j,k}+p_{j,k} \le s_{j',k'} + H\left(1-y_{(j,k),(j',k')}\right)
&& \forall (j,k)\ne(j',k'):\; m_{j,k}=m_{j',k'}, \\
& s_{j',k'}+p_{j',k'} \le s_{j,k} + H y_{(j,k),(j',k')}
&& \forall (j,k)\ne(j',k'):\; m_{j,k}=m_{j',k'}, \\
& C_{\max} \ge s_{j,k}+p_{j,k} && \forall j\in J,\; k\in O_j, \\
& s_{j,k}\ge 0 && \forall j\in J,\; k\in O_j, \\
& C_{\max}\ge 0, \\
& y_{(j,k),(j',k')}\in\{0,1\}
&& \forall (j,k)\ne(j',k'):\; m_{j,k}=m_{j',k'}.
\end{aligned}
\]
This aligns with the visual: the diagram provides operation order, required machines, and durations, while intentionally withholding start times and any feasible schedule.

\paragraph{Example D: Energy Dispatch and Unit Commitment (Multi-Panel Planning Visual).}
\label{subsec:appendix_ex_energy}

\begin{figure}[h]
    \centering
    \includegraphics[width=0.82\linewidth]{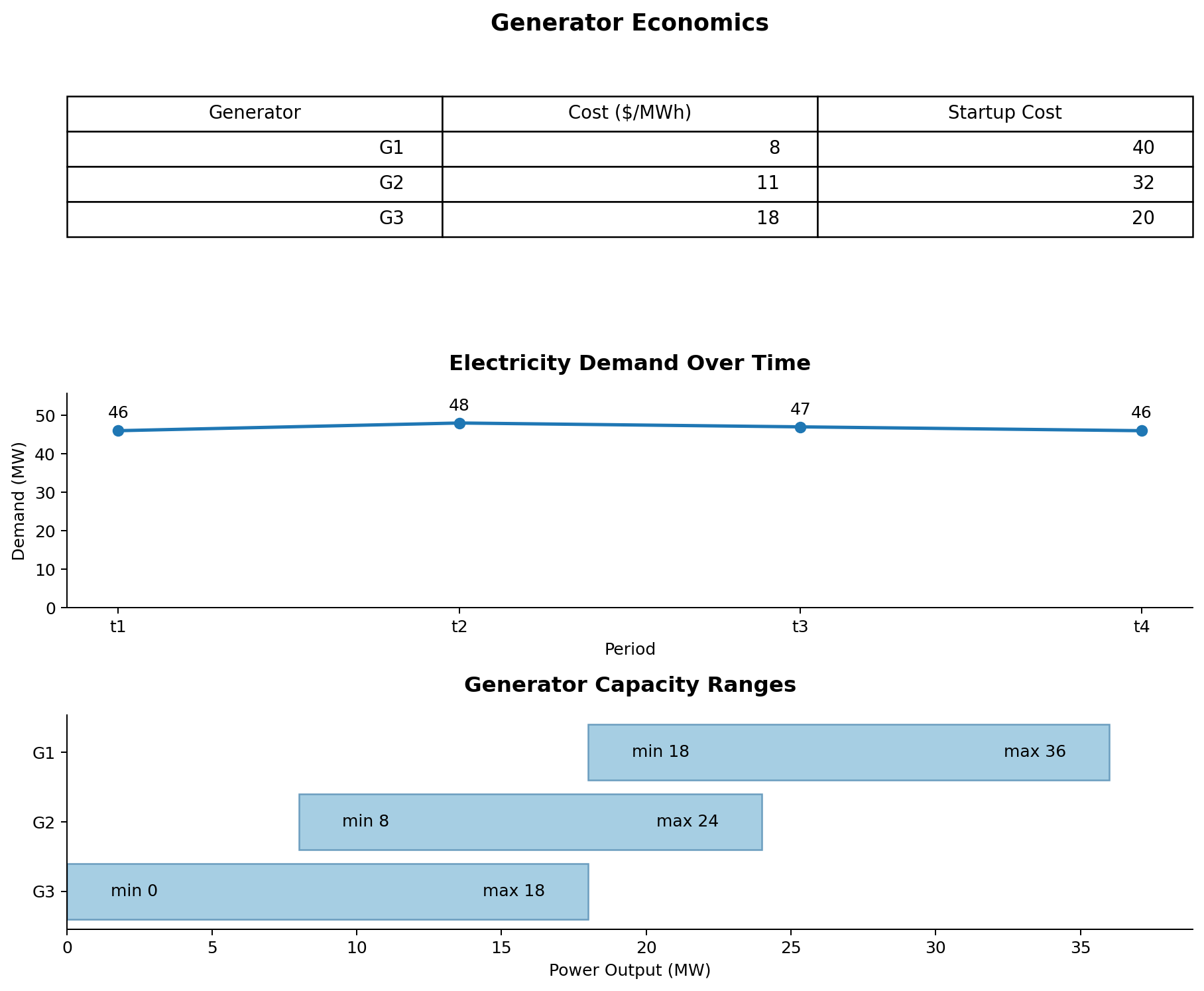}
    \caption{(Example D) Energy-dispatch visual input. Generator economics, period-wise demand, and capacity ranges define a time-coupled unit-commitment and dispatch model without revealing the optimal commitment or dispatch plan.}
    \label{fig:appendix_energy_dispatch}
\end{figure}

Figure~\ref{fig:appendix_energy_dispatch} shows an energy-dispatch and unit-commitment instance in the released multi-panel format. The top panel gives generator economics, the middle panel gives period-wise electricity demand, and the bottom panel gives generator minimum and maximum output ranges. Consistent with the multi-period construction described in App.~\ref{subsec:data_pipeline_multiperiod}, these panels jointly specify the public instance data needed for the optimization model, while omitting the optimal on/off schedule and dispatch quantities.

\noindent\textbf{Original task text (\texttt{task\_input.txt}).}
\begin{quote}
\small
A power-system operations instance is specified by the accompanying visual artifact.
The figure provides all instance-specific numerical data required for the planning problem.

- The top panel is a generator economics table listing each generator, its marginal generation cost, and its startup cost.

- The middle panel is a demand timeline showing the exact system demand in every period.

- The bottom panel is a generator-capacity chart showing the minimum and maximum output range of each generator.

Operational rules:
- Each generator may be either off or on in each period.

- If a generator is on, its output must lie between its minimum and maximum capacity.

- If a generator is off, its output must be zero.

- Total generation in each period must exactly match the demand shown in the figure.

- Turning a generator on in a period incurs its startup cost if it was off in the immediately preceding period.

- The initial on/off status before the first period is: G1=off, G2=off, G3=off.

- No ramping constraints, minimum up-time constraints, minimum down-time constraints, transmission constraints, reserve constraints, load shedding, or stochastic effects are included.

Task:
Build a mathematical optimization model for this exact instance and provide executable solver code.
\end{quote}

\noindent\textbf{Ground-truth notes.}
The corresponding \texttt{math\_model.md} defines a deterministic single-zone energy dispatch and unit-commitment model. The sets are \(G=\{G1,G2,G3\}\) and \(T=\{t1,t2,t3,t4\}\). The formulation uses continuous generation variables \(p_{g,t}\ge 0\), binary commitment variables \(u_{g,t}\), and binary startup variables \(v_{g,t}\), with demand \(D_t\), marginal cost \(c_g\), startup cost \(f_g\), output limits \(P^{\min}_g,P^{\max}_g\), and initial status \(u^0_g\). The complete model is
\[
\begin{aligned}
\min \quad & \sum_{g\in G}\sum_{t\in T}\left(c_g p_{g,t}+f_g v_{g,t}\right) \\
\text{s.t.}\quad
& \sum_{g\in G}p_{g,t}=D_t && \forall t\in T, \\
& p_{g,t}\ge P^{\min}_g u_{g,t} && \forall g\in G,\; t\in T, \\
& p_{g,t}\le P^{\max}_g u_{g,t} && \forall g\in G,\; t\in T, \\
& v_{g,t_1}\ge u_{g,t_1}-u^0_g && \forall g\in G, \\
& v_{g,t}\ge u_{g,t}-u_{g,t^-} && \forall g\in G,\; t\in T\setminus\{t_1\}, \\
& p_{g,t}\ge 0 && \forall g\in G,\; t\in T, \\
& u_{g,t}\in\{0,1\},\; v_{g,t}\in\{0,1\} && \forall g\in G,\; t\in T.
\end{aligned}
\]
Here \(t^-\) denotes the period immediately preceding \(t\). The easy-level instance excludes ramping limits, minimum up/down constraints, transmission constraints, reserve requirements, and stochastic effects, matching the original \texttt{math\_model.md}. The visual panels provide all numeric data needed for this time-indexed model.

These examples show the intended MM-OptBench interaction pattern at instance level: the text states the modeling task and output contract, while the visual supplies essential public data that must be translated into variables, parameters, constraints, objectives, and executable solver logic. The full benchmark repeats this pattern across all six families and multiple difficulty regimes.

\section{Representative Family-Specific Data Generation Pipelines}
\label{app:data_generation_pipeline}

This appendix gives representative family-specific construction examples rather than the full set of subcategory pipelines. To keep the paper appendix readable, we include one subcategory from each major optimization family: minimum-cost flow for \textit{Network Optimization} (App.~\ref{subsec:pipeline_mcf_example}), facility location for \textit{Location, Covering, and Assignment} (App.~\ref{subsec:pipeline_facility_example}), job-shop scheduling for \textit{Scheduling and Sequencing} (App.~\ref{subsec:pipeline_jss_example}), energy dispatch and unit commitment for \textit{Multi-Period and System Planning} (App.~\ref{subsec:pipeline_educ_example}), the traveling salesman problem (TSP) for \textit{Routing and Tour Optimization} (App.~\ref{subsec:pipeline_tsp_example}), and graph coloring for \textit{Pure Combinatorial and Logical Models} (App.~\ref{subsec:pipeline_gc_example}). Complete subcategory-level construction descriptions for all 26 benchmark subcategories are provided in the online supplementary material. The shared protocol and validation logic are described in Apps.~\ref{subsec:appendix_pipeline} and~\ref{subsec:appendix_qa}; the examples below retain the main family-specific details needed to understand how structure is sampled, how invalid candidates are rejected, how the reference optimum is verified, and how the visual artifact exposes instance data without revealing the solution.
\subsection{Data Generation Pipelines for Network Optimization}
\label{subsec:data_pipeline_network}

\label{subsec:pipeline_mcf_example}

The \textit{Network Optimization} family contains five subcategories: minimum-cost flow, maximum flow, resource-constrained shortest path, capacitated network design, and time-expanded multi-period network flow. We use minimum-cost flow as the representative example here because it exposes the core visual structure of this family---nodes, directed arcs, capacities, costs, and flow-balance semantics---in a compact network diagram. Full construction details for the other network subcategories are provided in the online supplementary material.

\begin{figure}[h]
    \centering
    \includegraphics[width=1\linewidth]{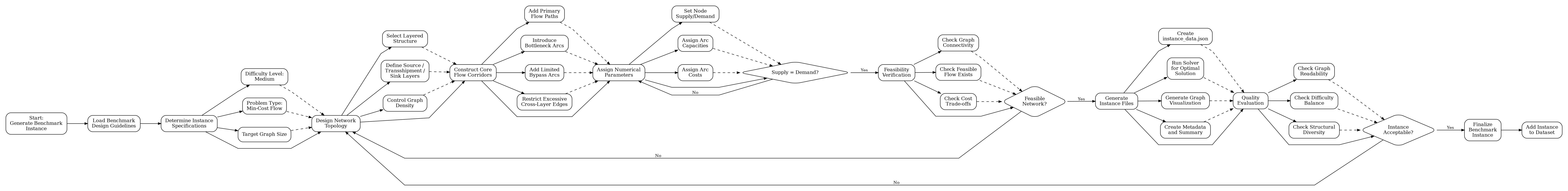}
    \caption{Representative minimum-cost-flow construction pipeline. Candidate networks are generated from structural guidelines, assigned supplies, demands, capacities, and costs, verified for feasibility and optimality, and retained only after visual and artifact-level quality checks.}
    \label{fig:generation_pipeline_mcf}
\end{figure}

The minimum-cost-flow pipeline represents the network family, where the visual artifact is itself part of the optimization specification. Nodes encode supplies and demands, directed arcs encode admissible movement, and arc labels encode the capacity and unit transportation cost that determine the feasible region and objective. As shown in Figure~\ref{fig:generation_pipeline_mcf}, generation begins from difficulty-specific guidelines that specify the admissible number of nodes and arcs, the source--transshipment--sink layout, and the amount of structural interaction expected in the network.

The topology is not sampled as an arbitrary dense graph. The generator first builds a readable layered backbone so that every supply node has a route toward the demand side. It then adds controlled bypass arcs, cross-layer arcs, or coupling arcs when the target difficulty calls for stronger interaction among routes. This construction creates alternatives that are visually interpretable but not reducible to a single obvious path. Overly dense graphs, disconnected graphs, and layouts with ambiguous edge crossings are rejected before numerical parameters are assigned.

After the graph structure is fixed, the generator assigns node balances, arc capacities, and arc costs. Balance checks enforce equality between total supply and total demand, while capacity checks ensure that the network can support at least one feasible shipment plan. Cost regimes are also screened: instances in which one route dominates all alternatives, or in which costs make the objective essentially irrelevant, are discarded. These checks are important because a visually valid network may still be too trivial as an optimization-modeling instance.

The verified candidate is then solved with a reference minimum-cost-flow model. The solver output is accepted only when flow conservation holds at all transshipment nodes, arc flows satisfy capacity bounds, all supply and demand requirements are met, and the returned objective is certified optimal. Accepted candidates are materialized into the aligned benchmark artifacts, including \texttt{instance\_data.json}, \texttt{math\_model.md}, \texttt{solver\_ref.py}, \texttt{solution\_ref.json}, task text, and the rendered network visual. The visual contains all public data needed to formulate the model, but it does not show the optimal flow values. A final audit checks agreement between the visual labels and the structured instance record, label readability, and consistency with the target difficulty level.

\subsection{Data Generation Pipelines for Location, Covering, and Assignment}
\label{subsec:data_pipeline_location}

\label{subsec:pipeline_facility_example}

The \textit{Location, Covering, and Assignment} family contains facility location, \(p\)-median/\(p\)-center, set covering/partitioning, bipartite assignment and matching, and generalized assignment. We use facility location as the representative example because it combines the main modeling motifs of the family: open-or-not decisions, assignment variables, geometric feasibility, and cost information induced by spatial layout. The remaining subcategory-level pipelines are included in the online supplementary material.

\begin{figure}[h]
    \centering
    \includegraphics[width=1\linewidth]{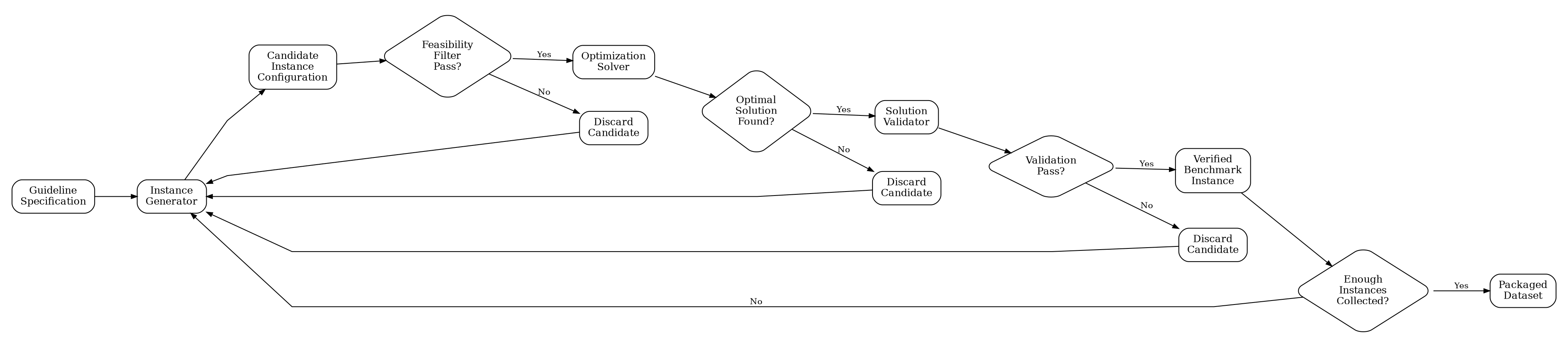}
    \caption{Representative facility-location construction pipeline. Spatial layouts and coverage relations are generated from guidelines, screened for feasibility, solved with a reference optimizer, and retained only after assignment, coverage, and visual-consistency checks.}
    \label{fig:generation_pipeline_fl}
\end{figure}

Facility location is representative of problems in which spatial information determines both feasible assignments and economic trade-offs. The generator begins with difficulty-specific guidelines that define the number of candidate facilities, the number of customers, the admissible coordinate range, the coverage-radius regime, and the cost or capacity variants activated in the instance. It then samples a spatial configuration containing candidate facility sites, customer locations, opening costs, customer demands, and, when used, facility capacities.

The spatial layout is screened before optimization. The generator checks that each customer has at least one feasible facility within coverage, that the coverage graph is not so dense that all facilities become interchangeable, and that the layout remains readable after labels and coverage circles are drawn. In capacitated variants, aggregate capacity and local capacity pressure are checked jointly so that infeasibility is avoided without making capacity constraints vacuous. Candidates that fail these tests are discarded before solver time is spent.

For candidates that pass screening, the reference model selects open facilities and assigns each customer to an open feasible facility. The solver output is validated against the intended formulation: every customer must be assigned, assignments must respect coverage and opening decisions, capacities must be respected when present, and the objective must combine fixed opening costs and assignment costs according to the subcategory definition. The pipeline also rejects instances whose optimal solution is structurally uninformative, for example when all facilities must open or when a single facility trivially covers all customers at lowest cost.

The accepted instance is then rendered as a spatial visual containing the public information required for modeling: facility and customer markers, coverage regions, coordinates or labels, and the relevant cost or demand annotations. The visualization intentionally omits assignment edges and open/closed decisions, because those are part of the solution rather than the input specification. The final consistency audit checks that every coverage relation shown in the figure matches \texttt{instance\_data.json}, that numerical annotations agree with the reference artifacts, and that the coverage geometry remains legible at the benchmark scale.

\subsection{Data Generation Pipelines for Scheduling and Sequencing}
\label{subsec:data_pipeline_scheduling}

\label{subsec:pipeline_jss_example}

The \textit{Scheduling and Sequencing} family contains job-shop scheduling, parallel-machine scheduling, resource-constrained project scheduling, and flexible/hybrid flow-shop scheduling. We select job-shop scheduling as the representative example because it makes the interaction between visual structure and optimization formulation especially explicit: the figure specifies operation order, machine requirements, and processing durations, while the solver model must introduce precedence and non-overlap constraints. Complete pipelines for the remaining scheduling subcategories are provided in the online supplementary material.

\begin{figure}[h]
    \centering
    \includegraphics[width=1\linewidth]{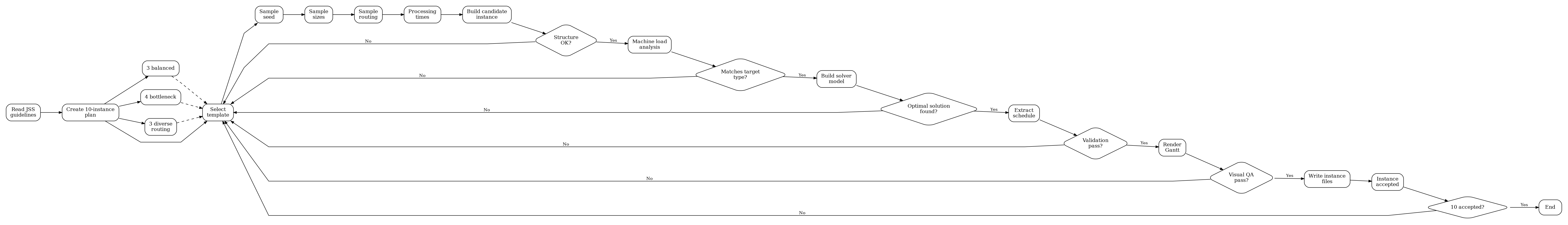}
    \caption{Representative job-shop-scheduling construction pipeline. Candidate job routings and processing times are sampled under difficulty guidelines, checked for meaningful machine conflicts, verified by an exact scheduler, and rendered as operation-layout visuals that do not reveal a solved schedule.}
\label{fig:generation_pipeline_jss}
\end{figure}

Job-shop scheduling illustrates a case where the visual artifact encodes precedence and resource requirements but intentionally omits the solved schedule. The generator starts from guidelines that specify the number of jobs, number of machines, number of operations per job, routing-pattern regime, and processing-time regime. It then constructs one ordered machine route for each job, so that the visual row for a job corresponds to the required operation sequence rather than to a scheduled timeline.

The routing structure is checked before processing times are assigned. Each job route must be well formed, machine identifiers must be valid, and machine usage must create meaningful contention across jobs. Candidates are rejected when they decompose into nearly independent chains, when one machine never interacts with the others, or when the routing pattern would make the non-overlap constraints irrelevant. This stage is what turns the visual layout from a list of operations into a genuine scheduling instance.

Processing times are then generated under the selected duration regime. Balanced regimes produce moderate congestion across machines, heterogeneous regimes create uneven operation lengths, and bottleneck regimes make selected machines or job segments especially influential for makespan. The reference scheduler verifies that the instance admits a feasible schedule and computes the optimal makespan. The solution is accepted only if job precedences are respected, machine-capacity conflicts are resolved through non-overlap, and the objective value recorded in \texttt{solution\_ref.json} matches the certified optimum.

The released visual is an operation-layout diagram rather than a Gantt chart. Each row corresponds to a job, each block corresponds to an operation, and the block text gives the machine requirement and processing duration. It does not display start times, completion times, machine timelines, or an optimal operation order on each machine. The final audit checks that the operation order, machine labels, durations, and row layout match the structured instance record and that the figure remains readable across difficulty levels.

\subsection{Data Generation Pipelines for Multi-Period and System Planning}
\label{subsec:data_pipeline_multiperiod}

\label{subsec:pipeline_educ_example}

The \textit{Multi-Period and System Planning} family contains lot-sizing and production planning, energy dispatch and unit commitment, workforce planning and shift scheduling, and inventory routing. We use energy dispatch and unit commitment as the representative example because it illustrates the multi-panel nature of this family: generator economics, temporal demand, and capacity ranges must be integrated into one time-coupled optimization model. Full construction details for the other planning subcategories are provided in the online supplementary material.

\begin{figure}[h]
    \centering
    \includegraphics[width=1\linewidth]{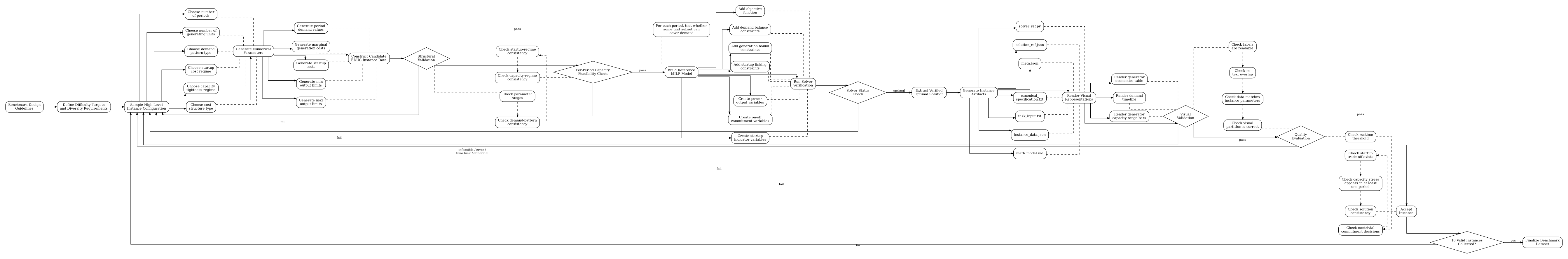}
    \caption{Representative energy-dispatch and unit-commitment construction pipeline. Candidate planning instances combine generator economics, temporal demand, and capacity ranges; verified instances are retained only after dispatch feasibility, solver optimality, and multi-panel visual consistency checks.}
\label{fig:generation_pipeline_educ}
\end{figure}

Energy dispatch and unit commitment is representative of multi-period planning tasks whose instance data are distributed across coordinated panels. The guidelines specify the number of generating units, planning horizon length, temporal demand regime, marginal generation costs, startup costs, and minimum/maximum generation ranges. These choices determine both the economic structure of the objective and the time-coupled feasibility structure of the commitment decisions.

The generator samples unit-level and period-level parameters jointly. Demand may be increasing, peaked, cyclic, or mildly irregular; generator costs may create a merit order; startup costs may make frequent switching undesirable; and capacity ranges may create tight or loose reserve margins. Before solving, the pipeline checks that aggregate capacity can meet demand in every period and that the instance is not degenerate, such as a case where one unit is always sufficient or all units must be on in every period regardless of costs.

Verified candidates are solved using a reference unit-commitment and economic-dispatch model. The solver must produce commitment and dispatch decisions satisfying period-wise demand balance, generator minimum and maximum output limits, and the startup accounting used in the benchmark formulation. The objective combines generation cost and startup cost, and the recorded value is accepted only after feasibility and optimality checks. This verification is especially important because visually plausible dispatch panels can still hide infeasible capacity combinations or trivial commitment structure.

The visual carrier is deliberately multi-panel. One panel summarizes generator economics, a second panel plots demand over the planning horizon, and a third panel displays generator capacity ranges. Together these panels provide the public data needed to build the model, while omitting the optimal on/off schedule and dispatch quantities. The final audit checks cross-panel consistency with \texttt{instance\_data.json}, alignment with the reference formulation, label readability, and absence of solution leakage.

\subsection{Data Generation Pipelines for Routing and Tour Optimization}
\label{subsec:data_pipeline_routing}

\label{subsec:pipeline_tsp_example}

The \textit{Routing and Tour Optimization} family contains the traveling salesman problem, vehicle routing, traveling salesman with time windows, and heterogeneous-fleet vehicle routing. We use TSP as the representative example because it isolates the central route-construction requirement of the family: the model must convert a map-like coordinate visual into a globally connected tour without being shown the complete graph or the optimal route. The remaining routing pipelines are given in the online supplementary material.

\begin{figure}[h]
    \centering
    \includegraphics[width=1\linewidth]{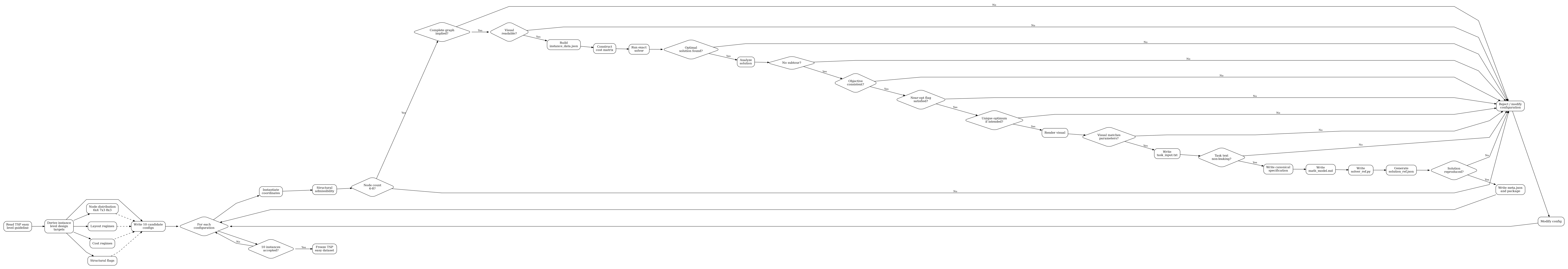}
    \caption{Representative TSP construction pipeline. City coordinates and distance regimes are sampled under spatial-layout guidelines, verified by an exact TSP solver, and rendered as map-like inputs that show locations but not the optimal tour.}
\label{fig:generation_pipeline_tsp}
\end{figure}

The TSP pipeline represents routing and tour-optimization instances where the visual input is a map-like coordinate layout. Guidelines determine the number of cities, the admissible spatial regime, and the distance convention used to construct travel costs. Candidate layouts include uniform point sets, clustered regions, perturbed circular or grid-like patterns, and mild outlier configurations. Layouts that are nearly collinear, excessively crowded, or visually ambiguous are rejected because they reduce either routing difficulty or human readability.

Once the coordinate layout is accepted, the generator constructs the pairwise travel-cost matrix from the specified Euclidean or rounded-distance rule. This matrix is the optimization input used by the reference solver and is also the latent structure that a model must infer from the visual coordinates. The pipeline checks that the cost matrix is symmetric when the variant requires symmetry, that diagonal entries and distance rounding are handled consistently, and that the resulting instance is not dominated by a trivial nearest-neighbor pattern.

The reference solver then computes an optimal Hamiltonian cycle. Verification checks that every city is visited exactly once, that the route returns to the start, that subtours are eliminated, and that the objective value recorded in the released solution file is the certified optimum. Candidates that fail solver verification or produce unstable objective records are regenerated.

The released visual shows the city identifiers and coordinate locations but omits the complete graph and the optimal tour. This choice keeps the figure inspectable and prevents solution leakage, while still requiring the model to construct the complete pairwise routing structure needed for a solver-executable TSP formulation. The final quality pass checks visual spacing, label placement, coordinate consistency, and agreement among the visual, \texttt{instance\_data.json}, reference solver, and metadata.

\subsection{Data Generation Pipelines for Pure Combinatorial and Logical Models}
\label{subsec:data_pipeline_combinatorial}

\label{subsec:pipeline_gc_example}

The \textit{Pure Combinatorial and Logical Models} family contains multi-dimensional knapsack, graph coloring, set packing, and logical constraint satisfaction. We use graph coloring as the representative example because it tests purely discrete structure through an adjacency-matrix visual and requires solver verification of the intended chromatic-number regime. The complete pipelines for the other combinatorial and logical subcategories are provided in the online supplementary material.

\begin{figure}[h]
    \centering
    \includegraphics[width=1\linewidth]{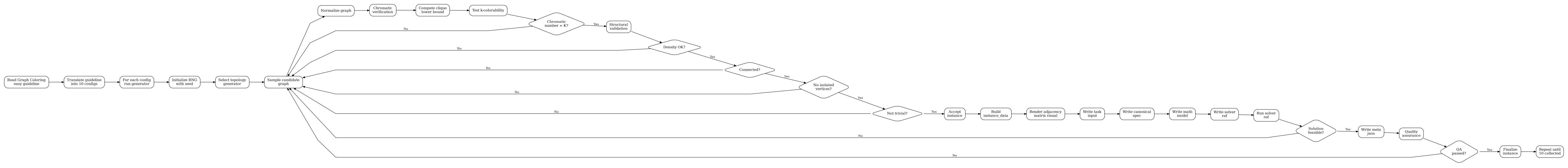}
    \caption{Representative graph-coloring construction pipeline. Candidate graphs are sampled under size, density, topology, and target-chromatic-number guidelines, verified by an exact coloring solver, and rendered as adjacency-matrix visuals.}
\label{fig:generation_pipeline_gc}
\end{figure}

Graph coloring is included as the representative pure combinatorial example because its difficulty is controlled by discrete structure rather than geometry or time. Guidelines specify the number of vertices, the available color budget, the admissible edge-density range, the topology regime, and the target chromatic-number regime. These choices allow the generator to control both visual scale and combinatorial hardness.

Candidate graphs are constructed from several topology families. Bipartite-like graphs provide low-chromatic baselines; grid-based graphs create sparse structured adjacency; odd-cycle augmentations introduce small but decisive chromatic obstructions; and planted-coloring regimes generate graphs with randomized cross-group edges while preserving a known target structure. The generator uses these regimes to avoid a dataset consisting only of random graphs, since such a dataset would be less interpretable visually and less diagnostic of modeling behavior.

Each candidate is verified through structural and solver-based checks. Structural validation removes disconnected graphs, isolated vertices, density outliers, and graphs whose adjacency patterns are either too sparse to be meaningful or too dense to be visually interpretable. Solver verification computes the chromatic number through the reference coloring model and confirms that it matches the intended target regime. The candidate is accepted only when the graph is colorable under the benchmark specification and the verified chromatic behavior agrees with the generated metadata.

Accepted instances are packaged with the canonical coloring formulation and rendered as adjacency-matrix visuals. Rows and columns index vertices, and marked cells indicate adjacency relations. This representation exposes the exact graph needed to formulate coloring constraints, but it does not reveal any feasible or optimal color assignment. The final audit checks symmetry of the matrix for undirected graphs, consistency between marked cells and \texttt{instance\_data.json}, readability of row and column labels, and agreement with the reference solution record.

\section{Experimental Protocol, Extended Results, and Diagnostics}
\label{app:extended_experiment_results}

This appendix collects the experimental protocol, extended results, and diagnostic analyses behind Section~\ref{sec:experiments}. We first give the setup and prompt contracts needed to interpret the research questions in the main text (App.~\ref{subsec:app_exp_setup_details}), then report the extended performance tables and pass@4 heatmap supporting the overall, family-level, and difficulty-level results (App.~\ref{subsec:app_extended_performance_tables}). We next define the failure taxonomy used for diagnostic attribution (App.~\ref{app:failure_taxonomy_details}) and analyze the per-model and pooled failure patterns for the six general-purpose MLLMs (App.~\ref{subsec:app_general_purpose_failure_analysis}). We then analyze why the math-specialized MLLMs fail on MM-OptBench, including base-model context and an oracle-reading easy-case diagnostic for MM-Eureka's base model (App.~\ref{subsec:app_math_specialized_failure_analysis}). Finally, we report reference-runtime statistics and explain the role of the 1800-second evaluation timeout (App.~\ref{subsec:app_runtime_analysis}). App.~\ref{subsec:app_controlled_evaluations} reports controlled repair, image-downscaling, and temperature evaluations.

\subsection{Experimental Setup Details}
\label{subsec:app_exp_setup_details}

This subsection provides the experimental details that are compressed in Section~\ref{sec:experiments}. We specify the evaluated model groups, the official input--output interface, prompt templates and response contracts, the run protocol and metrics, the execution environment and correctness criteria, and the diagnostic procedures used for failure analysis. These details fix the evaluation contract behind the main-text results and make clear which measurements are official scores and which are diagnostic ablations.

\textbf{Model groups.} We evaluate two complementary model groups because no existing system is purpose-built for multimodal optimization modeling. The first group consists of six frontier general-purpose MLLMs: GPT-5.4, Gemini 3.1 Pro Preview, Claude Sonnet 4.6, Qwen3-VL-Plus, Qwen-VL-Max, and GLM-4.5V. These models were selected to cover strong contemporary multimodal systems from multiple model families and providers, rather than a single vendor or architecture, and to test whether broadly capable vision-language models can assemble solver-correct optimization artifacts from structured visual evidence. The second group tests whether specialization for multimodal mathematical reasoning transfers to this setting. We include MathCoder-VL-8B~\citep{wang2025mathcoder}, which emphasizes vision-to-code mathematical reasoning; MM-Eureka~\citep{meng2025mm}, which studies rule-based reinforcement learning for multimodal reasoning; and MM-PRM~\citep{du2025mm}, which uses scalable step-level supervision for multimodal mathematical reasoning. These models are relevant stress tests because MM-OptBench also requires mathematical abstraction, symbolic structure, and executable code, but differs from standard visual-math benchmarks by requiring a complete optimization formulation and solver-executable implementation. In the main benchmark comparison, all systems are evaluated as prompting-based black boxes, without MM-OptBench-specific fine-tuning, retrieval augmentation, or external agentic repair. This keeps the comparison focused on the capabilities already present in current MLLMs, rather than on benchmark-specific adaptation.

\paragraph{Model-specific inference settings.}
Table~\ref{tab:cr_inference_settings} specifies the reasoning modes and generation limits of the six general-purpose models. Generated code has a 1,800-second execution limit for every model; the prompt contract and allowed solver environment are shared across models.

\begin{table}[htbp]
\centering
\footnotesize
\setlength{\tabcolsep}{3pt}
\renewcommand{\arraystretch}{1.15}
\caption{Inference settings for the six general-purpose models. Total generation caps are in tokens; Qwen-VL-Max uses its provider limit. Code-execution timeout is 1,800 seconds throughout.}
\label{tab:cr_inference_settings}
\begin{tabularx}{\linewidth}{@{}>{\raggedright\arraybackslash}p{3.8cm}>{\raggedright\arraybackslash}p{2.15cm}>{\raggedright\arraybackslash}X r@{}}
\toprule
Model / version & Reasoning state & Effort or thinking budget & Token cap \\
\midrule
GPT-5.4\newline(\texttt{gpt-5.4-2026-03-05}) & Reasoning enabled & \texttt{high}; no numerical reasoning-token budget exposed & 32,768 \\
Gemini 3.1 Pro Preview\newline(\texttt{gemini-3.1-pro-preview}) & Dynamic thinking enabled & \texttt{high}; provider-managed dynamic budget & 32,768 \\
Claude Sonnet 4.6\newline(\texttt{claude-sonnet-4-6}) & Adaptive thinking enabled & \texttt{high}; no manual \texttt{budget\_tokens} setting & 32,768 \\
Qwen3-VL-Plus\newline(\texttt{qwen3-vl-plus-2025-12-19}) & Thinking enabled & No separate numerical thinking cap; reasoning usage recorded & 32,768 \\
Qwen-VL-Max\newline(\texttt{qwen-vl-max-2025-08-13}) & Non-thinking & Not applicable & 8,192 \\
GLM-4.5V\newline(\texttt{glm-4.5v}) & Thinking enabled & Forced thinking while enabled; no numerical budget exposed & 32,768 \\
\bottomrule
\end{tabularx}
\end{table}

\textbf{Official input and output.} For each benchmark instance, the model receives the model-facing inputs: \texttt{task\_input.txt} and the accompanying visual artifact(s). The requested output contains a mathematical formulation and a solver-executable artifact exposing \texttt{solve()}. Official scoring uses only this Stage-1 response. Oracle-reading and verified-extraction are diagnostic protocols and never alter Valid Code Rate, pass@1, or pass@4.

\textbf{Prompt templates and response contracts.} Table~\ref{tab:app_prompt_protocol} summarizes the stable prompt contracts used in the experiments. The templates below replace case-specific text, images, and structured payloads with placeholders; the released \texttt{task\_input.txt} files and visual artifacts provide the instance-specific content.

\begin{table}[h]
\centering
\footnotesize
\setlength{\tabcolsep}{4pt}
\renewcommand{\arraystretch}{1.22}
\caption{Prompt contracts used for official scoring and diagnostic ablations. Stage 1 is the only official scoring interaction; Stage 2, oracle-reading, and verified-extraction are diagnostic only.}
\label{tab:app_prompt_protocol}
\begin{tabular}{@{}>{\raggedright\arraybackslash}m{2.55cm} >{\raggedright\arraybackslash}m{3.15cm} >{\raggedright\arraybackslash}m{7.55cm}@{}}
\hline
Interaction & Model input & Required response \\
\hline
Stage 1: official scoring
& \texttt{task\_input.txt} plus visual artifact(s)
& Sections \texttt{ASSUMPTIONS}, \texttt{MATHEMATICAL\_MODEL}, and \texttt{PYTHON\_CODE}; the code must define a top-level \texttt{solve()} function returning a JSON-serializable dictionary. \\
\hline
Stage 2: diagnostic attribution
& Same \texttt{task\_input.txt} plus visual artifact(s), but only after a failed Stage-1 sample
& Valid JSON containing \texttt{case\_id}, \texttt{problem\_type}, and \texttt{extracted\_data}; the model is explicitly told not to solve the problem or write solver code. \\
\hline
Oracle-reading
& Ground-truth \texttt{instance\_data.json} packaged as a public-instance payload, with the original task wording only for context
& Same formulation-and-\texttt{solve()} response contract as Stage 1; this measures downstream formulation and code generation after public-instance reading is bypassed. \\
\hline
Verified-extraction
& A Stage-2 extraction that exactly matches \texttt{instance\_data.json}, repackaged as a public-instance payload
& Same formulation-and-\texttt{solve()} response contract as Stage 1; this measures downstream capability conditional on independently verified extraction. \\
\hline
\end{tabular}
\end{table}

\begin{prompttemplate}{System Message}
\begin{Verbatim}[breaklines,breakanywhere,fontsize=\footnotesize]
You are a senior operations research scientist and optimization modelling expert.
You can jointly interpret heterogeneous multimodal inputs, including natural-language descriptions, multiple images, tables, charts, coordinate plots, schedules, maps, network diagrams, legends, and annotations.
Your job is to extract instance-specific data faithfully, identify the optimization objective, formulate a correct mathematical model, choose an appropriate exact or heuristic solution strategy, and write runnable Python code.
Prefer exact methods whenever justified by the problem structure; if a direct exact algorithm is simpler than a generic solver, implement it directly in Python.
Favor compact, faithful data encodings and concise solver code over bloated representations.
When returning structured Python results, keep all dict keys JSON-safe strings rather than tuples or other non-serializable key types.
Never invent unreadable or missing data. Surface uncertainty explicitly, keep assumptions minimal, and make intermediate artefacts inspectable through structured outputs.
Follow the requested output contract exactly and use English only.
\end{Verbatim}
\end{prompttemplate}

\begin{prompttemplate}{Stage-1 User Template}
\begin{Verbatim}[breaklines,breakanywhere,fontsize=\footnotesize]
Case ID: <case_id>
You will receive a benchmark optimization instance via text plus one or more attached visuals.
Use all provided modalities jointly, and treat the visuals as a primary source of instance-specific data.
Your task is end-to-end full solving: understand the instance, formulate the model, and write runnable Python code.
Task statement: <contents of task_input.txt>
Attached visuals in order: <visual file names>
Output requirements: ASSUMPTIONS, MATHEMATICAL_MODEL, PYTHON_CODE. The Python code must be self-contained, include the instance data explicitly, define solve(), return a JSON-serializable dictionary, avoid markdown fences, and use only the allowed solver/scientific packages.
\end{Verbatim}
\end{prompttemplate}

\begin{prompttemplate}{Stage-2 User Template}
\begin{Verbatim}[breaklines,breakanywhere,fontsize=\footnotesize]
Case ID: <case_id>
You will receive a benchmark optimization instance via text plus one or more attached visuals.
Use all provided modalities jointly, and treat the visuals as a primary source of instance-specific data.
Your task is extraction only: recover the instance-specific structured data faithfully from the text and visuals. Do not solve the optimization problem and do not write solver code.
Task statement: <contents of task_input.txt>
Attached visuals in order: <visual file names>
Output requirements: Return valid JSON only, without markdown fences. Use exactly the fields case_id, problem_type, and extracted_data. Inside extracted_data, include structured_instance, assumptions, and uncertain_or_missing.
\end{Verbatim}
\end{prompttemplate}

\begin{prompttemplate}{Oracle-Reading and Verified-Extraction User Template}
\begin{Verbatim}[breaklines,breakanywhere,fontsize=\footnotesize]
Case ID: <case_id>
You will receive a benchmark optimization instance via the task statement plus a verified public-instance payload.
Treat the supplied payload as authoritative public data for this case.
Your task is downstream solving only: use the verified public-instance payload to formulate the optimization model and write runnable Python code. Do not re-infer the instance from visuals.
Structured-payload usage rules: <payload-is-authoritative rules>
Structured public-instance payload: <instance_data.json or verified extraction>
Original benchmark wording for context only: <contents of task_input.txt>
Output requirements: ASSUMPTIONS, MATHEMATICAL_MODEL, PYTHON_CODE, with the same solve() contract as Stage 1.
\end{Verbatim}
\end{prompttemplate}

\textbf{Runs and metrics.} We report Valid Code Rate, pass@1, and pass@4 as defined in Sec.~\ref{subsec:setting_protocol}. For the main benchmark comparison, every evaluated model is run five times. For models with stochastic sampling support, the five runs consist of one temperature-zero run at $0.0$ and four stochastic runs at temperature $0.4$. Valid Code Rate and pass@1 are summarized as mean$\pm$standard deviation over the five runs, while pass@4 is reported as the multi-sample success rate over the four stochastic runs.

\textbf{Execution environment and correctness.} Generated solver artifacts are executed in an isolated Python 3.12.7 environment with common optimization and scientific-computing packages available, including \texttt{networkx} 3.6.1, \texttt{scipy} 1.17.1, \texttt{pulp} 3.3.0, OR-Tools 9.15.6755, \texttt{gurobipy} 13.0.1, and \texttt{mip} 1.17.6. During benchmark construction, candidate instances are retained only after their reference solvers certify an optimum under the family-specific construction budget; most subcategories are designed to solve in seconds to tens of seconds, and the released-reference runtimes are well below the evaluation cap. For model evaluation, we use a more permissive 1800-second timeout as a grace period for slower generated implementations, rather than as a claim that the benchmark instances themselves require this much time; App.~\ref{subsec:app_runtime_analysis} gives the supporting runtime analysis. A sample is marked incorrect if the executable artifact fails to compile or run, times out, or fails the task-specific correctness criterion in Section~\ref{subsec:setting_protocol}. The default numeric tolerance is $10^{-6}$; continuous-objective location families whose released reference objectives are rounded to four decimals use an absolute-tolerance floor of $10^{-3}$.

\textbf{Failure analysis.} RQ4 uses a typical run because merged six-model official pass@1 varies only slightly across the five runs. For each failed Stage-1 sample, Stage 2 asks the model to extract the public instance data without solving the problem; this extraction is compared with \texttt{instance\_data.json} to distinguish mismatching, matching, and unresolved extraction follow-ups. Oracle-reading separately supplies the ground-truth public instance record and measures downstream formulation-and-code generation after the multimodal reading burden is removed.

\subsection{Extended Performance Tables and pass@4 Heatmap}
\label{subsec:app_extended_performance_tables}

This subsection gives the numerical results behind the compact plots and claims in Section~\ref{sec:experiments}. App.~\ref{subsubsec:app_overall_metrics} reports the model-level official metrics used to interpret overall capability and math-specialized transfer. Apps.~\ref{subsubsec:app_family_results} and~\ref{subsubsec:app_difficulty_results} provide the family-level and difficulty-level slices behind Figure~\ref{fig:exp_family_heatmap}. App.~\ref{subsubsec:app_pass4_heatmap} repeats the same structural analysis under pass@4 to check whether multi-sample generation changes the conclusions. Unless otherwise stated, \emph{Six-model Avg.} refers to the average over the six general-purpose MLLMs only; the three math-specialized models are analyzed separately in App.~\ref{subsec:app_math_specialized_failure_analysis} because they solve no official instances in any slice. All reported values in this subsection follow the five-run protocol in App.~\ref{subsec:app_exp_setup_details}.

\subsubsection{Overall Metrics and Model-Level Performance}
\label{subsubsec:app_overall_metrics}

This subsubsection provides the model-level numbers behind Figure~\ref{fig:exp_overall_summary}. Table~\ref{tab:app_exp_overall} shows how often each model produces executable code, how often a single generation is solver-correct, and how much multi-sample generation helps under pass@4. The aggregate row averages only the six general-purpose MLLMs, so it reflects the performance trend among models that obtain nonzero official solves. The three math-specialized MLLMs remain in the table as a separate stress-test group.

\begin{table}[h]
\centering
\caption{Overall modeling performance for all nine evaluated MLLMs on the finalized six-family benchmark under the official scoring stage. Valid Code Rate and pass@1 report the five-run mean with standard deviation in the subscript; pass@4 reports the multi-sample success rate. The Six-model Avg. row averages the six general-purpose MLLMs only.}
\label{tab:app_exp_overall}
\setlength{\tabcolsep}{5pt}
\begin{tabular}{llccc}
\hline
Model group & Model & Valid Code Rate (\%) & pass@1 (\%) & pass@4 (\%) \\
\hline
\multirow{6}{*}{\makecell[c]{General-\\purpose}}
& GPT-5.4 & $91.8_{\pm 0.9}$ & $52.1_{\pm 0.9}$ & 68.3 \\
& Gemini 3.1 Pro Preview & $92.5_{\pm 0.8}$ & $51.3_{\pm 0.7}$ & 64.4 \\
& Claude Sonnet 4.6 & $83.3_{\pm 2.6}$ & $23.5_{\pm 1.5}$ & 31.3 \\
& Qwen-VL-Max & $64.0_{\pm 1.5}$ & $16.8_{\pm 0.9}$ & 25.5 \\
& Qwen3-VL-Plus & $78.6_{\pm 0.7}$ & $20.3_{\pm 1.3}$ & 34.7 \\
& GLM-4.5V & $69.0_{\pm 2.5}$ & $15.4_{\pm 1.0}$ & 29.1 \\
\hline
\makecell[c]{General-purpose\\average} & Six-model Avg. & 79.9 & 29.9 & 42.2 \\
\hline
\multirow{3}{*}{\makecell[c]{Math-\\specialized}}
& MathCoder-VL-8B & $0.0_{\pm 0.0}$ & $0.0_{\pm 0.0}$ & 0.0 \\
& MM-Eureka & $6.6_{\pm 4.5}$ & $0.0_{\pm 0.0}$ & 0.0 \\
& MM-PRM & $0.0_{\pm 0.0}$ & $0.0_{\pm 0.0}$ & 0.0 \\
\hline
\end{tabular}

\end{table}

\textbf{Overall metric interpretation.} Table~\ref{tab:app_exp_overall} supports five observations.

\noindent\textbf{(1) Executability is far from correctness.} The strongest two models, GPT-5.4 and Gemini 3.1 Pro Preview, both produce runnable artifacts on more than 90\% of samples, but solve only about half of the benchmark on a single attempt: $52.1\%\pm0.9\%$ and $51.3\%\pm0.7\%$ pass@1, respectively. This leaves gaps of 39.7 and 41.2 percentage points between executable code and solver-correct modeling, showing that most remaining errors are not merely syntax or environment failures.

\noindent\textbf{(2) The executability--correctness gap is larger for mid-tier systems.} Claude Sonnet 4.6 and Qwen3-VL-Plus have high Valid Code Rates, $83.3\%\pm2.6\%$ and $78.6\%\pm0.7\%$, but their pass@1 values are only $23.5\%\pm1.5\%$ and $20.3\%\pm1.3\%$. Qwen-VL-Max and GLM-4.5V show the same pattern at lower executability levels. Thus, being able to emit runnable solver code is a necessary but weak proxy for constructing the intended optimization model.

\noindent\textbf{(3) Multiple samples help, but do not remove the core difficulty.} pass@4 improves every general-purpose model, with GPT-5.4 rising from 52.1\% to 68.3\% and Gemini 3.1 Pro Preview from 51.3\% to 64.4\%. However, even the best pass@4 result leaves nearly one third of instances unsolved after four stochastic attempts. The mid-tier models remain well below 35\% pass@4, indicating that additional sampling does not compensate for systematic modeling failures.

\noindent\textbf{(4) The six-model aggregate remains low.} Averaging over the six general-purpose MLLMs gives 79.9\% Valid Code Rate, but only 29.9\% pass@1 and 42.2\% pass@4. This aggregate makes the benchmark-level gap clear: the typical frontier MLLM can often produce code that runs, yet still fails to encode the intended optimization model.

\noindent\textbf{(5) Math-specialized MLLMs do not transfer under the official interface.} MathCoder-VL-8B, MM-Eureka, and MM-PRM obtain 0.0\% pass@1 and 0.0\% pass@4. MM-Eureka is the only one with a nonzero Valid Code Rate, $6.6\%\pm4.5\%$, but this executability does not translate into any official solve. These models are therefore listed in Table~\ref{tab:app_exp_overall} but excluded from the Six-model Avg. row and the family/difficulty aggregates; otherwise, their uniformly zero official success would obscure the structural trends among general-purpose MLLMs. App.~\ref{subsec:app_math_specialized_failure_analysis} gives a dedicated failure analysis for these models.

\subsubsection{Family-Level Performance}
\label{subsubsec:app_family_results}

This subsubsection examines whether model performance is tied to the mathematical structure of the optimization family. Table~\ref{tab:app_exp_family} reports pass@1 and pass@4 for each general-purpose MLLM across the six major families, together with the Six-model Avg. columns used in the main-text family analysis. These numbers show which formulation regimes remain difficult after averaging over model-specific strengths and after allowing multiple stochastic attempts.

\begin{table*}[h]
\centering
\scriptsize
\setlength{\tabcolsep}{4pt}
\caption{Modeling performance of the six general-purpose MLLMs across major optimization families. pass@1 reports the five-run mean with standard deviation in the subscript; pass@4 reports the corresponding multi-sample success rate. The rightmost columns report the average over these six models.}
\label{tab:app_exp_family}
\resizebox{\textwidth}{!}{%
\begin{tabular}{lcccccccccccccc}
\hline
& \multicolumn{2}{c}{GPT-5.4}
& \multicolumn{2}{c}{Gemini 3.1 Pro Preview}
& \multicolumn{2}{c}{Claude Sonnet 4.6}
& \multicolumn{2}{c}{Qwen-VL-Max}
& \multicolumn{2}{c}{Qwen3-VL-Plus}
& \multicolumn{2}{c}{GLM-4.5V}
& \multicolumn{2}{c}{Six-model Avg.} \\
\cline{2-15}
Family
& pass@1 & pass@4
& pass@1 & pass@4
& pass@1 & pass@4
& pass@1 & pass@4
& pass@1 & pass@4
& pass@1 & pass@4
& pass@1 & pass@4 \\
\hline
Network Optimization
& $39.6_{\pm 2.0}$ & 57.3
& $68.5_{\pm 1.7}$ & 74.7
& $8.5_{\pm 1.7}$ & 14.7
& $12.8_{\pm 1.7}$ & 24.7
& $24.7_{\pm 4.8}$ & 36.7
& $16.4_{\pm 2.6}$ & 32.7
& 28.4 & 40.1 \\
Location / Covering / Assignment
& $59.7_{\pm 2.5}$ & 70.0
& $63.1_{\pm 3.0}$ & 67.3
& $32.9_{\pm 2.3}$ & 41.3
& $17.6_{\pm 1.2}$ & 25.3
& $27.3_{\pm 0.8}$ & 44.7
& $25.5_{\pm 2.0}$ & 46.0
& 37.7 & 49.1 \\
Scheduling / Sequencing
& $46.8_{\pm 3.5}$ & 74.2
& $39.5_{\pm 3.5}$ & 67.5
& $20.2_{\pm 3.4}$ & 29.2
& $3.7_{\pm 1.0}$ & 9.2
& $7.7_{\pm 2.1}$ & 22.5
& $5.7_{\pm 3.1}$ & 15.0
& 20.6 & 36.3 \\
Multi-Period / System Planning
& $43.8_{\pm 3.1}$ & 60.0
& $33.7_{\pm 1.4}$ & 53.3
& $26.8_{\pm 4.7}$ & 36.7
& $9.0_{\pm 2.7}$ & 15.8
& $17.7_{\pm 3.6}$ & 30.0
& $10.2_{\pm 1.7}$ & 16.7
& 23.5 & 35.4 \\
Routing / Tour Optimization
& $41.5_{\pm 2.2}$ & 56.7
& $46.7_{\pm 1.6}$ & 55.8
& $19.8_{\pm 2.5}$ & 25.8
& $18.0_{\pm 1.5}$ & 24.2
& $15.3_{\pm 2.5}$ & 29.2
& $7.8_{\pm 1.7}$ & 20.8
& 24.9 & 35.4 \\
Combinatorial / Logical Models
& $82.3_{\pm 5.7}$ & 94.2
& $49.2_{\pm 3.3}$ & 64.2
& $33.8_{\pm 2.3}$ & 41.7
& $40.3_{\pm 2.8}$ & 54.2
& $26.3_{\pm 1.7}$ & 42.5
& $23.8_{\pm 0.7}$ & 38.3
& 42.5 & 55.8 \\
\hline
\end{tabular}}
\end{table*}

\textbf{Family-level interpretation.} Table~\ref{tab:app_exp_family} supports four observations.

\noindent\textbf{(1) Model rankings depend strongly on optimization structure.} Gemini 3.1 Pro Preview is the strongest model on Network Optimization (68.5\% pass@1) and Location / Covering / Assignment (63.1\%), whereas GPT-5.4 leads on Scheduling / Sequencing (46.8\%), Multi-Period / System Planning (43.8\%), and Combinatorial / Logical Models (82.3\%). The benchmark therefore does not induce a single global ordering of models; different formulation primitives favor different systems.

\noindent\textbf{(2) Family difficulty is not determined by visual size alone.} The rightmost columns make the family averages explicit: six-model pass@1 is highest for Combinatorial / Logical Models (42.5\%) and Location / Covering / Assignment (37.7\%), even though these families can contain dense matrices, incidence structures, and spatial relations. By contrast, Scheduling / Sequencing (20.6\%) and Multi-Period / System Planning (23.5\%) are harder because correct solutions require globally consistent temporal, resource, or balance constraints rather than local extraction of visible entries.

\noindent\textbf{(3) Multi-sample generation raises success rates but preserves the structural pattern.} pass@4 improves every general-purpose model--family pair. At the aggregate level, the rightmost columns show gains of roughly 10--16 points over pass@1 across all families; the largest average gain appears in Scheduling / Sequencing (20.6\% to 36.3\%). The gains are especially large for GPT-5.4 (46.8\% to 74.2\%) and Gemini 3.1 Pro Preview (39.5\% to 67.5\%) on Scheduling / Sequencing, suggesting that some failures are recoverable through additional stochastic attempts. However, the family ordering remains similar: Combinatorial / Logical Models and Location / Covering / Assignment remain the strongest slices, while Scheduling / Sequencing, Multi-Period / System Planning, and Routing / Tour Optimization remain substantially lower on average.

\noindent\textbf{(4) The weaker models expose family-specific bottlenecks rather than uniform underperformance.} Qwen-VL-Max performs relatively better on Combinatorial / Logical Models (40.3\% pass@1) than on Scheduling / Sequencing (3.7\%), while GLM-4.5V is stronger on Location / Covering / Assignment (25.5\%) than on Routing / Tour Optimization (7.8\%). These differences suggest that failure is shaped by the interaction between the model and the required optimization structure, not merely by overall model scale or formatting robustness.

\subsubsection{Difficulty-Level Performance}
\label{subsubsec:app_difficulty_results}

This subsubsection tests whether the released difficulty levels correspond to measurable increases in modeling difficulty. Table~\ref{tab:app_exp_difficulty} reports pass@1 and pass@4 for the six general-purpose MLLMs across easy, medium, and hard instances. The aggregate columns summarize how performance changes as instances require denser indexing, stronger coupling, longer temporal or routing structure, and more integrated use of visual evidence.

\begin{table*}[h]
\centering
\scriptsize
\setlength{\tabcolsep}{4pt}
\caption{Modeling performance of the six general-purpose MLLMs across benchmark difficulty levels. pass@1 reports the five-run mean with standard deviation in the subscript; pass@4 reports the corresponding multi-sample success rate. The rightmost columns report the average over these six models.}
\label{tab:app_exp_difficulty}
\resizebox{\textwidth}{!}{%
\begin{tabular}{lcccccccccccccc}
\hline
& \multicolumn{2}{c}{GPT-5.4}
& \multicolumn{2}{c}{Gemini 3.1 Pro Preview}
& \multicolumn{2}{c}{Claude Sonnet 4.6}
& \multicolumn{2}{c}{Qwen-VL-Max}
& \multicolumn{2}{c}{Qwen3-VL-Plus}
& \multicolumn{2}{c}{GLM-4.5V}
& \multicolumn{2}{c}{Six-model Avg.} \\
\cline{2-15}
Difficulty
& pass@1 & pass@4
& pass@1 & pass@4
& pass@1 & pass@4
& pass@1 & pass@4
& pass@1 & pass@4
& pass@1 & pass@4
& pass@1 & pass@4 \\
\hline
Easy
& $67.6_{\pm 1.7}$ & 82.3
& $64.5_{\pm 1.5}$ & 80.4
& $38.8_{\pm 2.2}$ & 49.2
& $27.5_{\pm 1.3}$ & 40.4
& $32.8_{\pm 3.1}$ & 53.8
& $29.3_{\pm 1.9}$ & 50.4
& 43.4 & 59.4 \\
Medium
& $54.0_{\pm 2.5}$ & 70.8
& $56.1_{\pm 1.6}$ & 68.5
& $22.2_{\pm 2.7}$ & 29.2
& $17.2_{\pm 1.9}$ & 26.9
& $19.1_{\pm 2.2}$ & 32.7
& $13.2_{\pm 1.1}$ & 27.3
& 30.2 & 42.6 \\
Hard
& $34.7_{\pm 1.3}$ & 51.9
& $33.2_{\pm 3.4}$ & 44.2
& $9.4_{\pm 1.9}$ & 15.4
& $5.6_{\pm 1.4}$ & 9.2
& $9.1_{\pm 1.6}$ & 17.7
& $3.5_{\pm 1.0}$ & 9.6
& 15.9 & 24.7 \\
\hline
\end{tabular}}
\end{table*}

\textbf{Difficulty-level interpretation.} Table~\ref{tab:app_exp_difficulty} supports four observations.

\noindent\textbf{(1) Difficulty produces a monotone performance drop for every general-purpose model.} The rightmost columns show that the six-model pass@1 average decreases from 43.4\% on easy instances to 30.2\% on medium and 15.9\% on hard. The same monotone pattern appears under pass@4, which drops from 59.4\% to 42.6\% and then 24.7\%. This consistency across metrics indicates that the released difficulty levels capture real structural complexity rather than random variation in a single run.

\noindent\textbf{(2) The strongest models remain far from reliable on hard instances.} GPT-5.4 falls from 67.6\% pass@1 on easy instances to 34.7\% on hard, and Gemini 3.1 Pro Preview falls from 64.5\% to 33.2\%. Even under pass@4, the hard-tier success rates are only 51.9\% and 44.2\%. Thus, the hard split is not solved by current frontier general-purpose MLLMs, even with multiple samples.

\noindent\textbf{(3) Weaker models collapse sharply as structure becomes harder.} Claude Sonnet 4.6, Qwen3-VL-Plus, Qwen-VL-Max, and GLM-4.5V all fall below 10\% pass@1 on hard instances. Their hard-tier pass@4 values also remain low, ranging from 9.2\% to 17.7\%. This shows that additional sampling cannot compensate when the model lacks the ability to assemble the required global formulation.

\noindent\textbf{(4) The difficulty trend aligns with the intended benchmark design.} Easy cases still require multimodal formulation, but they involve smaller or cleaner structural interactions. Medium and hard cases add larger index sets, tighter couplings, denser constraints, or longer planning horizons. The observed degradation therefore supports the release design: difficulty is not only a change in instance size, but a change in the amount of structure that must be represented consistently in the generated solver artifact.

\subsubsection{pass@4 Structural Heatmap}
\label{subsubsec:app_pass4_heatmap}

\begin{figure*}[h]
\centering
\includegraphics[width=0.98\textwidth]{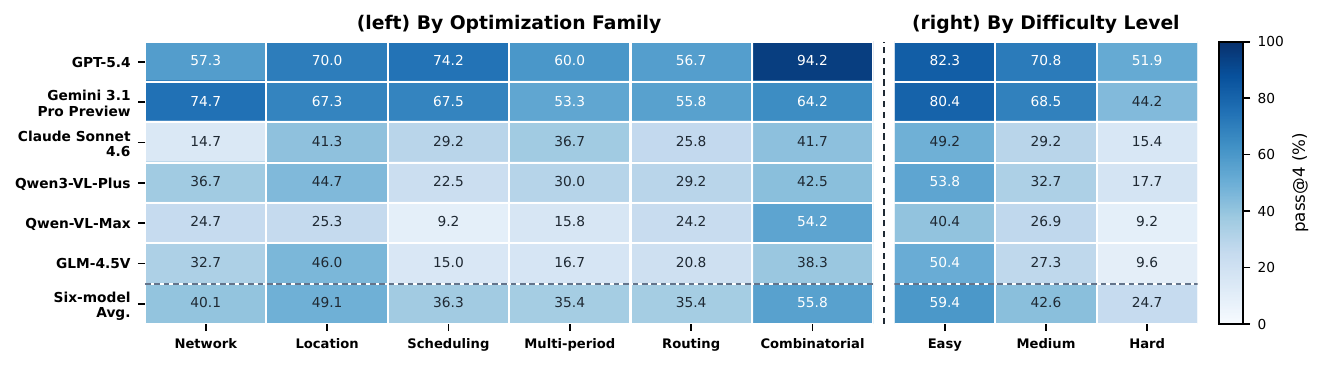}
\caption{pass@4 for the six general-purpose MLLMs across major optimization families (left) and released difficulty levels (right), using the same layout as the main-text pass@1 heatmap. The bottom row reports the average over these six models.}
\label{fig:app_exp_family_heatmap_pass4}
\end{figure*}

This subsubsection uses pass@4 to distinguish sampling limitations from persistent modeling limitations. If failures were mainly due to a model occasionally missing a correct formulation, allowing four stochastic attempts should substantially change the family and difficulty patterns. Figure~\ref{fig:app_exp_family_heatmap_pass4} therefore mirrors the main-text pass@1 heatmap with the same model axis, family/difficulty layout, and Six-model Avg. row, so that the structural trends can be compared directly.

\textbf{pass@4 heatmap interpretation.} Figure~\ref{fig:app_exp_family_heatmap_pass4} supports four observations.

\noindent\textbf{(1) pass@4 raises absolute success rates without changing the main structural conclusion.} Location / Covering / Assignment (49.1\% six-model average) and Combinatorial / Logical Models (55.8\%) remain the strongest family slices, while Scheduling / Sequencing (36.3\%), Multi-Period / System Planning (35.4\%), and Routing / Tour Optimization (35.4\%) remain lower. Repeated sampling therefore helps recover some missed solutions, but it does not erase family-specific weaknesses.

\noindent\textbf{(2) The leading systems benefit in different places.} GPT-5.4 reaches 94.2\% pass@4 on Combinatorial / Logical Models and improves strongly on Scheduling / Sequencing, rising from 46.8\% pass@1 to 74.2\% pass@4. Gemini 3.1 Pro Preview is strongest on Network Optimization (74.7\%) and remains high on Location / Covering / Assignment (67.3\%) and Scheduling / Sequencing (67.5\%). These gains indicate that stochastic sampling can help when at least some generations contain the right formulation template.

\noindent\textbf{(3) Multi-sample generation does not close the gap for weaker models.} Qwen-VL-Max remains below 26\% pass@4 on Scheduling / Sequencing, Multi-Period / System Planning, and Routing / Tour Optimization, and GLM-4.5V remains below 21\% on the same three families. These results suggest that the limiting factor is often not the lack of one lucky sample, but a persistent inability to construct the required model.

\noindent\textbf{(4) The difficulty panel confirms that hard instances remain hard after multiple attempts.} The six-model average pass@4 falls from 59.4\% on easy instances to 42.6\% on medium and 24.7\% on hard. Even the strongest systems solve only about half of hard instances under pass@4, with GPT-5.4 at 51.9\% and Gemini 3.1 Pro Preview at 44.2\%. This supports the main-text claim that released difficulty reflects structural complexity rather than sampling variance alone.

\subsection{Detailed Failure Taxonomy}
\label{app:failure_taxonomy_details}

\begin{table}[h]
\centering
\footnotesize
\setlength{\tabcolsep}{3pt}
\renewcommand{\arraystretch}{1.15}
\renewcommand{\tabularxcolumn}[1]{m{#1}}
\caption{Detailed failure taxonomy used in MM-OptBench. The table summarizes where each failure type arises, its typical symptom, and representative boundary cases. Figure~\ref{fig:exp_failure_breakdown} uses a coarser legend for readability: \emph{Modeling / algorithmic} merges Structural modeling and Solution-procedure / algorithmic; \emph{Runtime} corresponds to Execution-level; and \emph{Other code / artifact} denotes response-contract or solver-artifact failures. This regrouping is only for visualization; the table preserves the finer category boundaries.}
\label{tab:failure_taxonomy_appendix}
\begin{tabularx}{\textwidth}{>{\centering\arraybackslash}m{2.35cm} >{\centering\arraybackslash}m{2.65cm} >{\centering\arraybackslash}m{3.35cm} >{\raggedright\arraybackslash}X}
\toprule
\textbf{Category} & \textbf{Where the fault lives} & \textbf{Typical symptoms} & \textbf{Representative examples} \\
\midrule
Upstream reading / extraction
& Multimodal extraction of the public instance before formulation and code generation
& Wrong entities, counts, relations, or public parameters are extracted from the multimodal input
& Reading the wrong adjacency relation from a graph matrix; miscounting customers, jobs, facilities, or time periods from a visual panel; extracting incorrect capacities, time windows, processing times, or subset memberships; extracting the wrong feasible assignment or coverage relation from a map-like figure.
\\
\midrule
Other code / artifact
& Response contract or solver-artifact interface
& The output does not expose a usable solver artifact in the form required by the evaluation program
& Missing the required \texttt{solve()} entry point; returning prose or an unsupported object instead of the required result dictionary; mixing incompatible code fragments; relying on external files that are not provided; omitting required data bindings or producing an artifact with the wrong interface.
\\
\midrule
Execution-level
& Running the generated solver code
& The generated solver never reaches a valid candidate solution because it fails to compile, import, run, or terminate
& Malformed Python syntax; missing solver-library imports such as \texttt{gurobipy}, \texttt{pulp}, or \texttt{ortools}; runtime key/index/type errors during model construction or output formatting; exact or search-based procedures that exceed the runtime budget.
\\
\midrule
Structural modeling
& The optimization formulation itself
& Code executes, but the encoded optimization problem is not the intended one
& Omitting flow-conservation or capacity constraints in network optimization; dropping assignment exclusivity in location-allocation; relaxing binary or integer decisions into continuous variables; missing precedence or machine non-overlap constraints in scheduling; using the wrong objective sense or an incomplete objective definition.
\\
\midrule
Solution-procedure / algorithmic
& The method used to solve the encoded problem
& The model appears broadly plausible, but the returned answer is still wrong or not exact enough
& Using a heuristic where exact solving is required by the benchmark; implementing an incorrect dynamic program, branch-and-bound procedure, or search logic; solving only a relaxation and returning it as the final answer; terminating local search early and reporting the incumbent as optimal; returning a feasible but non-optimal solution when the task requires proven optimality.
\\
\bottomrule
\end{tabularx}
\end{table}

This subsection defines the failure labels used in Section~\ref{subsec:exp_failures}. These labels are diagnostic only: they are assigned after a sample has already failed the official Stage-1 score, and they do not change Valid Code Rate, pass@1, or pass@4. These labels describe two checks: whether the model extracts the input data correctly in Stage 2, and what errors remain when it is given the correct input data directly.

\textbf{Stage-1 and Stage-2 roles.} Stage 1 is the official scoring interaction: the model receives the text-plus-visual input and returns a formulation together with a solver-executable artifact. If that artifact is solver-correct, no failure label is needed. If it fails, Stage 2 asks a narrower extraction-only question: the model must recover the public instance data from the same input, such as nodes, capacities, coordinates, jobs, processing times, demands, or time-indexed parameters, without solving the optimization problem. The extraction is compared with \texttt{instance\_data.json}, the hidden machine-readable record from which the benchmark input and reference solution were derived.

\textbf{Top-level split.} If the data extracted in Stage 2 do not match \texttt{instance\_data.json}, we label the case \emph{Upstream reading / extraction}. If the data match, we label it \emph{Downstream after matched extraction}: the model extracts the data correctly in Stage 2, but its earlier Stage-1 response failed to solve the problem. Stage 2 is a separate test, so correct extraction there does not tell us whether the model read the data correctly in Stage 1. If the Stage-2 follow-up is missing or cannot be parsed, the case is kept as unresolved rather than forced into either branch.

\textbf{Downstream categories and figure labels.} Table~\ref{tab:failure_taxonomy_appendix} gives the detailed downstream categories. \emph{Other code / artifact} covers response-contract or solver-artifact interface failures; \emph{Execution-level} covers code that cannot compile, import, run, or terminate during evaluation; \emph{Structural modeling} covers runnable artifacts that encode the wrong optimization model; and \emph{Solution-procedure / algorithmic} covers cases where the solving method is heuristic, incomplete, or otherwise not exact enough. Figure~\ref{fig:exp_failure_breakdown} uses shorter category names for readability. Table~\ref{tab:failure_figure_label_definitions} gives the exact definitions of the figure categories and maps them to the detailed taxonomy.

\begin{table}[h]
\centering
\footnotesize
\setlength{\tabcolsep}{4pt}
\renewcommand{\arraystretch}{1.12}
\renewcommand{\tabularxcolumn}[1]{m{#1}}
\caption{Definitions of the compact failure labels used in Figure~\ref{fig:exp_failure_breakdown}. The left-panel labels come from the Stage-2 attribution protocol; the right-panel labels are downstream oracle-reading failure groups derived from Table~\ref{tab:failure_taxonomy_appendix}.}
\label{tab:failure_figure_label_definitions}
\begin{tabularx}{\textwidth}{>{\centering\arraybackslash}m{2.45cm} >{\centering\arraybackslash}m{2.0cm} >{\raggedright\arraybackslash}X}
\toprule
\textbf{Figure label} & \textbf{Panel} & \textbf{Definition} \\
\midrule
Reading mismatch
& Left
& The Stage-2 extraction is parseable but disagrees with \texttt{instance\_data.json}. \\
\midrule
Downstream after matched extraction
& Left
& The Stage-2 extraction matches \texttt{instance\_data.json}, but the original Stage-1 solver artifact remains solver-incorrect. \\
\midrule
Schema / interrupted
& Left
& The Stage-2 diagnostic follow-up is missing, interrupted, or not parseable as the expected extraction payload. These are unresolved diagnostic cases rather than a separate modeling-failure category. \\
\midrule
Modeling / algorithmic
& Right
& Oracle-reading failure caused by an incorrect encoded optimization model or an incomplete/incorrect solving procedure; this label merges \emph{Structural modeling} and \emph{Solution-procedure / algorithmic} from Table~\ref{tab:failure_taxonomy_appendix}. \\
\midrule
Runtime
& Right
& Oracle-reading failure caused by code that fails to compile, import, run, or terminate during evaluation; this corresponds to \emph{Execution-level} in Table~\ref{tab:failure_taxonomy_appendix}. \\
\midrule
Other code / artifact
& Right
& Oracle-reading failure caused by response-contract or solver-artifact interface problems, such as missing \texttt{solve()}, unsupported return objects, or unusable code artifacts; this corresponds to \emph{Other code / artifact} in Table~\ref{tab:failure_taxonomy_appendix}. \\
\bottomrule
\end{tabularx}
\end{table}

\textbf{Taxonomy interpretation.} The categories indicate where improvement would be needed. Upstream reading/extraction failures point to better extraction of instance data from text and visuals. Other code/artifact failures point to stronger adherence to the benchmark interface. Execution-level failures point to more reliable solver code. Structural-modeling failures require better translation from instance data into variables, constraints, and objectives. Solution-procedure failures require exact or certified solving rather than heuristic or incomplete algorithms. This is why MM-OptBench reports both official solver-grounded scores and diagnostic attribution: the same wrong objective value can arise from very different failure mechanisms.

\subsection{Failure Analysis for General-Purpose MLLMs}
\label{subsec:app_general_purpose_failure_analysis}

This subsection analyzes the failure patterns of the six general-purpose MLLMs shown in Figure~\ref{fig:exp_failure_breakdown}. The unit is a \emph{failed model-instance output}, not a benchmark instance: each model produces outputs on the 780 instances, and the tables below either separate those failures by model or pool them in the final row. Therefore the totals are larger than 780. The compact labels used in Figure~\ref{fig:exp_failure_breakdown} are defined in Table~\ref{tab:failure_figure_label_definitions}, with the fuller diagnostic taxonomy in Table~\ref{tab:failure_taxonomy_appendix}. Table~\ref{tab:app_exp_failure_breakdown_official} reports the official two-stage failure counts behind the left panel of Figure~\ref{fig:exp_failure_breakdown}; Table~\ref{tab:app_exp_failure_breakdown_oracle} reports the oracle-reading failure counts behind the right panel; and Table~\ref{tab:app_exp_failure_schema} expands the parse-failure component of the compact \emph{Schema / interrupted} label.

\begin{table}[h]
\centering
\footnotesize
\setlength{\tabcolsep}{4pt}
\caption{Official two-stage failure breakdown behind the left panel of Figure~\ref{fig:exp_failure_breakdown}. Category cells report count/780 for individual models and count/4680 for the pooled row; parenthetical values are percentages within that row's failed official outputs. The final column reports total failed official outputs under the same denominator convention.}
\label{tab:app_exp_failure_breakdown_official}
\resizebox{\linewidth}{!}{
\begin{tabular}{lrrrr}
\hline
Model & Reading mismatch & Downstream after matched extraction & Schema / interrupted & Total failed \\
\hline
GPT-5.4 & 300/780 (82.0\%) & 58/780 (15.8\%) & 8/780 (2.2\%) & 366/780 \\
Gemini 3.1 Pro Preview & 245/780 (64.8\%) & 101/780 (26.7\%) & 32/780 (8.5\%) & 378/780 \\
Claude Sonnet 4.6 & 450/780 (75.8\%) & 97/780 (16.3\%) & 47/780 (7.9\%) & 594/780 \\
Qwen3-VL-Plus & 343/780 (53.7\%) & 271/780 (42.4\%) & 25/780 (3.9\%) & 639/780 \\
Qwen-VL-Max & 338/780 (51.9\%) & 288/780 (44.2\%) & 25/780 (3.8\%) & 651/780 \\
GLM-4.5V & 373/780 (55.3\%) & 264/780 (39.2\%) & 37/780 (5.5\%) & 674/780 \\
\hline
Six-model Overall & 2049/4680 (62.1\%) & 1079/4680 (32.7\%) & 174/4680 (5.2\%) & 3302/4680 \\
\hline
\end{tabular}
}
\end{table}

\begin{table}[h]
\centering
\footnotesize
\setlength{\tabcolsep}{4pt}
\caption{Oracle-reading failure breakdown behind the right panel of Figure~\ref{fig:exp_failure_breakdown}. Category cells report count/780 for individual models and count/4680 for the pooled row; parenthetical values are percentages within that row's failed oracle-reading outputs. The final column reports total failed oracle-reading outputs under the same denominator convention.}
\label{tab:app_exp_failure_breakdown_oracle}
\begin{tabular}{lrrrr}
\hline
Model & Modeling / algorithmic & Runtime & Other code / artifact & Total failed \\
\hline
GPT-5.4 & 129/780 (64.2\%) & 60/780 (29.9\%) & 12/780 (6.0\%) & 201/780 \\
Gemini 3.1 Pro Preview & 180/780 (74.7\%) & 47/780 (19.5\%) & 14/780 (5.8\%) & 241/780 \\
Claude Sonnet 4.6 & 128/780 (50.8\%) & 63/780 (25.0\%) & 61/780 (24.2\%) & 252/780 \\
Qwen3-VL-Plus & 315/780 (70.6\%) & 88/780 (19.7\%) & 43/780 (9.6\%) & 446/780 \\
Qwen-VL-Max & 260/780 (48.5\%) & 203/780 (37.9\%) & 73/780 (13.6\%) & 536/780 \\
GLM-4.5V & 324/780 (55.6\%) & 223/780 (38.3\%) & 36/780 (6.2\%) & 583/780 \\
\hline
Six-model Overall & 1336/4680 (59.1\%) & 684/4680 (30.3\%) & 239/4680 (10.6\%) & 2259/4680 \\
\hline
\end{tabular}
\end{table}

\begin{table}[h]
\centering
\footnotesize
\setlength{\tabcolsep}{5pt}
\caption{Subtype distribution for the 104 unresolved Stage-2 extraction parse failures. These parse failures are part of the compact \emph{Schema / interrupted} official category in Table~\ref{tab:app_exp_failure_breakdown_official}; the remaining 70 cases in that compact category are follow-ups that were not run or were interrupted. Shares are normalized within the 104 parse failures.}
\label{tab:app_exp_failure_schema}
\begin{tabular}{lrr}
\hline
Schema subtype & Count & Share (\%) \\
\hline
Incomplete JSON & 55 & 52.9 \\
Other malformed JSON & 26 & 25.0 \\
Python literal, not JSON & 20 & 19.2 \\
Numeric keys not quoted & 2 & 1.9 \\
Empty response & 1 & 1.0 \\
\hline
\end{tabular}
\end{table}

\textbf{Network Optimization results.} In the representative run, Gemini and GPT-5.4 solve 68.7\% and 38.7\% of network instances using the original text and images. With the correct data supplied directly (oracle reading), their scores rise to 86.7\% and 88.0\%. This smaller gap suggests that reading the diagrams and their node and edge values contributes to the original performance difference.

\textbf{Failure-count interpretation.} Tables~\ref{tab:app_exp_failure_breakdown_official}, \ref{tab:app_exp_failure_breakdown_oracle}, and~\ref{tab:app_exp_failure_schema} support four observations.

\noindent\textbf{(1) Most failed cases also have data extraction errors in Stage 2.} Of the 4680 official outputs pooled over the six general-purpose MLLMs, 2049 also have data extraction errors in Stage 2; among the 3302 failed official outputs, this accounts for 62.1\%. This is the largest official category for every general-purpose model, ranging from 51.9\% for Qwen-VL-Max to 82.0\% for GPT-5.4. The evaluation program can read these extracted data, but they do not match the reference data in \texttt{instance\_data.json}.

\noindent\textbf{(2) Correct extraction does not imply correct modeling.} Another 1079/4680 official outputs have matched Stage-2 extractions but still fail the solver-grounded check; within failed official outputs, this is 32.7\%. This downstream-after-matched-extraction share is especially large for Qwen3-VL-Plus (42.4\%) and Qwen-VL-Max (44.2\%). Stage 2 asks the model to read the input again. Getting the data right on this second attempt does not mean it read them correctly in Stage 1.

\noindent\textbf{(3) Oracle-reading failures are mostly substantive downstream failures.} In the oracle-reading view, the model is given the ground-truth public instance record directly. Of the 4680 oracle-reading outputs, 2259 fail; among these failed oracle-reading outputs, 1336 (59.1\%) are modeling/algorithmic mismatches, 684 (30.3\%) are runtime failures, and 239 (10.6\%) are other code/artifact failures. The modeling/algorithmic share is the largest oracle-reading failure category for all six models, while runtime failures are particularly visible for Qwen-VL-Max (37.9\%) and GLM-4.5V (38.3\%). Thus the downstream problem is not merely that code crashes; the dominant category is a solver-running or solver-facing artifact that encodes the wrong optimization logic or uses an incomplete solution procedure.

\noindent\textbf{(4) Unresolved diagnostics are small and intentionally not reassigned.} The compact \emph{Schema / interrupted} row contains 174/4680 official outputs, or 5.2\% of failed official outputs. Of these, 104 are extraction parse failures and 70 are follow-ups that were not run or were interrupted. Table~\ref{tab:app_exp_failure_schema} shows that most parse failures are incomplete JSON (52.9\%), other malformed JSON (25.0\%), or Python-literal payloads (19.2\%). We keep these cases separate instead of forcing them into reading or downstream categories, because the diagnostic follow-up itself is unusable. This avoids overstating either source of failure.

\subsection{Failure Analysis for Math-Specialized MLLMs}
\label{subsec:app_math_specialized_failure_analysis}

This subsection expands the RQ2 analysis in Section~\ref{subsec:exp_overall}. The three math-specialized MLLMs are designed for multimodal mathematical reasoning, code-oriented reasoning, or process supervision, but MM-OptBench asks for a different artifact: a complete optimization formulation together with solver-executable code that returns the verified optimum. We therefore analyze these models separately from the six general-purpose MLLMs used in the main family, difficulty, and failure-attribution figures.

\begin{table}[h]
\centering
\footnotesize
\setlength{\tabcolsep}{3pt}
\caption{Math-specialized MLLM diagnostic summary with base-model context. Official pass@1 and pass@4 are zero across five runs. Valid Code Rate is summarized over the same five-run protocol; oracle-reading is diagnostic and is not merged into the official score.}
\label{tab:app_math_specialized_summary}
\resizebox{\linewidth}{!}{
\begin{tabular}{llcccl}
\hline
Model & Base model & Official success & Valid Code Rate (\%) & Oracle success & Dominant official failure \\
\hline
MathCoder-VL-8B~\citep{wang2025mathcoder} & InternVL2-8B & 0/780 & 0.0 & 0/780 & Other code / artifact (72.7\%) \\
MM-Eureka~\citep{meng2025mm} & Qwen2.5-VL-7B-Instruct & 0/780 & 6.6 & 1/780 & Execution-level (87.3\%) \\
MM-PRM~\citep{du2025mm} & InternVL2.5-8B & 0/780 & 0.0 & 0/780 & Other code / artifact (100.0\%) \\
\hline
\end{tabular}
}
\end{table}

\FloatBarrier

\textbf{Summary and base-model context.} Table~\ref{tab:app_math_specialized_summary} records the backbone associated with each specialized checkpoint. MathCoder-VL-8B is built on InternVL2-8B, MM-Eureka uses Qwen2.5-VL-7B-Instruct, and MM-PRM uses InternVL2.5-8B. These results describe the three models tested here. They do not tell us whether training for mathematical reasoning improves optimization modeling, because model size and the ability to write solver code may also affect performance. All three models solve zero official instances, and oracle-reading remains essentially zero even when the ground-truth public instance record is supplied: MathCoder-VL-8B solves $0/780$ diagnostic oracle-reading cases, MM-PRM solves $0/780$, and MM-Eureka solves only $1/780$. This means the failure is not explained solely by multimodal instance extraction; once the visual-reading burden is removed, these models still rarely construct a solver-correct optimization artifact.

\FloatBarrier

\begin{figure}[h]
\centering
\includegraphics[width=0.95\linewidth]{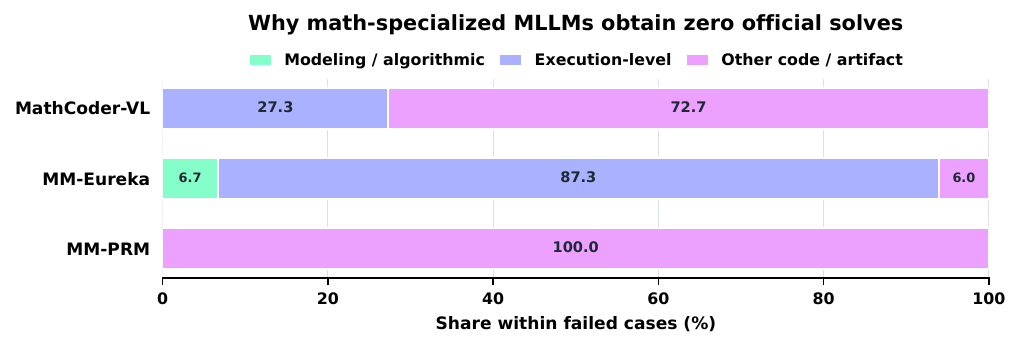}
\caption{Failure composition for the three math-specialized MLLMs using the taxonomy labels from Appendix~\ref{app:failure_taxonomy_details}. Percentages are normalized over 780 official failed cases per model.}
\label{fig:app_exp_math_specialized_failures}
\end{figure}

\FloatBarrier

\begin{table}[h]
\centering
\footnotesize
\setlength{\tabcolsep}{4pt}
\caption{Engineering-level failure subtypes underlying Figure~\ref{fig:app_exp_math_specialized_failures}. The taxonomy-group column is the aggregation used in the figure; the subtype column gives the lower-level execution log used for interpretation. Percentages are normalized over 780 official failed cases per model.}
\label{tab:app_math_specialized_engineering_subtypes}
\begin{tabular}{lllrr}
\hline
Model & Taxonomy group in Fig.~\ref{fig:app_exp_math_specialized_failures} & Engineering subtype & Count & Share (\%) \\
\hline
\multirow{3}{*}{MathCoder-VL-8B}
& Other code / artifact & Response contract/schema & 567 & 72.7 \\
& Execution-level & Syntax error & 168 & 21.5 \\
& Execution-level & Dependency/import error & 45 & 5.8 \\
\hline
\multirow{6}{*}{MM-Eureka}
& Execution-level & Runtime error & 558 & 71.5 \\
& Execution-level & Syntax error & 118 & 15.1 \\
& Modeling / algorithmic & Modeling/solving error & 52 & 6.7 \\
& Other code / artifact & Response contract/schema & 28 & 3.6 \\
& Other code / artifact & Output-format error & 19 & 2.4 \\
& Execution-level & Dependency/import error & 5 & 0.6 \\
\hline
MM-PRM & Other code / artifact & Response contract/schema & 780 & 100.0 \\
\hline
\end{tabular}
\end{table}

\FloatBarrier

\textbf{Math-specialized failure composition.} Figure~\ref{fig:app_exp_math_specialized_failures} and Table~\ref{tab:app_math_specialized_engineering_subtypes} support four observations.

\noindent\textbf{(1) All three math-specialized MLLMs fail before producing any official solve.} The figure normalizes over 780 failed official cases per model, because MathCoder-VL-8B, MM-Eureka, and MM-PRM each solve 0/780 instances. The plot therefore explains where the failed attempts break down, rather than comparing successful and unsuccessful subsets.

\noindent\textbf{(2) MathCoder-VL-8B and MM-PRM mostly fail at the artifact-contract level.} MathCoder-VL-8B assigns 72.7\% of cases to \emph{Other code / artifact}, and MM-PRM assigns 100.0\% to this category. Table~\ref{tab:app_math_specialized_engineering_subtypes} shows that these are response-contract/schema failures: the model usually does not expose a benchmark-usable solver artifact, so the evaluation cannot even reach a substantive solver-correctness check.

\noindent\textbf{(3) MM-Eureka has a different bottleneck: it more often reaches execution but fails there.} Its dominant category is \emph{Execution-level} failure (87.3\%), with only 6.7\% categorized as \emph{Modeling / algorithmic}. The lower-level logs show that runtime errors account for 558/780 cases (71.5\%), syntax errors for 118/780 (15.1\%), and dependency/import errors for 5/780 (0.6\%). This explains why MM-Eureka has a small nonzero Valid Code Rate but still no official solves: it often enters the code-generation regime, but the produced artifact is not robust enough to run and return the verified optimum.

\noindent\textbf{(4) The plotted categories are taxonomy-level groups, not a new taxonomy.} Table~\ref{tab:app_math_specialized_engineering_subtypes} expands the engineering logs behind Figure~\ref{fig:app_exp_math_specialized_failures}: response-contract/schema and output-format errors are grouped as \emph{Other code / artifact}; syntax, runtime, and dependency/import errors are grouped as \emph{Execution-level}; and modeling/solving errors are grouped as \emph{Modeling / algorithmic}. The table should therefore be read as a diagnostic expansion of the figure.

\begin{table}[h]
\centering
\footnotesize
\setlength{\tabcolsep}{5pt}
\caption{Oracle-reading easy-case diagnostic for MM-Eureka's base model, Qwen2.5-VL-7B-Instruct. The sweep covers all 26 subcategories with 10 easy cases each. Failure columns report counts.}
\label{tab:app_qwen25vl7b_oracle_easy}
\begin{tabular}{lrrrrr}
\hline
Setting & Cases & Solved & Other code / artifact & Execution-level & Modeling / algorithmic \\
\hline
Full easy sweep & 260 & 0 & 59 & 201 & 0 \\
\hline
\end{tabular}
\end{table}

\FloatBarrier

\textbf{MM-Eureka and its base model.} Table~\ref{tab:app_qwen25vl7b_oracle_easy} checks whether MM-Eureka's failure is specific to the specialized checkpoint or already present in its Qwen2.5-VL-7B-Instruct base. To remove visual-reading difficulty, this diagnostic uses oracle-reading: the model receives the ground-truth \texttt{instance\_data.json} and only needs to formulate and implement the solver artifact. Even under this easier condition, the base model solves $0/260$ easy cases across all 26 subcategories. The dominant failures are execution-level: 201/260 cases (77.3\%) fail during compilation, dependency resolution, or runtime execution, while 59/260 cases (22.7\%) fail as other code/artifact errors before a usable solver artifact is exposed. Since the public instance data are supplied exactly, these errors cannot be attributed to multimodal extraction. We therefore did not extend this base-model diagnostic to medium or hard instances: those tiers add structural coupling and visual/modeling complexity, so the evaluated Qwen2.5-VL-7B-Instruct checkpoint already fails to generate solver-correct artifacts on this easy oracle-reading subset. MM-Eureka also exhibits this downstream difficulty under the same interface.

\textbf{Implication.} These results show that the three models trained for mathematical reasoning still struggle to formulate and solve the optimization problems in MM-OptBench. MM-OptBench requires the model to satisfy a stricter contract than answering a visual math question: it must maintain the public instance data, formulate the correct variables, constraints, and objective, and implement a solver-compatible procedure. The Qwen2.5-VL-7B-Instruct oracle-reading check is especially revealing: even after removing the hardest multimodal reading step and restricting to easy instances, the base model produces no solver-correct artifacts. The math-specialized failures therefore highlight a gap between multimodal mathematical reasoning benchmarks and solver-grounded optimization modeling.

\subsection{Reference Runtime and Evaluation Timeout Analysis}
\label{subsec:app_runtime_analysis}

This subsection explains the runtime evidence behind the 1800-second execution cap used for generated solver artifacts in App.~\ref{subsec:app_exp_setup_details}. The key distinction is between \emph{reference-instance construction} and \emph{model-output evaluation}. During construction, candidate instances are retained only after family-specific reference solvers certify an optimum under construction-time safeguards. During evaluation, the timeout is deliberately more permissive because LLM-generated code may be less optimized than the reference implementation. We therefore separate construction-time controls, observed reference runtimes, long-tail reference cases, and the official generated-artifact timeout.

\begin{table}[h]
\centering
\footnotesize
\setlength{\tabcolsep}{5pt}
\renewcommand{\arraystretch}{1.15}
\caption{Construction-time solver-budget controls used during benchmark generation. These controls define feasible instance-generation regimes; they are not the official timeout for evaluating generated model outputs.}
\label{tab:app_runtime_construction_budgets}
\renewcommand{\tabularxcolumn}[1]{m{#1}}
\begin{tabularx}{\linewidth}{>{\raggedright\arraybackslash}m{3.2cm} >{\raggedright\arraybackslash}X}
\hline
Control & Role during construction \\
\hline
Guideline-level solvability targets
& Family guidelines choose instance scales intended to be exactly solvable by reference methods, typically in seconds to tens of seconds (e.g., common targets such as $<1$s, $<4$s, $<8$s, or $<15$--$30$s). \\
Generator process guards
& Outer guards reject pathological candidate runs during generation. Detected guards are usually far below the official evaluation cap: 19/33 use $\leq180$s, 24/33 use $\leq300$s, and 29/33 use $\leq600$s. \\
\hline
\end{tabularx}
\end{table}

\FloatBarrier

\begin{table}[h]
\centering
\footnotesize
\setlength{\tabcolsep}{4pt}
\caption{Reference-solver runtime summary for the 780 released MM-OptBench instances. The table reports family-level runtime quantiles and the number of instances whose reference solve exceeds 180 seconds.}
\label{tab:app_reference_runtime_by_family}
\begin{tabular}{lrrrrr}
\hline
Family & Instances & Median (s) & p95 (s) & Max (s) & \# $>180$s \\
\hline
Network Optimization & 150 & 0.021 & 0.048 & 0.405 & 0 \\
Location / Covering / Assignment & 150 & 0.005 & 0.050 & 0.489 & 0 \\
Scheduling / Sequencing & 120 & 0.078 & 36.853 & 344.151 & 2 \\
Multi-Period / System Planning & 120 & 0.019 & 34.923 & 89.345 & 0 \\
Routing / Tour Optimization & 120 & 0.163 & 12.224 & 90.207 & 0 \\
Combinatorial / Logical Models & 120 & 0.010 & 0.061 & 0.364 & 0 \\
\hline
Overall & 780 & 0.019 & 8.299 & 344.151 & 2 \\
\hline
\end{tabular}
\end{table}

\FloatBarrier

\begin{table}[h]
\centering
\footnotesize
\setlength{\tabcolsep}{4pt}
\caption{Longest reference runtimes among released instances. The long tail is concentrated in a few exact scheduling, routing, and planning instances; even these cases remain far below the 1800-second evaluation cap.}
\label{tab:app_reference_runtime_long_tail}
\begin{tabular}{lllr}
\hline
Instance & Family & Subcategory / difficulty & Runtime (s) \\
\hline
\texttt{pms\_m\_010} & Scheduling / Sequencing & Parallel-machine scheduling / medium & 344.151 \\
\texttt{fhfss\_h\_010} & Scheduling / Sequencing & Flexible hybrid flow-shop / hard & 211.059 \\
\texttt{fhfss\_h\_003} & Scheduling / Sequencing & Flexible hybrid flow-shop / hard & 106.922 \\
\texttt{tsptw\_m\_008} & Routing / Tour Optimization & TSP with time windows / medium & 90.207 \\
\texttt{irp\_h\_008} & Multi-Period / System Planning & Inventory routing / hard & 89.345 \\
\hline
\end{tabular}
\end{table}

\FloatBarrier

\begin{table}[h]
\centering
\footnotesize
\setlength{\tabcolsep}{5pt}
\renewcommand{\arraystretch}{1.15}
\caption{Relation between released-reference runtimes and the official generated-artifact evaluation timeout.}
\label{tab:app_runtime_evaluation_timeout}
\renewcommand{\tabularxcolumn}[1]{m{#1}}
\begin{tabularx}{\linewidth}{>{\raggedright\arraybackslash}m{3.2cm} >{\raggedright\arraybackslash}X}
\hline
Quantity & Interpretation \\
\hline
Released reference solves
& Across the 780 released instances, the median reference runtime is 0.019s, the p95 is 8.299s, and the maximum is 344.151s; no released instance exceeds 600s. \\
Official generated-artifact timeout
& Each LLM-generated solver artifact receives 1800s. This is a conservative grace period for slower generated implementations, not evidence that the released benchmark requires near-timeout reference solving. \\
\hline
\end{tabularx}
\end{table}

\FloatBarrier

\textbf{Runtime interpretation.} Tables~\ref{tab:app_runtime_construction_budgets}, \ref{tab:app_reference_runtime_by_family}, \ref{tab:app_reference_runtime_long_tail}, and~\ref{tab:app_runtime_evaluation_timeout} support four observations.

\noindent\textbf{(1) Construction-time budgets and evaluation-time budgets serve different purposes.} The construction pipeline uses family-specific guideline targets and generator process guards to reject infeasible, degenerate, or excessively expensive candidates. These budgets are part of instance generation and quality control. The 1800-second evaluation timeout is not used to define the released instance difficulty; it is applied later to LLM-generated solver artifacts.

\noindent\textbf{(2) The released benchmark is not a collection of near-timeout reference instances.} Across the 780 released instances, the median reference runtime is 0.019 seconds and the p95 is 8.299 seconds. Only two instances exceed 180 seconds, none exceed 600 seconds, and none approach the 1800-second evaluation cap.

\noindent\textbf{(3) The runtime long tail is narrow and interpretable.} The slowest cases are concentrated in exact scheduling, routing, and system-planning subcategories, such as parallel-machine scheduling, flexible hybrid flow-shop scheduling, TSP with time windows, and inventory routing. This is consistent with the benchmark design: these families require exact reasoning over global temporal or routing structure, but the retained instances still remain comfortably below the evaluation cap.

\noindent\textbf{(4) The 1800-second cap is conservative rather than permissive toward incorrect modeling.} A correct but less optimized generated implementation may require more time than the reference solver, so a short timeout would conflate modeling failure with implementation efficiency. The 1800-second cap gives generated code a generous opportunity to run, while correctness still follows the task-specific criterion in Section~\ref{subsec:setting_protocol}. If a generation times out under this cap, it is reasonable to treat it as an invalid solver artifact for MM-OptBench.

\subsection{Revision and Robustness Experiments}
\label{subsec:app_controlled_evaluations}

We examine whether one round of revision improves accuracy, and how performance changes when images are downscaled or the sampling temperature is varied.

\subsubsection{One-Round Revision}
\label{subsubsec:cr_revision}

At $T=0$, we ask Qwen3-VL-Plus, Qwen-VL-Max, and GLM-4.5V to revise each initially incorrect answer once. For each model, both revision settings start from the same answers and use the same output-token limit. In \emph{self-revision}, the model reviews its answer without execution feedback. In \emph{solver-feedback revision}, it also receives the execution status, error messages, timeout information, or returned output, but no reference answer. Table~\ref{tab:cr_revision} reports the percentage of all 780 instances solved either initially or after revision. This measures the benefit of revising an existing answer; pass@4 measures success across independent answers. Self-revision improves accuracy for all three models, and solver feedback gives a further, smaller improvement.

\begin{table}[htbp]
\centering
\small
\caption{Accuracy (\%) before and after one revision round on all 780 instances. Revised scores include instances solved initially.}
\label{tab:cr_revision}
\begin{tabular}{lrrr}
\toprule
Model & Initial & Self-revision & Solver feedback \\
\midrule
Qwen3-VL-Plus & 18.1 & 21.9 & 22.9 \\
Qwen-VL-Max & 15.6 & 17.6 & 17.7 \\
GLM-4.5V & 13.1 & 15.5 & 16.8 \\
\bottomrule
\end{tabular}
\end{table}

\subsubsection{Image Downscaling}
\label{subsubsec:cr_downscaling}

We test the same three models at $T=0$ after multiplying image width and height by $1.0$, $0.8$, $0.6$, $0.4$, or $0.2$. We preserve the aspect ratio and file format, without cropping or sharpening, and keep the prompts, output-token limits, and evaluation code unchanged. The original-resolution condition is rerun alongside the downscaled conditions. Table~\ref{tab:cr_downscaling} reports accuracy on all 780 instances for every model and image scale. Average accuracy is 16.1\% at the original resolution and 16.8\% at $0.8\times$. At $0.8\times$, Qwen-VL-Max and GLM-4.5V improve slightly, while Qwen3-VL-Plus changes little. Thus, a modest reduction in image size does not consistently reduce accuracy. Further downscaling lowers average accuracy to 15.6\%, 11.0\%, and 2.4\% at $0.6\times$, $0.4\times$, and $0.2\times$, respectively, with all three models performing substantially worse at the two smallest scales.

\begin{table}[htbp]
\centering
\small
\caption{Accuracy (\%) on all 780 instances at different image scales. Both image width and height are multiplied by the stated factor. Average is the mean accuracy across the three models.}
\label{tab:cr_downscaling}
\begin{tabular}{crrrr}
\toprule
Image scale & Qwen3-VL-Plus & Qwen-VL-Max & GLM-4.5V & Average \\
\midrule
$1.0\times$ & 21.0 & 14.5 & 12.8 & 16.1 \\
$0.8\times$ & 20.8 & 15.5 & 14.0 & 16.8 \\
$0.6\times$ & 19.4 & 15.8 & 11.5 & 15.6 \\
$0.4\times$ & 13.3 & 11.9 & 7.7 & 11.0 \\
$0.2\times$ & 3.2 & 1.5 & 2.4 & 2.4 \\
\bottomrule
\end{tabular}
\end{table}

\subsubsection{Temperature Sensitivity}
\label{subsubsec:cr_temperature}

Table~\ref{tab:cr_temperature} compares pass@4 at six sampling temperatures, $T\in\{0,0.2,0.4,0.6,0.8,1.0\}$. All six general-purpose models use the same 78 instances: one from each of the 26 subcategories at each of the three difficulty levels. No single temperature gives the highest score for all six models. The main stochastic evaluation uses a common temperature of $T=0.4$ for all models.

\begin{table}[htbp]
\centering
\small
\setlength{\tabcolsep}{5pt}
\caption{pass@4 (\%) at different sampling temperatures on the same 78 instances.}
\label{tab:cr_temperature}
\begin{tabular}{crrrrrr}
\toprule
$T$ & GPT-5.4 & \makecell{Gemini\\3.1 Pro} & \makecell{Claude\\Sonnet 4.6} & \makecell{Qwen3-\\VL-Plus} & \makecell{Qwen-\\VL-Max} & GLM-4.5V \\
\midrule
0.0 & 61.5 & 65.1 & 32.2 & 37.2 & 20.5 & 27.8 \\
0.2 & 59.0 & 62.2 & 29.1 & 34.6 & 21.8 & 26.7 \\
0.4 & 65.4 & 64.4 & 31.3 & 34.6 & 21.8 & 29.1 \\
0.6 & 59.0 & 63.5 & 30.4 & 38.5 & 21.8 & 29.0 \\
0.8 & 62.8 & 65.2 & 32.1 & 34.6 & 22.9 & 27.9 \\
1.0 & 62.8 & 63.1 & 30.0 & 37.2 & 17.9 & 26.4 \\
\bottomrule
\end{tabular}
\end{table}

\FloatBarrier

\section{Limitations and Impact Statement}
\label{app:limitations_impact}

Our study has several limitations. First, MM-OptBench adopts solver-grounded execution as its primary notion of correctness. This makes evaluation clear, auditable, and practically meaningful, but it does not fully capture all forms of structural equivalence between generated and canonical formulations. Some models may produce solver-correct implementations with substantially different symbolic organization, while other structural defects may only become evident under broader robustness or scenario testing than the benchmark currently performs. This limitation is not unique to MM-OptBench: existing optimization-modeling benchmarks also rely on objective agreement, executability, formulation matching, or solver-checked outputs as practical proxies for correctness, and therefore face the same difficulty of scoring all mathematically equivalent formulations and latent structural defects~\citep{Ramamonjison2023NL4Opt,huang2025orlm,huang2024mamo,lu2025optmath,wang2025orgeval,li2026constructing}.

Second, although MM-OptBench spans 6 major families and 26 subcategories, it does not exhaust the full range of optimization-modeling scenarios encountered in practice. Real deployments may involve longer documents, richer enterprise data interfaces, interactive human-in-the-loop refinement, domain-specific modeling languages, or substantially larger industrial instances. MM-OptBench is a controlled testbed over application-inspired rendered artifacts; its results do not establish transfer to naturally occurring, noisy, interactive, or industrial-scale inputs.

Third, our experiments focus on prompting-based evaluation of frontier general-purpose MLLMs and a small set of math-specialized MLLMs. We additionally examine one-round revision, but do not study MM-OptBench-specific fine-tuning, retrieval augmentation, or broader agentic modeling systems. An independent human baseline is also not included: the task requires joint visual interpretation, optimization modeling, and solver programming, and benchmark authors would not provide an independent comparison. Accordingly, the reported results should be interpreted as baseline performance for the benchmark setting, rather than as upper bounds on multimodal optimization-modeling performance.

The potential positive impact of this line of research is substantial. If multimodal optimization modeling becomes reliable, it could lower the expertise barrier that currently limits the use of optimization in domains such as logistics, manufacturing, planning, and energy systems, making formal decision support more accessible to a broader range of users and organizations. From a scientific perspective, MM-OptBench also encourages a more rigorous style of multimodal evaluation centered on verifiable structured artifacts rather than surface-level answer generation.

At the same time, this capability carries meaningful risks. Optimization models are often used in high-stakes settings, and flawed formulations may yield infeasible, unsafe, unfair, or economically harmful decisions while still appearing superficially plausible. A system that generates convincing but incorrect solver code could therefore create a false sense of reliability, especially for non-expert users. For this reason, MM-OptBench should be understood as an evaluation resource rather than evidence that current models are ready for autonomous deployment in decision-critical environments. We hope the benchmark helps promote progress toward stronger solver-grounded verification, better structural interpretability, clearer human oversight, and safer standards for optimization-oriented AI systems.

\end{document}